\pdfoutput=1
\documentclass[11pt]{article}
\usepackage{pgfplots}
\usepackage{enumerate}
\usepackage[OT1]{fontenc}
\usepackage{amsmath,amssymb}
\usepackage{dsfont}
\usepackage{pgfplots}
\usepackage{smile}
\usepackage{multirow}
\usepackage{rotating}
\usepackage{enumerate}
\usepackage{esvect}
\usepackage{tikz}
\usetikzlibrary{patterns}
\usetikzlibrary{arrows}
\usetikzlibrary{bayesnet}
\usepackage{authblk}
\usepackage{caption}
\usepackage{bbold}
\usepackage[colorlinks,
            linkcolor=red,
            anchorcolor=blue,
            citecolor=blue
            ]{hyperref}
 \usepackage{algorithm}
\usepackage{algorithmic}

\usepackage{subcaption}

\def\supp{\mathop{\text{supp}}}

\long\def\comment#1{}

\def\cS{{\mathcal{S}}}

\providecommand{\norm}[1]{\vvvert#1\vvvert}

\newcommand{\bel}{\begin{eqnarray}\label}
\newcommand{\eel}{\end{eqnarray}}
\newcommand{\bes}{\begin{eqnarray*}}
\newcommand{\ees}{\end{eqnarray*}}

\def\rd{\mathrm{d}}

\let\hat\widehat
\let\tilde\widetilde

\def\bigeps{\mathcal{E}_n}

\def\E{{\mathbb E}}

\def\supp{\mathop{\text{supp}\kern.2ex}}

\def\given{{\,|\,}}
\def\Given{\, \Big| \,}

\def\supp{\mathop{\text{supp}}}

\def\piestar{{\pi^*}}

\def\pie{{\pi}}

\def\piepessi{{\hat\pie}}
\def\prob{{\mathbb{P}}}

\theoremstyle{plain}

\usepackage{mathrsfs}
\usepackage{fullpage}

\usepackage{hyperref}
\usepackage[protrusion=false,expansion=true]{microtype}
\def\##1\#{\begin{align}#1\end{align}}
\def\$#1\${\begin{align*}#1\end{align*}}

\usepackage{undertilde}

\theoremstyle{mytheoremstyle}

\def\nend{\nonumber\\}

\def\tpr{\tilde{p}}

\def\pr{p}

\def\vT{{\mathcal{T}}}

\def\vX{X}

\def\vZ{Z}
\def\vh{h}
\def\vH{\cH}

\def\vTheta{{\Theta}}

\def\vcL{\cL}

\def\pob{{\pr_{\text{ob}}}}

\newcommand\hpessi[1]{{\hat h^{#1}}}

\def\hstaralpha{h^{*}_{\alpha}}

\newcommand{\RNum}[1]{\uppercase\expandafter{\romannumeral #1\relax}}
\def\sF{\mathscr{F}}

\newcommand\pin[1]{{p_{\text{in}}^#1}}

\usepackage{subcaption}
\usetikzlibrary{shapes, decorations, arrows, calc, arrows.meta, fit, positioning}
\tikzset{
    -Latex, auto, node distance =1 cm and 1 cm,semithick,
    state/.style ={circle, draw, minimum width = 1 cm},
    missingstate/.style ={circle, draw, minimum width = 1 cm, fill=lightgray},
    hiddenstate/.style ={circle, draw, minimum width = 1 cm, fill=lightgray, text=black},
    point/.style = {circle, draw, inner sep=0.04cm,fill,node contents={}},
    bidirected/.style={Latex-Latex,dashed},
    el/.style = {inner sep=2pt, align=left, sloped},
    normal/.style={-stealth', line width=1},
    double/.style={stealth'-stealth', line width=1},
}
\DeclareUnicodeCharacter{FF0C}{,}

\title{\huge Quantile-Optimal Policy Learning under Unmeasured Confounding} 

\author[1]{Zhongren Chen\thanks{ Emails: \texttt{\{zhongren.chen, siyu.chen.sc3226, xiaohong.chen, zhuoran.yang\}@yale.edu}}}

\author[1]{Siyu Chen$^*$} 

\author[2]{Zhengling Qi\thanks{Email: \texttt{qizhengling@gwu.edu}}}

\author[3]{Xiaohong Chen$^*$}  

\author[1]{Zhuoran Yang$^*$}

\affil[1]{
\small
\textit{Department of Statistics and Data Science, Yale University}}
\affil[2]{
\small
\textit{Department of Decision Sciences, George Washington University}}
\affil[3]{
\small
\textit{Cowles Foundation for Research in Economics, Yale University}}
\date{}
\pgfplotsset{compat=1.17}

\begin{document}
\maketitle
\vspace{-25pt}
\begin{abstract}
We study quantile-optimal policy learning where the goal is to find a policy whose reward distribution has the largest $\alpha$-quantile for some $\alpha \in (0, 1)$. 
We focus on the offline setting whose generating process involves unobserved confounders. Such a problem suffers from three main challenges: (i) nonlinearity of the quantile objective as a functional of the reward distribution,  (ii) unobserved confounding issue, and  (iii) insufficient coverage of the offline dataset. To address these challenges, we propose a suite of causal-assisted policy learning methods that provably enjoy strong theoretical guarantees under mild conditions. In particular, to address (i) and (ii), using causal inference tools such as instrumental variables and negative controls, we propose to estimate the quantile objectives by solving nonlinear functional integral equations. Then we adopt a minimax estimation approach with nonparametric models to solve these integral equations, and propose to construct conservative policy estimates that address (iii). The final policy is the one that maximizes these pessimistic estimates. In addition, we propose a novel regularized policy learning method that is more amenable to computation. Finally, we prove that the policies learned by these methods are $\tilde{\cO}(n^{-1/2})$ quantile-optimal under a mild coverage assumption on the offline dataset. Here, $\tilde {\cO}(\cdot) $ omits poly-logarithmic factors. To the best of our knowledge, we propose the first sample-efficient policy learning algorithms for estimating the quantile-optimal policy when there exist unmeasured confounding. 

\end{abstract}

\section{Introduction}

Offline reinforcement learning (RL) \citep{levine2020offline, prudencio2023survey} aims to learn the optimal decision-making policies from pre-collected datasets, where the learner has no control over the data collection process. This lack of control is further complicated by the fact that many real-world datasets suffer from incomplete information due to unmeasured confounders \citep{rubin1974estimating}—factors influencing both the actions and rewards, yet not observed by the learner. 
These unobserved confounders naturally arise in the \emph{partially observable} setting, 
where the behavior policy used to generate the actions is contingent on the state, while only noisy observations of the state are recorded in the data. 
For example, in healthcare, electronic health records (EHR) may lack key details about a patient’s condition or treatment environment \citep{hernan2016using}, but decisions about treatment still need to be optimized based on the available data.

In many decision-making scenarios, rather than expected rewards, we are interested in optimizing more nuanced objectives involving quantiles of the 
reward distribution. 
For instance, to measure the efficacy of job training programs, it may be more meaningful to optimize for the median income increase, as the mean could be skewed by a few extreme cases. 
Moreover, quantile-based objectives naturally arise when some notion of fairness is of interest \citep{yang2019fair, liu2022conformalized}.

In this paper, we tackle the problem of quantile-optimal policy learning in the offline setting where unmeasured confounders play a critical role. Our goal is to find a policy that maximizes the $\alpha$-th quantile of the reward distribution, conditioned on the context. Such a policy is learned based on a pre-collected offline dataset that involves unobserved confounders, which are hidden variables that simultaneously affect the rewards and actions stored in the data. 
  
This problem is particularly challenging due to three core issues: (i) the quantile objective is a nonlinear functional of the reward distribution, making statistical learning and estimation more difficult; (ii) unmeasured confounders introduce bias into the learning problem, which can lead to misleading results if not properly accounted for; and (iii) the offline dataset often lacks full coverage, meaning that the distribution of the collected data might have insufficient overlap with that induced by some candidate policy, making it challenging to evaluate the performance of that policy. 

 To address these challenges, we propose a suite of novel, causal-assisted policy learning approaches. Our methods leverage powerful causal inference techniques, such as instrumental variables (IV) \citep{angrist1996identification, baiocchi2014instrumental} and negative controls (NC) \citep{lipsitch2010negative, tchetgen2020introduction}, to account for unmeasured confounding. 
 Leveraging these causal inference tools, we reduce the quantile-based learning problem to 
 the problem of solving {\bf nonlinear functional integral equations}, which are then solved as a minimax estimation in nonparametric nonlinear conditional moment models. 
  This enables us to address Challenges (i) and (ii). 
  Furthermore, to handle Challenge (iii), we propose to adopt the pessimism principle \citep{jin2021pessimism, xie2021bellman, rashidinejad2021bridging, buckman2020importance, lu2022pessimism}, which enables us to relax the requirement on data coverage. 
  We introduce two algorithms based on the pessimism idea --- a solution set constrained version and a regularized version. We note that pessimism penalizes candidate policies with large uncertainty and enables us to only focus on the policies that are covered by the dataset.
  In particular, both our algorithms work as long as the offline data set has sufficient coverage over the optimal-policy induced distribution, regardless of whether the data set contains sufficient information about the suboptimal policies. 

  In this paper, we use quantile regret as the loss function. We establish that the proposed algorithms achieve sample-efficient  $\widetilde O (n^{-1/2})$ suboptimality regret bounds, where $n$ is the sample size. 
  Here, $\widetilde O(\cdot)$ may include some logarithmic factors. 
  Compared to existing works that study offline policy learning with confounded datasets aiming to maximize expected rewards, our analysis is more complicated due to the nonlinear nature of the quantile-based objective function. 
  In particular, to get the suboptimality regret bounds, we leverage the local curvature of the nonlinear operator in the functional estimating equation. We examine the statistical error of the estimated quantile-based objective in terms of a local norm, defined using the first-order Taylor expansion of the nonlinear operator. 
  To the best of our knowledge, we establish the first provably sample-efficient algorithms for quantile-optimal policy learning under unmeasured confounding and insufficient support.

\paragraph{Our contributions.} 
Our contributions are threefold. 
First, leveraging causal tools such as IV and NC, we reduce the problem of quantile-optimal policy learning to a minimax estimation problem with nonlinear functional integral equations. To the best of our knowledge, we are the first to establish integral equations for the structural quantile function using NC. Second, to tackle the challenge of insufficient data coverage, we incorporate the principle of pessimism into our learning algorithm. 
Specifically, we introduce two versions of pessimistic algorithms, which are both proved to be sample efficient.
Third, to analyze the pessimistic algorithms for quantile-optimal policy learning, we need to construct confidence sets for the nonlinear quantile-based minimax objective. 
To this end, we establish a novel statistical analysis framework that leverages the local curvature of the nonlinear operator in the functional estimating equation. Our analysis can be extended to general nonlinear functional integral equations, which may be of independent interest.

\subsection{Related Literature}
Our work belongs to the intersection of offline decision making, quantile policy learning, causal inference with unobserved confounders, and limited support.

\paragraph{Offline decision making}
 There is already a large literature on offline reinforcement learning \citep{yin2022near, uehara2021finite, jin2021pessimism, xie2021bellman, rashidinejad2021bridging} and offline contextual bandits \citep{li2012unbiased, lee2021optidice, metevier2019offline}. Within this literature, \citet{cassel2023general, zhu2020thompson, prashanth2020concentration} study nonlinear objective functions in contextual bandits without confounding bias. \citet{chen2023unified, wang2021provably} investigate causal policy learning with confounding variables, but using linear objective functions. Our work makes a new contribution to this literature by considering nonlinear objective functions with unobserved confounding and insufficient data coverage. 

 \paragraph{Quantile policy learning}
 Quantile treatment effect has been extensively studied in Econometrics \citep{abadie2002instrumental, chernozhukov2005iv,horowitz2007nonparametric, chernozhukov2008instrumental, chen2012estimation,   gagliardini2012nonparametric}. Recently, a few papers have explored quantile policy learning. \citet{wang2018learning} investigates quantile-optimal treatment regimes; \citet{linn2017interactive} proposes quantile regression to indirectly and approximately optimize the quantile of outcomes within specific classes of decision rules; \cite{fang2023fairness} incorporates the quantile of the reward as a regularization term in their objective to maximize the average reward. However, there are no published papers studying quantile policy learning in the offline setting with unobserved confounding.
 
\paragraph{Causal inference}
A huge body of works in causal inference has focused on addressing confounding bias through (observed) covariate adjustment \citep{rubin1974estimating, rosenbaum1983central, lee2010improving, liu2024encoding}, instrumental variables \citep{angrist1996identification, ai2003efficient, newey2003instrumental, chen2003estimation, chernozhukov2008instrumental, chen2012estimation, baiocchi2014instrumental, hartford2017deep}, and negative controls \citep{miao2018identifying, kallus2021causal}. While the covariate adjustment approach cannot address the unobserved confounding, the IV and NC methods can account for the unobserved confounding. While earlier literature on IV and NC focuses on linear objective functions within parametric settings, \citet{chen2003estimation, chen2012estimation, chen2015sieve, chen2014local} and others have studied nonparametric quantile IV, and \citet{miao2018identifying, kallus2021causal} have considered nonparametric NC with linear objectives. In Theorem \ref{thm: NC conditional moment restrictions} of this paper, we extend the frameworks of \citet{chen2012estimation} and \citet{kallus2021causal} to establish nonlinear functional integral equations of the structural quantile function using NC. 
In addition, we incorporate the principle of pessimism to address the challenge of limited data coverage in offline quantile policy learning. There are some recent papers that combine pessimism with causal inference tools for confounded offline decision making under linear objective functions \citep{dong2023pasta, chen2023unified}. Our work extends that of \citet{chen2023unified} for maximizing an expected reward under pessimism to optimizing the quantile objective function under pessimism.

In summary, our paper aims at closing an important gap in the existing literature. By deriving integral equations that identify the structural quantile function using IV and NC, and by incorporating a principled pessimism step for offline policy learning with nonlinear objective functions, this work greatly expands the application capabilities of the current literature on offline policy learning.

\subsection{Notation}
We use upper-case letters to denote random variables and lower-case letters to denote the realizations of the corresponding random variables. For any set $\cB,$ we let $\Delta(\cB)$ denote the set of distributions over \(\mathcal{B}\). We let $p$ denote the density of random variables and $\prob$ denote the probability of an event or probability mass function of a discrete random variable. We use the notation $:=$ to indicate a definition or assignment. We use $\overset{\cE}{\leq}$ and $\overset{\cE}{\geq}$ to represent the inequality that holds on some event $\cE$. For any $n,$ we let $\cO(n)$ denote $Cn$ for some arbitrary positive constant $C$ and $\tilde{\cO}(n)$ to denote $\cO(n\cdot\text{poly}(\log n))$. 

 \subsection{Roadmap}
In Section \ref{sec: offline decision making}, we define the offline decision-making problem with a quantile objective when confounding bias is present. We also present several examples motivated from real-world applications. Sections \ref{sec: algo} and \ref{sec: theoretical results} focus on the IV approach to address unobserved confounding. In Section \ref{sec: algo}, we first explain the use of IV for the identification and estimation of the causal quantity of interest. We then present two algorithms for IV-assisted quantile-optimal policy learning. In Section \ref{sec: theoretical results}, we present a thorough theoretical analysis of the regret of our algorithm. Section \ref{sec: experiment} conducts simulation experiments to illustrate the performance of the algorithm. Section \ref{sec: conclusion} concludes the paper with a short discussion. In \S\ref{sec: other applications}, we present how our algorithm can be adapted to policy learning under other nonlinear objective functions. \S\ref{sec: negative controls} extends the IV results of Sections \ref{sec: algo} and \ref{sec: theoretical results} to those using the NC approach.

\section{Offline Decision Making}\label{sec: offline decision making}

In this section, we introduce the problem of quantile-optimal policy learning within an offline contextual bandit setting, with unmeasured confounders and observed auxiliary variables (IV or NC). We first present the general framework in detail. We then introduce a few real-world motivating examples.

\subsection{Offline decision making with a quantile objective}

In the following, we formulate the problem of offline decision making with a quantile objective under a contextual bandit framework. Compared with the standard contextual bandits, our setup features the existence of unmeasured confounders and auxiliary variables.

Let \(\mathcal{A}\) represent the space of actions and \(\Delta(\mathcal{A})\) denote the set of distributions over \(\mathcal{A}\). Let \(X \in \mathcal{X}\) be the context, \(Y \in \mathbb{R}\) the reward, and \(O\) the auxiliary variables. Let \(U \in \mathcal{U}\) denote the unmeasured confounder that causally affects $A$ and $Y$ simultaneously. 
The overarching learning problem is to learn a policy \(\pi: \cX\rightarrow\Delta(\cA)\) that maximizes the average structural quantile of the reward distribution based on an offline dataset, where the dataset does not include the unmeasured confounders $U$. 
Specifically, the decision-making process involves two steps: first, collecting an offline dataset from a given distribution, and then learning a policy from the offline data to apply in an interventional process.

\paragraph{The offline data collection process (ODCP)}
Suppose there is a joint distribution \(p(u, x, o, a, y)\) over the random variables \((U, X, O, A, Y) \), which 
 specifies the joint distribution of the observed data.  
 We note that the joint distribution can be factorized as
 \begin{align}\label{eq:odcp}
   \pr(u, x, o, a, y) = \pr(u, x, o) \cdot  \pr(a\given u, x, o) \cdot \pr(y\given u, x, o, a).
 \end{align}
 This factorization  illustrates the generating process of the offline data. In particular, \(\pr (u, x, o)\) is the joint distribution of the confounder $U$, context $X$, and auxiliary variables $O$. 
 The term \(\pr (a \given u, x, o)\) represents the behavior policy, which dictates how actions are selected based on \( (U, X, O) \). 
 The  reward $Y$ is generated conditioning on \( (U, X, O, A) \) via \(\pr (y \given u, x, o, a)\). 
Let $n$ denote the number of samples. 
Let  \(\{u_i, x_i, o_i, a_i, y_i\}_{i=1}^{n}\) denote $n$ i.i.d. observations from the joint distribution in \eqref{eq:odcp}. As $U$ is unmeasured, the offline dataset only includes \(\{x_i, a_i, o_i, y_i\}_{i=1}^{n}\). 

We note that, in the presence of unmeasured confounders, the auxiliary variables $O$ should be sufficiently informative for identifying the underlying causal structure. Thus, we need to pose additional assumptions to the joint distribution \(p(u, x, o, a, y)\). We will provide two examples of such assumptions in the subsequent subsection, where $O$ is an instrumental variable \citep{angrist1996identification} or a negative control exposure-outcome pair \citep{lipsitch2010negative}. 
 
\paragraph{The Interventional process} 
Suppose we learn a policy $\pie : \mathcal{X} \rightarrow \Delta(\mathcal{A})$ from the offline dataset,  we then apply this policy in an interventional process to assess its performance. 
In this process, we no longer observe the auxiliary variables. Moreover, the context $X$ may follow a different but known marginal distribution, denoted by $\widetilde p(x)$, than that in the ODCP. 
Upon observing a new context $X \sim \widetilde p(x)$ from the environment, we select an action 
$A \sim \pie(\cdot \given X)$ from the learned policy, which generates a reward $Y$ from the same confounded data generating process as in the ODCP. Specifically, let $ \pin{\pie}$ denote the joint distribution of random variables $(U, X, O, A, Y)$ in the interventional process under the policy $\pie$. We have the following factorization: 
\begin{align}\label{eq:interventional}
  \pin{\pie}(u, x, o, a, y) = \tpr(x) \cdot \pr(u, o \given x) \cdot \pie(a\given x)\cdot  \pr(y\given u, x, o, a).
\end{align} 
Here $\pr(y\given u, x, o, a)$ is the same as in \eqref{eq:odcp}. 
Comparing with the joint distribution in the ODCP, we note that the interventional process is a surgical intervention on the ODCP, where we let the action $A$ depend only on the context $X$. Additionally, the joint distribution of $(U, X, O)$ is allowed to be different. 
In the job training program example, the interventional process corresponds to the case where the government uses the learned policy to decide whether to allow a new worker to join the program based on their pre-intervention covariates.

\begin{figure}[h]
  \centering 
  \begin{subfigure}[b]{0.4\linewidth}
  \centering
  \begin{tikzpicture}
          \node[hiddenstate] (U) {$U$};
          \node[state] (X) [below=of U] {$X$};
          \node[state] (A) [left=of X, xshift=.27cm, yshift=-1cm] {$A$};
          \node[state] (Y) [right=of X, xshift=-.27cm, yshift=-1cm] {$Y$};
          \path[dashed] (U) edge (X);
          \path[normal] (U) edge (A);
          \path[normal] (X) edge (A);
          \path[normal] (A) edge (Y);
          \path[normal] (U) edge (Y);
          \path[normal] (X) edge (Y);
  
          \node[state] (Z) [left=of A] {$Z$};
          \path[normal] (Z) edge (A);
      \end{tikzpicture} 
  \caption{DAG of the ODCP for the IV. Note that $Z \indep Y \given (X, A, U ).$}\label{fig: IV ODCP DAG}
  \end{subfigure}\qquad
  \begin{subfigure}[b]{0.4\linewidth}
    \centering
      \begin{tikzpicture}
          \node[hiddenstate] (U) {$U$};
          \node[state] (X) [below=of U] {$X$};
          \node[state] (A) [left=of X, xshift=.27cm, yshift=-1cm] {$A$};
          \node[state] (Y) [right=of X, xshift=-.27cm, yshift=-1cm] {$Y$};
          \path[dashed] (U) edge (X);
          \path[normal] (X) edge (A);
          \path[normal] (A) edge (Y);
          \path[normal] (U) edge (Y);
          \path[normal] (X) edge (Y);
      \end{tikzpicture}
      \caption{DAG of the interventional process for the IV.}\label{fig: IV interventional DAG}
    \end{subfigure}\qquad
    \caption{(a) A DAG illustrating the causal relationship between random variables during ODCP when the IV is observed. The dashed edge implies that the causal relationship may be absent. The grey node indicates that $U$ is unmeasured. Here $Z$ is the auxiliary variables. Note that for $Z$ being a valid IV, $Z$ and $Y$ should be independent conditioning on $(X, A, U)$. (b) A DAG encoding the causal relationship between random variables in the interventional process. All arrows coming into the node $A$ have been removed other than the one from the node $X.$}\label{fig: IV DAG}
\end{figure}
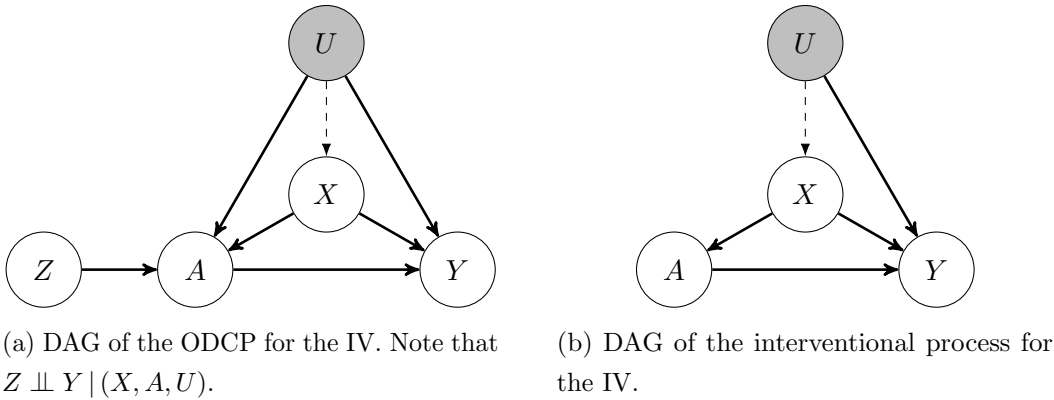

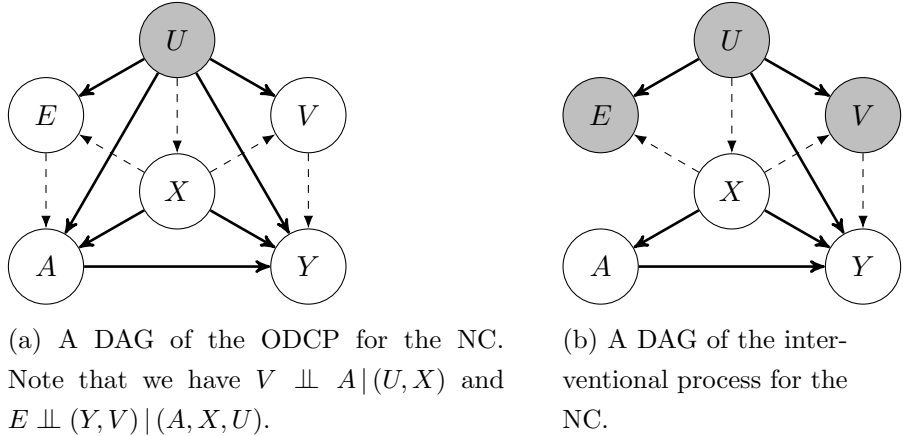
\begin{figure}[h]
  \centering 
  \begin{subfigure}[b]{0.4\linewidth}
  \begin{tikzpicture}
      \node[hiddenstate] (U) {$U$};
      \node[state] (X) [below=of U] {$X$};
      \node[state] (A) [left=of X, xshift=.27cm, yshift=-1cm] {$A$};
      \node[state] (Y) [right=of X, xshift=-.27cm, yshift=-1cm] {$Y$};
      
      \path[normal] (X) edge (A);
      \path[normal] (A) edge (Y);
      \path[normal] (U) edge (Y);
      \path[normal] (U) edge (A);
      \path[normal] (X) edge (Y);
      \path[dashed] (U) edge (X);

      \node[state] (E) [above=of A] {$E$};
      \node[state] (V) [above=of Y] {$V$};
      \path[normal] (U) edge (E);
      \path[normal] (U) edge (V);
      \path[dashed] (X) edge (V);
      \path[dashed] (X) edge (E);
      \path[dashed] (E) edge (A);
      \path[dashed] (V) edge (Y);
      \end{tikzpicture} 
  \caption{A DAG of the ODCP for the NC. Note that we have $V \indep A \given (U, X)$ and $E \indep (Y, V) \given (A, X, U).$}\label{fig: NC ODCP DAG}
  \end{subfigure}\qquad
  \begin{subfigure}[b]{0.23\linewidth}
  \centering
    \begin{tikzpicture}  
      \node[hiddenstate] (U) {$U$};
      \node[state] (X) [below=of U] {$X$};
      \node[state] (A) [left=of X, xshift=.27cm, yshift=-1cm] {$A$};
      \node[state] (Y) [right=of X, xshift=-.27cm, yshift=-1cm] {$Y$};
      
      \path[normal] (X) edge (A);
      \path[normal] (A) edge (Y);
      \path[normal] (U) edge (Y);
      \path[normal] (X) edge (Y);
      \path[dashed] (U) edge (X);

      \node[hiddenstate] (E) [above=of A] {$E$};
      \node[hiddenstate] (V) [above=of Y] {$V$};
      \path[normal] (U) edge (E);
      \path[normal] (U) edge (V);
      \path[dashed] (X) edge (V);
      \path[dashed] (X) edge (E);
      \path[dashed] (V) edge (Y);
  \end{tikzpicture}
  \caption{A DAG of the interventional process for the NC.}\label{fig: NC interventional DAG}
  \end{subfigure}\qquad
  \caption{(a) A DAG illustrating the causal relationship between random variables of the ODCP when the NCE and the NCO are observed. The dashed edge implies that the causal relationship may be absent. (b) A DAG encoding the causal relationship between random variables in the interventional process. All arrows coming into the node $A$ have been removed other than the one from the node $X.$ Note in addition to $U,$ $E$ and $V$ are also not observed, indicated by the grey nodes.}\label{fig: NC DAG}
\end{figure}

\paragraph{Causal graphs --- IV and NC} From a causal inference perspective, ODCP and the interventional process can be viewed as two different data generating processes, with two different causal graphs. And the interventional process is a surgical intervention on the ODCP, where we remove the arrows coming into the node of $A$ except the one that comes from the context $X$. See Figures \ref{fig: IV DAG} and \ref{fig: NC DAG} for illustrations of the causal graphs of the ODCPs and interventional processes, when $O$ corresponds to an IV $Z$ or a negative control exposure-outcome pair $(E, V)$. 
Informally, an IV is a random variable that affects the action $A$ but does not directly affect the reward $Y$. NC are a pair of random variables $(E, V)$, 
which are called negative control exposure (NCE) and negative control outcome (NCO) respectively.
Intuitively, $E$ is an random variable that does not causally affect the reward $Y$, and $V$ is a  random variable causally unaffected by either the action $A$ or the exposure $E$. We will focus on these causal assumptions in the rest of the paper. To be consistent with the notion in causal inference literature, in the sequel, we replace $O$ with $Z$ when it represents the IV \citep{angrist1996identification}, and by $(E, V)$ when it represents the negative control exposure-outcome pair \citep{lipsitch2010negative}.

\paragraph{Quantile objective and regret}
In the sequel, we denote the expectation over the distribution of random variables in ODCP as $\EE.$ 
For any fixed policy $\pie,$ we denote the expectation over the distribution of random variables in the interventional process as $\EE_{\pin{\pie}}.$ We aim to maximize the average structural quantile of the reward distribution. Let $\alpha \in (0,1)$ be a fixed scalar. Formally, given $X=x,$ for any action $a,$ we let $\hstaralpha(a, x)$ denote the $\alpha$-quantile of the potential outcome $Y(a)$, which is called the structural quantile function \citep{chernozhukov2008instrumental}. In the language of do-calculus \citep{pearl1995causal}, we have  
\begin{align}\label{eq:quantile_obj}
\prob[Y \leq \hstaralpha(A, X)\given X=x, \text{do}(A=a)]=\alpha
\end{align} under the interventional process $\pin{\pie}$. 
We assume that the potential outcome $Y(a)$ given $X$ is continuous; therefore, $\hstaralpha(A, X)$ is uniquely defined. Thus, mathematically, our goal is to learn a  policy $\pi: \cX\rightarrow\Delta(\cA)$ that maximizes the average structural quantile function: 
\begin{align}\label{def: v}
    v^\pie_{\alpha} := \EE_{\pin{\pie}}\sbr{\hstaralpha(A, X)},
\end{align}
where $X \sim \tilde p(x)$ and $A \sim \pie(\cdot \given X).$
Let $\piestar$ denote the optimal policy that maximizes the average structural quantile function. To measure the performance of a given policy $\pie$, we define the regret of $\pi$ as 
\begin{align}\label{def: SubOpt}
  \text{Regret}(\pie) = v^\piestar_{\alpha} - v^\pie_{\alpha}.
\end{align}
The regret is always nonnegative and characterizes the suboptimality of the policy $\pi$. 
Additionally, we would like to highlight that our objective $v^\pie_{\alpha}$ is a nonlinear functional of the distribution of $Y$, which poses significant challenges to the theoretical analysis.

\subsection{Motivating Examples}

In many real-world applications, the objective often shifts from maximizing the average outcome to optimizing a specific quantile of the outcome distribution. This shift is particularly relevant when the focus is on distributional fairness or minimizing financial risk. Our framework is designed to address these challenges by enabling quantile optimization, which offers a more equitable approach compared to traditional average-based methods.

\paragraph{Labor market economics} Consider the previously mentioned example of job training programs designed to improve the distribution of workers' incomes \citep{abadie2002instrumental, chernozhukov2008instrumental}. In this context, the auxiliary variable $Z_i$ serves as an IV, which could be whether the worker was invited to participate in the program---this affects their likelihood of participation but does not directly impact their wage. Governments often aim to ensure that the majority of workers experience income gains.
Since the average utility can be heavily influenced by a few high-income workers, the focus shifts to optimizing specific quantiles of the utility distribution, such as the median or lower quantiles. This approach helps to ensure that the program benefits a broader range of workers, particularly those at the lower end of the income distribution.
  
\paragraph{Financial portfolio optimization}

Consider the design of models for optimizing financial portfolios, where the goal is to enhance portfolio gains. In this setting, let \(Y_i\) represent the \emph{gain} (or return) generated by the \(i\)-th portfolio over a specific period. The context \(X_i\) includes pre-investment characteristics such as market conditions, existing portfolio allocations, and investor risk tolerance. The action \(A_i\) denotes the portfolio strategy implemented, for instance, the allocation weights across different asset classes or the choice of specific financial instruments. An IV \(Z_i\) might be a factor that influences the chosen strategy \(A_i\) but is assumed not to affect the portfolio's gains \(Y_i\) directly, except through the strategy itself; examples could include changes in brokerage fee structures or certain types of financial advice that nudge strategy selection without containing direct information about future returns.

Financial institutions often aim for robust performance beyond simply maximizing expected gains. The distribution of portfolio gains can be heavily skewed, particularly by infrequent but large positive or negative events, making the mean gain an unreliable measure of typical performance or risk-adjusted return. Therefore, decision-makers may focus on optimizing specific quantiles of the gain distribution. For instance, they might aim to maximize a lower quantile (e.g., the 5th or 10th percentile gain) to ensure a degree of capital preservation or a minimum performance level, effectively managing downside risk. Alternatively, they might maximize the median gain for a robust measure of central tendency, or a higher quantile (e.g., the 75th percentile) to target ambitious but achievable upside potential.

A key challenge is the potential presence of unmeasured confounders \(U_i\) that could affect both the chosen portfolio strategy \(A_i\) and the realized gain \(Y_i\). Examples include evolving investor sentiment or private information, which may simultaneously influence asset allocation decisions and market performance, or unobserved market liquidity shocks that drive both strategic adjustments and gain outcomes. Furthermore, adapting concepts like spectral risk measures \citep{dowd2006after} to spectral utility measures, which integrate gains over various quantiles using preference-based weighting functions, is also relevant. In Example \ref{ex: quantile-based risk measures}, we demonstrate how our method can be adapted to minimize this measure.

\paragraph{Public health policy} Consider the scenario of reducing extreme hospital readmission times for patients post-surgery. Here, \(Y_i\) represents the post-surgery readmission time for the \(i\)-th patient, where lower values are better. \(A_i\) represents the type of post-surgery care provided, such as intensive monitoring versus a routine discharge process. \(X_i\) is a set of pre-surgery covariates of the patient (e.g., age, comorbidities).
To handle unmeasured confounders \(U_i\) (e.g., unrecorded severity of the patient's underlying condition that affects both the choice of post-surgery care and readmission risk), NC might be used. For example, 
\begin{itemize}
    \item The Negative Control Exposure (NCE) \(E_i\) could be the primary language spoken by the patient's initial care team. It influences communication and thus the nuanced details of care \(A_i\), but is not believed to directly affect the biological process of readmission \(Y_i\), conditional on \(A_i, X_i, U_i\).
    \item The Negative Control Outcome (NCO) \(V_i\) could be a measure of patient satisfaction with hospital administrative processes during admission for the surgery. It is affected by general hospital quality \(U_i\), which also influences readmission time \(Y_i\), but is not affected by the specific post-surgery care \(A_i\) or the NCE \(E_i\).
\end{itemize}
In this context, the goal of the health system is often not just to reduce the average readmission time but to specifically mitigate the risk of very long readmission times (extreme cases), as these often indicate severe adverse events or poor post-surgical outcomes. Thus, the focus is on policy learning that optimizes an upper quantile of the readmission time distribution (e.g., minimizing the 90th or 95th percentile of readmission times).

\section{Causal-Assisted Pessimistic Policy Learning Algorithms}\label{sec: algo}

The offline quantile policy learning problem introduced in Section \ref{sec: offline decision making} involves three challenges: (i) the inherent nonlinearity of the quantile objective as a functional of the reward distribution; (ii) the presence of unmeasured confounders that bias naive estimation; and (iii) insufficient coverage in the offline dataset.  
To surmount these obstacles, we develop a suite of algorithms founded on causal inference principles and the pessimism paradigm. 
Specifically, to address challenges (i) and (ii), we leverage causal inference tools such as instrumental variables (IV) and negative controls (NC) to identify the structural quantile function nonparametrically, which can be solved using a minimax estimation approach to \emph{nonlinear functional integral equations}. This tackles the complication stemming from the nonlinear quantile objective with confounders. Furthermore, to handle challenge (iii), we incorporate the \emph{principle of pessimism}, which leads to the construction of conservative policy estimates. This enables us to focus on policies whose performance can be reliably evaluated with the available offline data.

For clarity, we will first detail these methodologies in the context of IV. The framework for NC, which shares a similar conceptual basis, will be discussed in \S\ref{sec: negative controls}.

\subsection{Causal Identification via Instrumental Variables}\label{sec: model}

Optimizing the average structural quantile function \( \EE_{\pin{\pie}}\sbr{\hstaralpha(A, X)} \) requires an initial, reliable estimate of the structural quantile function \(\hstaralpha(A, X)\) itself. 
Recall that $\hstaralpha(A, X)$ is defined in \eqref{eq:quantile_obj}. 
A naive quantile regression of \(Y\) on \(A\) and \(X\) is inadequate due to endogeneity introduced by unmeasured confounders \(U\) \citep{chernozhukov2008instrumental}, which leads to a confounding bias.

To address the confounding bias, we propose to use the IV $Z$ to identify \(\hstaralpha(A, X)\). 
A standard nonparametric IV model requires the IV $Z$ to satisfy three conditions: (i) relevance: the distribution of the action is not constant in the IV, (ii) exclusion restriction: the IV affect the reward only through action, and (iii) unconfounded instrument: the IV are conditionally independent of the error term \citep{newey2003instrumental,hartford2017deep}. The general NPQIV model, however, necessitates a more stringent set of assumptions \citep{chen2014local}. Following the treatment of Section 6 in \cite{chen2012estimation}, we employ a simplified version of the NPQIV model incorporating the additive error assumption and conditional moment restriction:
\begin{assumption}[Model Assumption for Quantile IV]\label{asp: model}
    We assume that the following conditions hold for the IV model in the ODCP:
    \begin{align} \label{eq:model_assumption}
        Y=\hstaralpha(A, X)+\epsilon \qquad \text{ and }\qquad \prob(\epsilon \le 0 \given X, Z)=\alpha.
    \end{align}
Here, \(\epsilon\) is the structural error, dependent on \(U\), and its \(\alpha\)-quantile is zero conditional on \(X\) and \(Z\), but not necessarily conditional on \(X\) and \(A\).
    \end{assumption}
 
By \citet{chen2014local}, with some regularity conditions, Assumption \ref{asp: model} holds by the underlying causal structure of the ODCP in the IV case, as depicted in Figure \ref{fig: IV DAG}-(a). In contrast to a standard regression model where the mean of the error term is assumed to be zero, we define $\epsilon$ as a  structural error that adheres to a distinct conditional distribution assumption and is responsible for the confounding effect induced by the unmeasured confounders $U$. 
Consequently, $\epsilon$ is not independent of $(A, X)$ and the condition $\prob(\epsilon \le 0 \given X, A) = \alpha$ does not hold. This subtlety renders the conventional quantile regression of $Y$ on $A$ and $X$ erroneous. Fortunately, the existence of the IV allows us to write Assumption \ref{asp: model} as
\begin{align*}
\EE \sbr{\ind \{Y \leq \hstaralpha (A, X)\}\given X, Z}=\alpha.
\end{align*}
If we write $\ind \{Y \leq \hstaralpha(A, X)\} - \alpha$ as $W(D; \hstaralpha)$ where $D=(Y, X, A),$ we would then establish the relationship: $\EE \sbr{W(D; \hstaralpha)\given X, Z}=0.$ 
That is, $\hstaralpha$ is the solution to the following conditional moment restriction:
\begin{align}\label{def: conditional moment equation}
    \EE \sbr{W(D; \vh)\given X, Z}=0
\end{align} with respect to $\vh$. 
If $\hstaralpha$ further is the unique solution to \eqref{def: conditional moment equation}, we can estimate it by solving the conditional moment restriction in \eqref{def: conditional moment equation} based on the offline data. 

We note that $W(D, \vh)$ is a nonlinear function of $\vh$ and the conditional moment restriction in \eqref{def: conditional moment equation} is a nonlinear functional integral equation. 
For any function $\vh$, we define a nonlinear operator $\vT^{\textrm{IV}}$ that maps from a function of $(A, X)$ to a function of $(X, Z)$ as follows:
\begin{align*}
    (\vT^{\textrm{IV}} \vh )(x,z) :=\EE\sbr{W(D; \vh)\given (X, Z)=(x,z)}.
\end{align*}
Then we have $\vT^{\textrm{IV}} \hstaralpha = 0$, i.e., a zero function, almost surely. 
The equation of the form $\cT^{\textrm{IV}} h =0$ is referred to as conditional moment restriction in econometrics \citep{chamberlain1992efficiency, ai2003efficient, newey2003instrumental, chen2009nonlinear}. 
Under the identifiability and realizability assumption (i.e., Assumption \ref{asp: identifiability and realizability}) introduced later, this equation identifies the structural quantile function $\hstaralpha$.

As we will show in Theorem \ref{thm: NC conditional moment restrictions}, we can also use NC to identify $\hstaralpha$ under a similar set of assumptions, where we introduce a similar nonlinear operator $\cT^{\textrm{NC}}$. 
Consequently, estimating $\hstaralpha$ reduces to solving the conditional moment restrictions $\cT^{\text{IV}} h = 0$ or $\cT^{\text{NC}} h = 0$.  The only distinction between IV and NC is that $\cT^{\text{NC}}$ requires solving two conditional moment restrictions simultaneously, while $\cT^{\text{IV}}$ involves solving one. This difference is minimal, allowing us to use a similar approach for both cases. In fact, our methodology applies to any case where $\hstaralpha$ can be identified by solving a conditional moment restriction of the form $\cT h = 0$ for some nonlinear operator $\cT$.
In the sequel, to highlight such generality, we use $\cT$ to denote $\cT^{\textrm{IV}}$ for the IV case.

\subsection{Minimax Estimation}
The conditional moment restriction in  \eqref{def: conditional moment equation} is essentially a nonparametric inverse problem for $\vh$. It is impossible to find a closed-form solution. Various methods have been proposed to estimate the structural quantile function based on the conditional moment restriction. We adopt a minimax estimation strategy, inspired by \citet{dikkala2020minimax}, to estimate \(\hstaralpha(A,X)\). Our approach transforms the moment restrictions into an unconditional loss function via Fenchel duality and estimates $\hstaralpha$ by minimizing the loss function. 

We first define a conditional residual mean squared error (RMSE) with respect to $\vh$ as 
\begin{align}\label{def: RMSE}
    \nbr{\vT\vh}^2_{2}:= \EE\sbr{\rbr{\EE \sbr{W(D; h)\given X, Z}}^2}.  
\end{align}
Note that  for any function $h$ we have 
\begin{align*}
    \nbr{\vT\vh}^2_{2} \geq 0, \text{~~ and~~} \nbr{\vT\vh}^2_{2}=0 \text{ if and only if } \EE \sbr{W(D; h)\given X, Z}=0. 
\end{align*}  
Therefore, we can construct an estimator $\vh\in\vH$ of $\hstaralpha$ by minimizing \eqref{def: RMSE} in a suitable hypothesis function space $\vH$. However, $\eqref{def: RMSE}$ is a squared conditional expectation. Estimating it using the squared empirical expectation introduces variance terms, leading to a bias. To address this, we first utilize the Fenchel duality of the function $x^2/2$ and reformulate $1/2 \cdot \nbr{\vT\vh}^2_{2}$ as

\begin{align}\label{def: loss function}
    \frac{1}{2}\nbr{\vT\vh}^2_{2} &= \EE\sbr{\sup_{\theta \in \Theta}\vT\vh(\vX, \vZ) \cdot \theta(\vX, \vZ)- \frac{1}{2} \cdot \abr{\theta (X,Z)}^2 }\nend
    &= \sup_{\theta \in \Theta} \cbr{\EE\sbr{W(D; h) \cdot  \theta(\vX, \vZ) - \frac{1}{2} \cdot \abr{\theta (X,Z)}^2 } }.
\end{align}
Here $\Theta$ is a function class over $\cX \times \cZ$ that contains the function $\vT\vh$, and the last equality holds by the interchangeability principle \citep{dai2017learning}. Therefore, we replace the problem of minimizing \eqref{def: RMSE} by minimizing \eqref{def: loss function}. 
Note that the expectation in \eqref{def: loss function} is taken with respect to the joint distribution of $(Y, A, X, Z)$ and can be unbiasedly estimated from data. 
Let $\cL_{n}(\vh)$ denote the empirical version of \eqref{def: loss function}, i.e., 
\begin{align}\label{def: empirical loss function}
    \cL_{n}(\vh)&:=\sup_{\theta\in\Theta}\cbr{\EE_n\sbr{W(D; h) \cdot \theta(\vX, \vZ)}-\frac 1 2 \norm{\theta}_{n, 2}^2}\nend
    &= \sup_{\theta\in\Theta} \cbr{\frac{1}{n}\sum_{i=1}^{n}\sbr{W(D_i; h)\cdot \theta(\vX_i, \vZ_i)} - \frac{1}{2n}\sum_{i=1}^{n}\theta^2(\vX_i, \vZ_i)},
\end{align}
where we introduce a real-valued test function class $\Theta$ on $\cX \times \cZ$ such that $\vT\vh$ can be well approximated by some function in $\Theta$. As $\cL_{n}(\vh)$ can now be computed from the offline data, we can then estimate $\hstaralpha$ by minimizing $\cL_{n}(\vh)$ with respect to $\vh \in \mathcal{H}$.

\subsection{Two Pessimistic Algorithms for Policy Learning}\label{sec: algorithm}

Based on the loss function $\cL_{n}(\vh)$ defined in \eqref{def: empirical loss function}, we can construct an estimator of $\hstaralpha$  and use it to learn the optimal policy $\piestar$ in the interventional process. We introduce two versions of algorithms based on the pessimism principle. 


\paragraph{Challenge of insufficient data coverage.}
A major challenge of offline decision-making is to deal with the distribution shift between the ODCP that generates the observed action and the oracle policy. To see this, given $\vh$ and $\pie,$ we denote $v(h, \pie) :=\EE_{\pin{\pie}}\sbr{h(A, X)}.$ Suppose we have already obtained an estimator $\hat{h}$ of $\hstaralpha$ from the offline data, e.g., by minimizing $\cL_{n}(\vh)$ in \eqref{def: empirical loss function}. We can then construct a policy by greedily maximizing $v (\hat{h}, \pie),$ which gives $\tilde{\pie} := \argsup_{\pie}v ( \hat{h}, \pie).$ 

To understand the performance of $\tilde{\pie}$, we can decompose its regret as
\begin{align} \label{eq: suboptimality decomposition}
    \text{Regret}(\tilde{\pie}) &=v^{\piestar}_{\alpha}-v^{\tilde{\pie}}_{\alpha} \nend
    &= \underbrace{v^{\piestar}_{\alpha} - v ( \hat{h}, \piestar) }_{\displaystyle \text{(i)}} + \underbrace{v ( \hat{h}, \piestar) - v (\hat{h}, \tilde{\pie}) }_{\displaystyle \text{(ii)}} + \underbrace{v ( \hat{h}, \tilde{\pie})  - v^{\tilde{\pie}}_{\alpha}}_{\displaystyle \text{(iii)}}. 
\end{align}
Term (i) only depends on the oracle policy $\piestar$ and the estimator $\hat{h}$. It captures the distribution shift between the interventional process and the ODCP in terms of $X$. When such a distribution shift is mild and $\hat{h}$ is a good estimator of $\hstaralpha$, this term is small. 
Term (ii) is upper bounded by zero by the optimality of $\tilde{\pie}$. 
Term (iii) presents a unique subtlety: $\tilde{\pie}$ and $\hat{h}$ are both derived from offline data, creating spurious correlation. 
That is, suppose $\hat h$ has a high estimation uncertainty for some suboptimal context-action pair $(x, a)$, causing $\tilde {\pie}$ to choose this suboptimal action at context $x$. Term (iii) will be large. In other words, because $\tilde{\pie}$ is learned from data and is inherently random, to make Term (iii) small, we need to ensure that $\hat{h}$ is a good estimator of $\hstaralpha$ for all context-action pairs. Such a requirement is often too strong, especially when some parts of the context-action space are underexplored in the offline data.

\paragraph{The Solution set algorithm.} To tackle this challenge, we learn a pessimistic policy by doing uncertainty quantification on $\cL_{n}(\vh)$ \citep{yu2020mopo, kumar2020conservative, kidambi2020morel, jin2021pessimism, xie2021bellman, rashidinejad2021bridging, buckman2020importance}. Specifically, we first construct a solution set $\cS(e_n)$ for $\vh$ based on $\cL_{n}(\vh)$ as: 
\begin{align}\label{def: CI}
    \cS(e_n) := \cbr{\vh\in\vH:  \vcL_{n}(\vh) \le  \inf_{\vh\in\vH}\vcL_{n}(\vh) + e_n},
\end{align}
where $e_n$ is a small positive threshold we will determine later. We will show in Theorem \ref{thm: uncertainty quantification} (i) that, with high probability, $\hstaralpha$ lies in $\cS(e_n)$ by choosing $e_n$ properly. 
Then, based on $ \cS(e_n)$, we select the policy that optimizes the pessimistic average reward function: 
\begin{align}\label{eq: pessimistic estimator}
    \piepessi := \arg\sup_{\pie} \inf_{h\in\cS(e_n)} v(h, \pie).
\end{align}
Here \eqref{eq: pessimistic estimator} yields a pessimistic solution because 
$ \inf_{h\in\cS(e_n)} v(h, \pie) \leq v(\hstaralpha, \pie) = v_{\alpha}^{\pi}$ for any $\pie$ and $h\in\cS(e_n)$,
when $\hstaralpha\in\cS(e_n)$, which we will later show holds with high probability.
In other words, $\piepessi$ is the policy that maximizes a pessimistic estimate of the objective function.

We state the details of this algorithm in  Algorithm \ref{alg: meta}. To summarize, our algorithm consists of three steps.
First, we reduce the problem of estimating $\hstaralpha$ to the problem of conditional moment restriction  in \eqref{def: conditional moment equation}, which is solved by minimizing the loss function $\vcL_{n}(\vh)$ in \eqref{def: loss function}. 
Second, we construct a solution set \eqref{def: CI} based on the sublevel sets of the loss function in \eqref{def: empirical loss function}. 
Finally, we construct a policy $\piepessi$ from the solution set \eqref{def: CI} by solving the optimization problem in \eqref{eq: pessimistic estimator}.
  
\begin{algorithm}
\caption{Pessimistic Policy Learning Algorithm with Solution Sets}
\small
\begin{algorithmic}\label{alg: meta}
\REQUIRE Offline dataset $\cbr{a_i, x_i, z_i, y_i}_{i=1}^{n}$, hypothesis space $\vH$, test function space $\Theta$ and threshold $e_n$.
\STATE (i) Construct solution set $\cS(e_n)$ as the sublevel set of $\vH$ with respect to metric $\cL_n(\cdot)$ and threshold $e_n$.
\STATE (ii) $\piepessi :=\arg\sup_{\pie}\inf_{h\in \cS(e_n)} v(h, \pie)$.
\ENSURE $\piepessi$.
\end{algorithmic}
\end{algorithm}

\paragraph{Benefit of pessimism.} To see the benefit of the  pessimism principle, recall the regret decomposition in \eqref{eq: suboptimality decomposition} for $ \piepessi$ constructed in \eqref{eq: pessimistic estimator}. 
We let $\hat{h} := \arg\inf_{h\in\cS(e_n)} v(h, \piepessi)$. 
Then, by \eqref{eq: suboptimality decomposition}, we have 
$$
v ( \hat{h}, \piestar)  - v ( \hat{h}, \piepessi ) \leq 0 , \qquad   v ( \hat{h},  \piepessi)  - v^{ \piepessi}_{\alpha} = v ( \hat{h},  \piepessi)  -  v ( \hstaralpha,  \piepessi)  \leq 0. 
$$
Here, the first inequality is due to the optimality of $\piepessi$ in \eqref{eq: pessimistic estimator}, and the second inequality is due to the fact that $\hstaralpha\in\cS(e_n)$ by our choice of $e_n$ (see Theorem \ref{thm: uncertainty quantification} (i)).
As a result, we have 
\begin{align}
\label{eq:pess_upperbound}
\text{Regret}(\piepessi) \leq   v^{\piestar}_{\alpha} - v ( \hat{h}, \piestar)  = v\rbr{\hstaralpha,  \piestar} - v ( \hat{h},  \piestar) .  
\end{align}
Therefore, we reduce the regret of the pessimistic policy $\piepessi$ to a term that essentially measures the error of the estimator $\hat{h}$ in the interventional process.
In particular, this error is evaluated on the distribution induced by $\piestar$, which is the oracle optimal policy. 
In other words, as long as the offline data is sufficiently explorative to cover the support of the oracle policy, we can expect the regret of the pessimistic policy $\piepessi$ to be small.

\paragraph{The Regularized algorithm.} As an intermediate step to learn $\piepessi,$ we need to minimize $v(h, \pie)$ for $\vh$ restricted to $\cS(e_n).$ This is an optimization problem with a data-dependent constraint, which is often computationally intractable in practice. 
We now introduce a computationally benign version of the policy learning algorithm. 
Let $\bigeps(h)$ denote $\vcL_{n}(\vh) -  \inf_{\vh\in\vH}\vcL_{n}(\vh).$ We modify the objective function in \eqref{eq: pessimistic estimator} by adding a regularization term $\lambda_n \cdot \bigeps(h).$ The regularized version of the pessimistic policy is then defined as:
\begin{align}\label{eq:pess_reg}
    \piepessi_{R}:=\arg\sup_{\pie}\inf_{h\in \vH} \{v(h, \pie) + \lambda_n \cdot \bigeps(h)\}.
\end{align}
This approach reformulates the data-dependent constrained optimization problem in \eqref{eq: pessimistic estimator} into its augmented Lagrangian counterpart. The idea is motivated from \cite{rashidinejad2022optimal}, which studies offline bandit and reinforcement learning without confounders. The additional regularization term $\lambda_n \cdot \bigeps(\vh)$ can be viewed as the uncertainty of $\vh$. Thus, by adding this term, we effectively penalize the estimator $\vh$ based on its uncertainty.
Furthermore, when the estimator $\vh$ closely approximates $\hstaralpha$, the regularization term approaches $\lambda_n \cdot \bigeps(\hstaralpha)$, thereby becoming a small quantity. Consequently, we expect less bias when optimizing the augmented Lagrangian over $\pie$. 
Thus, the regularization term does not induce significant bias to the optimization problem, while it can significantly penalize the uncertainty of the estimator $\vh$. 
We summarise the regularized version of the algorithm in Algorithm \ref{alg: regularized version}.

Comparing \eqref{eq:pess_reg} with \eqref{eq: pessimistic estimator}, we see that the regularized version of the pessimistic policy $\piepessi_{R}$ is obtained from an unconstrained optimization problem. 
In practice, this is often more amenable to gradient-based optimization.
For example, \cite{xie2021bellman} uses the mirror descent method to optimize the regularized version of the pessimistic policy. We remark that while \cite{rashidinejad2022optimal} and \cite{xie2021bellman} advocate for the utilization of a regularized objective within their respective algorithms, these are explicitly tailored for the domains of offline bandit and offline reinforcement learning without addressing the issue of confounding effect. In contrast, our proposed algorithm is designed to remove the confounding bias, a pervasive challenge of causal inference under the setting of offline decision making.

\begin{algorithm}
    \caption{Pessimistic Policy Learning Algorithm with Regularization}
    \small
    \begin{algorithmic}\label{alg: regularized version}
    \REQUIRE Offline dataset $\cbr{a_i, x_i, z_i, y_i}_{i=1}^{n}$ from the ODCP, hypothesis space $\vH$, test function space $\Theta$ and regularization parameter $\lambda_n$.
    \STATE $\piepessi_{R}:=\arg\sup_{\pie}\inf_{h\in \vH} \{v(h, \pie) + \lambda_n \cdot \bigeps(h)\}$.
    \ENSURE $\piepessi_R$.
    \end{algorithmic}
    \label{algo:regularized}
    \end{algorithm}


\section{Theoretical Results}\label{sec: theoretical results}
In this section, we establish regret guarantees for the pessimistic policy learning algorithms established in \eqref{eq: pessimistic estimator} and \eqref{eq:pess_reg}. 
The theoretical analysis for these algorithms, derived for learning the quantile-optimal policy, is nontrivial and distinct from the existing literature on learning the average-optimal policy for two main reasons.  First, the standard concentration inequalities used in the existing literature do not work for $\nbr{\cT\vh}_{2}$ because of the nonlinearity of the operator $\cT$. Second, it is difficult to bound the regret of $\piepessi$ or $\piepessi_R$  in terms of $\nbr{\cT\vh}_{2}$.

To address the first challenge, we use bracketing number techniques from \citet{geer2000empirical, chen2003estimation}. 
Addressing the second challenge requires us to link the estimation error of the estimated structural quantile function $\hat h$ to $ \| \cT\hat h\| _{2}$. 
That is, we aim to link $\| \hat h - \hstaralpha \|_{\bullet} $ to $\| \cT\hat h\| _{2}$, where $\| \cdot \|_{\bullet}$ is some norm. 
To see this, consider the estimator $\hat \pie $ defined in \eqref{eq: pessimistic estimator}. 
When the solution set $\mathcal{S}(e_n)$ contains $\hstaralpha$ and the offline data is sufficiently regular, by the regret decomposition in \eqref{eq: suboptimality decomposition}, we can bound $\mathrm{Regret}(\hat \pie )$ by $\| \hat h - \hstaralpha \|_2 $. 
However, we cannot directly bound $ \| \hat h - \hstaralpha\| _{2} $ by analyzing the loss function $\cL_{n}(\vh)$ in \eqref{def: empirical loss function} due to the nonlinearity of the operator $\cT$.
To bypass this issue, we perform a local expansion of $\cT\hat h$ around $\hstaralpha$ and define a pseudo-metric $\norm{\cdot}_{\mathrm{ps}}$ on the solution set $\cS(e_n)$. This pseudo-metric acts as a bridge that connects the regret of $\hat \pie$ to $\| \cT\hat h \| _{2}$, which can be bounded using concentration tools. 
As we show below, although we are focused on the instrumental variable setting, every theorem presented has an equivalent counterpart for negative controls, as detailed in \S\ref{app: theoretical analysis NC}. Moreover, the analysis can be extended to general settings where $\hstaralpha$ is the solution to a nonlinear conditional moment restriction. 

In the following, we first present the theoretical guarantees for the solution set algorithm, which yields the pessimistic estimator $\piepessi$ defined in \eqref{eq: pessimistic estimator}. We will focus on the theory of the regularized algorithm in Section \ref{sec: theoretical analysis for the regularized set version algorithm}. We introduce several technical assumptions in the sequel and provide a detailed discussion of some of them—specifically, Assumptions \ref{asp: identifiability and realizability}, \ref{asp: regularity of Density}, \ref{asp: local curvature}, \ref{asp: data coverage}, and Condition \ref{con: high probability event}---in \S\ref{app: discussion assumptions}.

\subsection{Analysis of Solution Set $\cS (e_n)$} \label{sec: analysis of solution set}

We first establish theoretical guarantees for the solution set $\cS (e_n)$ constructed in \eqref{def: CI}. We will prove that, under proper conditions, $\cS (e_n)$ contains the true structural quantile function $\hstaralpha$ with high probability. 
Moreover, for any function $h$ in $\cS (e_n)$, we establish a uniform upper bound on the RMSE $\nbr{\cT\vh}_{2}$ defined in \eqref{def: RMSE}. To begin with, we impose a few conditions on the hypothesis and test function spaces. Note that a function class $\sF$ is star-shaped if for every $f \in \sF$ and  $r \in [0, 1],$ we have $rf \in \sF.$ 

\begin{assumption}  [Identifiability and Realizability]\label{asp: identifiability and realizability}
    We assume $\hstaralpha \in \vH$ and $\EE \sbr{W(D; \hstaralpha)\given X, Z}=0.$ Moreover, for any $\vh \in \vH$ with $\EE \sbr{W(D; \vh)\given X, Z}=0,$ we have $\nbr{\vh - \hstaralpha}_{\infty}=0.$
\end{assumption}

\begin{assumption}[Compatibility of Test Function Class]\label{asp: compatibility of test function class}
    We assume that $\Theta$ is sufficiently large such that, 
for any $\vh\in\vH,$ $\inf_{\theta\in\Theta} \nbr{\theta - \cT\vh}_{2}=\epsilon_{\Theta}$ with   $\epsilon_{\Theta}=\cO(n^{-1/2})$. 

\end{assumption}

\begin{assumption}[Regularity of Function Classes]\label{asp: regularity of function classes}
 We assume $\vH$ is compact with respect to the supremum norm $\nbr{\cdot}_{\infty} $ and $\Theta$ is star-shaped. We also assume the support of $\hstaralpha(A, X)$ is bounded, i.e., $\nbr{(\hstaralpha(A, X))}_\infty\le L_Y$. Moreover, it holds that $\sup_{h\in\vH}\nbr{h}_\infty\le L_{h}$ and $\sup_{\theta\in\Theta}\nbr{\theta}_\infty\le L_{\theta}$.
\end{assumption}
Assumption \ref{asp: identifiability and realizability} states that the conditional moment restriction identifies $\hstaralpha$ and that the hypothesis space $\vH$ captures $\hstaralpha.$ Moreover, $\hstaralpha$ is the unique solution to the conditional moment restriction. This set of assumptions is typically required in the econometric literature \citep{chen2012estimation}. Assumption \ref{asp: compatibility of test function class} ensures that the test function class is rich enough to approximate $\cT\vh$ for all $\vh \in \cH.$ Assumption \ref{asp: regularity of function classes} can be easily satisfied by choosing $\cH$ and $\Theta$ to be standard uniformly bounded and closed function classes.

By assuming $\Theta$ is star-shaped, $\cL_{n}(\cdot)$ becomes a nonnegative function as $\Theta$ contains the zero function.
Due to the form of $\vcL_{n}(\cdot ),$ it is natural to consider the function class
\begin{align*}
    \cQ  =\cbr{W(\cdot  ; h)\cdot \theta(\cdot ): \vh\in\vH, \theta\in\Theta},
\end{align*}
which is parameterized by $(\vh, \theta)$. 
Each element in $\cQ$ is a function of $ (Y, X, A, Z)$. 
Since $\vcL_{n}(\vh)$ is represented as the supremum of the addition of two terms, we define an event that quantifies the approximation error between each term in $\vcL_{n}(\vh)$ and its population counterpart. This event will later be proved to have a high probability by using empirical process theory. Fix a sequence of small positive constants $\eta_n$ that decreases with $n.$ Let $\cE$ denote the event
\begin{align}\label{def:Event}
\cE \;=\;
\Bigl\{
    \;\bigl|
        \EE_n[q(D)] - \EE[q(D)]
    \bigr|
    \;\le\;
    \eta_n\Bigl(\|\theta\|_{2}+\eta_n\Bigr),
    \; \text{and}\;
    \bigl|
        \|\theta\|_{n,2}^{2} - \|\theta\|_{2}^{2}
    \bigr|
    \;\le\;
    \tfrac12\Bigl(\|\theta\|_{2}^{2}+\eta_n^{2}\Bigr),
    \;\forall\,q\in\cQ
\Bigr\}.
\end{align}

We impose a condition that ensures $\cE$ holds with high probability, which can be verified with concrete instantiations of $\vH$ and $\Theta$ in \S\ref{app: bracketing concentration}.

\begin{condition}\label{con: high probability event}
    Suppose that Assumption \ref{asp: regularity of function classes} holds. For any $\xi>0,$ there exists $\eta_n > 0$  such that the event $\cE$ holds with probability at least $1-2\xi.$
\end{condition}
The set $\cE$ encapsulates two events that characterize the concentration of the function classes $\cQ$ and $\Theta$ separately. The term $\eta_n$ can be interpreted as the convergence rate of the tail bounds associated with the concentration of $\cQ$ and $\Theta$. In cases where $W(\cdot; h)$ is smooth in $\vh$, existing literature on the minimax estimation approach \citep{dikkala2020minimax, uehara2021finite, miao2023personalized} leverages the critical radius of the localized Rademacher complexity of the function classes as developed in \citet{wainwright2019high} to demonstrate that $\eta_n = \tilde{\cO}(n^{-1/2})$. Nevertheless, given that $W(\cdot; h)$ in our context includes an indicator function parametrized by $\vh$, this method is no longer applicable. To achieve a fast convergence rate, we borrow the tools from \citet{geer2000empirical}, which measure the complexity of the function class $\cQ$ and obtain the concentration bound via the bracketing number. For the detailed computation, we direct the reader to  \S\ref{app: bracketing concentration}, where we establish that Condition \ref{con: high probability event} is satisfied for some $\eta_n = \tilde{\cO}(n^{-1/2})$ given that $\cH$ and $\Theta$ are suitably selected. Consequently, Assumption \ref{asp: compatibility of test function class} can be stated as $\inf_{\theta\in\Theta} \nbr{\theta - \cT\vh}_{2} = \cO(\eta_n)$. 

The theoretical results for both the solution set and the regularized algorithms are built on the event $\cE$. So for the rest of the paper, we assume Condition \ref{con: high probability event} holds.  
In the following theorem, we prove that, on event $\cE$,  the solution set $\cS(e_n)$ exhibits some favorable properties.
\begin{theorem}[Uncertainty Quantification]\label{thm: uncertainty quantification}
Suppose that Assumptions \ref{asp: identifiability and realizability}, \ref{asp: compatibility of test function class}, and \ref{asp: regularity of function classes} hold. 
\begin{itemize}
    \item[(i).] On event  $\cE$, $\cL_{n}(\hstaralpha) \le 13/4 \cdot \eta_n^2.$ Moreover, if we set $e_n> 13/4 \cdot \eta_n^2$, then $\hstaralpha\in\cS(e_n)$.
    \item[(ii).] On event $\cE$, for all $\vh\in\cS(e_n)$, we have,
\begin{align*}
    \nbr{\cT\vh}_{2} = \cO\rbr{\sqrt{e_n}} +  \cO\rbr{\eta_n}.
\end{align*}
\end{itemize}
\begin{proof}
See \S\ref{pro: uncertainty quantification} for a detailed proof.
\end{proof}
\end{theorem}
Theorem \ref{thm: uncertainty quantification} (i) states that when $e_n$ is chosen properly, the solution set $\cS(e_n)$ captures $\hstaralpha$ with high probability. This fact is indispensable for establishing pessimism, which further leads to the regret upper bound in \eqref{eq:pess_upperbound}. 
Theorem \ref{thm: uncertainty quantification} (ii) characterizes the rate of conditional RMSE for all $\vh$ in $\cS(e_n)$ uniformly. 
As we will show subsequently, this result is crucial for establishing the final regret upper bound of the pessimistic algorithms. 

In the following corollary, we apply Theorem \ref{thm: uncertainty quantification} (i) to the regret decomposition in \eqref{eq: suboptimality decomposition}  and establish an upper bound of $\mathrm{Regret}(\piepessi)$.

\begin{corollary}[Regret Decomposition] \label{cor: sub-optimality decomposition}
  Recall that $\piepessi := \argsup_{\pie} \inf_{h\in\cS(e_n)} v(h, \pi).$ Suppose that Assumptions \ref{asp: identifiability and realizability}, \ref{asp: compatibility of test function class} and \ref{asp: regularity of function classes} hold. If $e_n>13/4 \cdot \eta_n^2,$ then on the event $\cE,$ the regret corresponding to $\piepessi$ is bounded by
\begin{align}  \label{eq:simplify_regret1}
    \text{Regret}(\piepessi){\leq} v^{\piestar}_{\alpha} - v ( \hpessi{\piestar}, \piestar ) = \EE_{\pin{\piestar}}\sbr{\hstaralpha(A, X)-\hpessi{\piestar}(A, X)}.
\end{align}
\begin{proof}
    See \S\ref{proof: pessimism} for a detailed proof.
\end{proof}
\end{corollary}
\subsection{Local Expansion within the Solution Set}\label{sec: local expansion CI}

The upper bound in Corollary \ref{cor: sub-optimality decomposition} involves $\hpessi{\piestar} - \hstaralpha$, whose error cannot be directly connected to the RMSE $\nbr{\cT h}_{2}$ for some $h$. 
In the following, we employ a local expansion of $\cT\hpessi{\piestar}$ around $\hstaralpha$ to bridge $\| \hpessi{\piestar} - \hstaralpha\|_{\bullet}$ for some norm $\| \cdot \|_{\bullet}$ to $\| \cT\hpessi{\piestar}\| _{2}$. 
To this end, we need to first ensure that any element in the solution set $\cS(e_n)$ is a consistent estimator of $\hstaralpha$.

Before presenting such a result, we first impose a regularity condition on the conditional density of $Y$ given $(A, X, Z)$.
\begin{assumption}[Regularity of Conditional Density]\label{asp: regularity of Density}
We assume that the conditional density of $Y$ given $(A, X, Z),$ $\pr_{Y\given A, X, Z=(a, x, z)}(y)$ is continuous in $(y, a, x, z)$ and $\sup_{y}\pr_{Y\given A, X, Z}(y) \leq  K$ for some $K > 0$ for almost all $(A, X, Z).$ 
\end{assumption}
Assumption \ref{asp: regularity of Density} imposes some mild assumptions on the conditional density of $Y$ given $(A, X, Z),$ which is used to ensure that $\vT$ is a continuous operator with respect to $\nbr{\cdot}_{\infty}.$ The assumption is standard in NPQIV \citep{chen2012estimation}. We now give a consistent result for any $\vh \in \cS(e_n).$ 
\begin{theorem}[Consistency of functions in $\cS(e_n)$]\label{thm: consistency}
    Suppose that Assumptions \ref{asp: identifiability and realizability}, \ref{asp: compatibility of test function class}, \ref{asp: regularity of function classes} and \ref{asp: regularity of Density}  hold. If $e_n=\cO(\eta_n^2 )$,  then on the event $\cE,$ for any $\vh \in \cS(e_n),$  we have $\nbr{\vh - \hstaralpha}_{\infty} = o_{p}\rbr{1}$  and hence $\nbr{\vh - \hstaralpha}_{2} = o_{p}\rbr{1}.$ 
    \begin{proof}
    See \S\ref{pro: consistency} for a detailed proof.
    \end{proof}
\end{theorem}

Theorem \ref{thm: consistency} shows that if the threshold of the solution set is chosen properly, then any $\vh$ in $\cS(e_n)$ is a consistent estimator of $\hstaralpha.$ With consistency, one remaining step is required before advancing with local expansion: establishing the correct definition of the derivative of $\vT\vh$ with respect to $\vh.$ Following \citet{ai2012semiparametric}, we introduce the notion of pathwise derivative of $\cT\vh$ evaluated at $\hstaralpha$,  in the direction $\vh-\hstaralpha$,  as follows: 
\begin{align*}
    \frac{d\cT\hstaralpha}{d\vh}[\vh - \hstaralpha] := \frac{d\EE [W(D;(1-r)\hstaralpha + r\vh) \given X, Z]}{dr}\Given_{r=0}.
\end{align*}
We then define a pseudo-metric on $\cS(e_n)$ based on the pathwise derivative:
\begin{align}
\label{eq: pseudometric}
\norm{\vh_1-\vh_2}_{\mathrm{ps}} := \sqrt{\EE \bigg[ \Big( \frac{d\cT\hstaralpha}{d\vh}[\vh_1-\vh_2] \Big) ^{2} \bigg] },
\end{align}
where the term inside the expectation is defined as
\begin{align*}
\frac{d\cT\hstaralpha}{d\vh}[\vh_1-\vh_2] = \frac{d\cT\hstaralpha}{d\vh}[\vh_1-\hstaralpha] - \frac{d\cT\hstaralpha}{d\vh}[\vh_2-\hstaralpha].
\end{align*}
In other words, for any $h_1$ and $h_2$, to compute the pseudo-metric $\norm{h_1-h_2}_{\mathrm{ps}}$, we first compute the pathwise derivative of $\cT\hstaralpha$ evaluated at $\hstaralpha$ in the direction of $h_1-\hstaralpha$ and $h_2-\hstaralpha$, and then take the square of the difference of the pathwise derivatives. In particular, suppose $\cT$ is a linear operator, e.g., $\cT$ can be viewed as a matrix. Then the pseudo-metric $\norm{\vh_1-\vh_2}_{\mathrm{ps}}$ reduces to a weighted $\ell_2$ norm, i.e., $\norm{\vh_1-\vh_2}_{\mathrm{ps}} ^2 =  ( \vh_1-\vh_2)^\top (\cT ^\top \cT )( \vh_1-\vh_2).$  
In Lemma \ref{lem: directional derivative}, we derive the closed form of the pseudo-metric for the IV case, which states that  $$\norm{\vh-\hstaralpha}_{\mathrm{ps}} = \sqrt{\EE[(\EE[\pr_{Y|{A,X,Z}}(\hstaralpha(A, X))\cbr{\vh(X,A) - \hstaralpha(A, X)}\given X, Z])^2]}.$$

Let $\vH_{\epsilon} := \left\{ \vh\in\vH: \nbr{\vh-\hstaralpha}_{2} \leq \epsilon\right\}\cap \cS(e_n)$ be the restricted space of $\cS(e_n)$ around the neighborhood of $\hstaralpha$ where $\epsilon$ is a sufficiently small positive number such that $\prob(\cE \cap  \{ \hpessi{\piestar} \in \vH_{\epsilon} \} ) \geq 1 - 3\xi.$ Such an $\epsilon$ is guaranteed to exist by Condition \ref{con: high probability event} and Theorem \ref{thm: consistency}.
We introduce the following assumption on the local curvature of $\cT$ around $\hstaralpha$. 
\begin{assumption}[Local Curvature]\label{asp: local curvature}
If we set $e_n>13/4 \cdot \eta_n^2$, then there exists a sufficiently small positive number $\epsilon$ and a finite constant $c_0>0$ such that for any $\vh \in \vH_{\epsilon},$ $\norm{\vh-\hstaralpha}_{\mathrm{ps}} \leq c_0 \nbr{\vT\vh}_{2}$. 
\end{assumption}
Assumption \ref{asp: local curvature} characterizes the local curvature of $\cT$. In particular, note that $\cT \hstaralpha = 0$. Thus, under (ii) we have $$\norm{\vh-\hstaralpha}_{\mathrm{ps}} \leq c_0 \cdot  \nbr{\vT\vh - \vT \hstaralpha}_2,$$
which ensures that the RMSE dominates the pseudo-metric in the neighborhood of $\hstaralpha$. 
The parameter $c_0$ controls the local curvature of $\cT$ around $\hstaralpha$. This assumption holds when the conditional moment restriction is locally linear around $\hstaralpha.$ 
See the detailed discussion in \S\ref{app: discussion assumptions}. 

\subsection{Regret of the Solution Set Offline Policy}\label{sec: regret SS}
Assumption \ref{asp: local curvature} allows us to do local expansion and link $\norm{\vh-\hstaralpha}_{\mathrm{ps}}$ to $\nbr{\vT\vh}_{2}.$ 
In the following, we establish the regret of the pessimistic estimator $\piepessi$ in \eqref{eq: pessimistic estimator} based on the solution set $\cS(e_n)$.

Before presenting the theorem, we introduce an assumption on the coverage of the offline data. 
\begin{assumption}[Change of Measure]\label{asp: data coverage}
For the marginal distribution of context $\tpr$ in the interventional process and the optimal interventional policy $\piestar$, there exists a function $b: \cX \times \cZ \rightarrow \RR$ such that $\EE\sbr{b^2(X, Z)} < \infty$ and
\begin{align}\label{cond: b1 PV extended}
    \EE\sbr{b(X, Z)\pr_{Y|{A,X,Z}}(\hstaralpha(A, X))\given A=a, X=x} = \frac{\tpr(x)\piestar(a\given x)}{\pr(x, a)}.
\end{align}
\end{assumption}
Assumption \ref{asp: data coverage} can be interpreted as a condition that requires the offline data in ODCP to cover the distribution induced by the oracle policy $\piestar$. The existence of $b(x, z)$ bridges the joint distribution of $(X, A)$ under the ODCP and the one induced by $\piestar.$ Recall that the upper bound of regret in Corollary \ref{cor: sub-optimality decomposition} involves evaluating the expectation under the interventional policy $\piestar.$ The ability to change between measures enables us to evaluate this expectation under the distribution of ODCP. 
Now we are ready to bound the   regret of $\piepessi.$
\begin{theorem}[Regret of Solution Set Algorithm]\label{thm: convergence of sub-optimality}
Suppose that structural model Assumption 
\ref{asp: model} holds.
Suppose that Assumptions
\ref{asp: identifiability and realizability},  \ref{asp: compatibility of test function class}, and 
\ref{asp: regularity of function classes} for function classes $\vH$ and $\Theta$ hold. Suppose also that Assumptions \ref{asp: local curvature} and \ref{asp: data coverage} hold.
If the threshold $e_n$ for the solution set satisfies $e_n> 13 /4 \cdot \eta_n^2,$ then, conditioning on the event $\cE \cap \{ \hpessi{\piestar} \in \vH_{\epsilon} \}$, the regret of $\piepessi$ is
\begin{align*}
    \text{Regret}(\piepessi) = c_0\cdot \nbr{b}_{2}\cdot  \cO\rbr{\sqrt{e_n} +\eta_n}.
\end{align*}
\begin{proof}
See \S\ref{proof: subopt} for a detailed proof.
\end{proof}
\end{theorem}

This theorem shows that the regret is bounded by  
$\cO\rbr{\sqrt{e_n} +\eta_n}$ 
when $c_0$ and $\| b\|_2$ are regarded as constants.
If we set $e_n = \cO(\eta_n^2),$ Theorem \ref{thm: convergence of sub-optimality} implies that the regret of $\piepessi$ is bounded by $\cO(\eta_n).$ As we will show in Theorem \ref{thm: concentration bound_linear} in  \S\ref{app: bracketing concentration},  $\eta_n = \tilde{\cO}(n^{-1/2})$ when $\vH$ and $\Theta$ are chosen to be linear spaces, which corresponds to a “fast statistical rate” \citep{uehara2021finite}. This rate aligns with those reported in the existing minimax estimation literature \citep{dikkala2020minimax, li2022pessimism}. It is noteworthy that this rate is equivalent to that observed in scenarios involving linear objectives \citep{chen2023unified}, where the average reward is maximized rather than the average quantile. This indicates that there is no loss of efficiency when extending the analysis to the nonlinear case.

\subsection{Theoretical Analysis for the Regularized Algorithm}\label{sec: theoretical analysis for the regularized set version algorithm}
For the regularized algorithm, we apply a similar approach using local expansion at $\hstaralpha$. 
Specifically, for any fixed policy $\pie$, we define $\hpessi{\pie}_{R}$ as 
\begin{align}\label{eq: hpessi reg}
    \hpessi{\pie}_{R} := \arg\inf_{h\in\vH} \{v(h, \pie) + \lambda_n \cdot \bigeps(h)\}.
\end{align}

By this definition, we can simplify the regret of $\piepessi_R$ similar to \eqref{eq:simplify_regret1} as follows.

\begin{lemma}[Regret Decomposition]\label{cor: sub-optimality decomposition reg}
The regret of $\piepessi_R$ is upper-bounded as follows:
\begin{align} \label{eq:regret_reg_bound}
    \text{Regret}(\piepessi_R) 
    &\leq \EE_{\pin{\piestar}}\left[\hstaralpha(A, X) - \hpessi{\piestar}_R(A, X)\right] 
    + \lambda_n \cdot \bigeps(\hstaralpha).
\end{align}
\end{lemma}

\begin{proof}
    See \S\ref{proof: convergence of sub-optimality reg} for a detailed proof.
\end{proof}
Lemma \ref{cor: sub-optimality decomposition reg} states that the regret of $\piepessi_R$ can be upper bounded by the average difference between $\hstaralpha $ and $\hpessi{\piestar}_R $ in the interventional process, plus a regularization term. Since we define $\bigeps(\hstaralpha)$ as $\vcL_{n}(\hstaralpha) - \inf_{\vh\in\vH}\vcL_{n}(\vh)$, this regularization term is comparatively straightforward to manage. The challenging part, $\vcL_{n}(\hstaralpha)$, has already been addressed in Theorem \ref{thm: uncertainty quantification} (i). 
Besides, $\inf_{\vh\in\vH}\vcL_{n}(\vh)$ is the minimum value of the empirical loss function $\vcL_{n}(\vh)$ over the hypothesis set $\vH$, which is often sufficiently small when the hypothesis set $\vH$ is rich enough. To capture this fact,  we impose the following assumption on $\inf_{\vh\in\vH}\vcL_{n}(\vh)$.

\begin{assumption}[Sample Criterion]\label{asp: sample criterion}
    We assume that $\inf_{\vh\in\vH}\vcL_{n}(\vh) = \cO(\eta_n^2).$
\end{assumption}

Assumptions similar to Assumption \ref{asp: sample criterion} are commonly encountered in the literature of M-estimation \citep{van2000asymptotic}. It holds when the empirical loss $\vcL_{n}(\vh)$ can be minimized to a negligible level across the hypothesis set $\vH$. This is often satisfied when the optimization method employed in practice is effective.
Consequently, by appropriately selecting $\lambda_n$, the regularization term mentioned in Lemma \ref{cor: sub-optimality decomposition reg} can be effectively bounded. 

It remains to bound the first term on the right-hand side of \eqref{eq:regret_reg_bound}. 
We adopt a similar strategy as in the solution set algorithm. Specifically, we prove that $\hpessi{\pie}_{R}$ is consistent to $\hstaralpha$ with any  $\pi$. Consequently, we perform a local expansion of $\hpessi{\piestar}_{R}$ around $\hstaralpha,$ allowing the regret to be linked to $\nbr{\vT\hpessi{\piestar}_{R}}_2$, which can be further bounded by analyzing the loss function $\vcL_{n}$. 

The following theorem establishes the consistency of $\hpessi{\pie}_{R}$ for $\hstaralpha$ under the regularized algorithm.
 
\begin{theorem}[Consistency of $\hpessi{\pie}_{R}$]\label{thm: consistency of the estimator of the regularized version algorithm}
Suppose that Assumptions \ref{asp: identifiability and realizability}, \ref{asp: compatibility of test function class}, \ref{asp: regularity of function classes}, \ref{asp: regularity of Density} and \ref{asp: sample criterion}  hold.  For any $\epsilon_{R} > 0,$ if we set $\lambda_n \ge \eta_n^{-(1 + \epsilon_{R})},$ then on the event $\cE,$ for any $\pie,$ we have $\|\hpessi{\pi}_{R} - \hstaralpha \| _{\infty} = o_{p}\rbr{1}$ and hence $  \| \hpessi{\pi}_{R} - \hstaralpha \| _{2} = o_{p}\rbr{1}.$ 
\begin{proof}
    See \S\ref{pro: consistency of the estimator of the regularized version algorithm} for a detailed proof.
\end{proof}
\end{theorem}
Theorem \ref{thm: consistency of the estimator of the regularized version algorithm} states that on  event $\cE,$ the estimated structural quantile function $\hpessi{\pi}_{R}$ is consistent to the ground-truth $\hstaralpha$. Note the regularization parameter $\lambda_n$ in Theorem \ref{thm: consistency of the estimator of the regularized version algorithm} only depends on $\eta_n,$ which characterizes the magnitude of the concentration error. The quantity $\epsilon_{R}$ appears in the definition of $\lambda_n$ is crucial for the consistency result. Although consistency is theoretically guaranteed for infinitesimally small $\epsilon_{R}$, the convergence rate of $\hpessi{\piestar}_{R}$ to $\hstaralpha$ becomes slow when $\epsilon_{R}$ is too small. As we will see later, the choice of $\epsilon_{R}$ induces a trade-off in the regret of $\piepessi_R$. 
Given the consistency result, we are ready to perform a local expansion of $\hpessi{\piestar}_{R}$ around $\hstaralpha$ in a restricted space of $\vH$ around the neighborhood of $\hstaralpha$. For a fixed $\lambda_n \ge \eta_n^{-\rbr{1 + \epsilon_{R}}},$ we fix a sufficiently small positive number $\epsilon$ such that the event  $\cE \cap \cbr{\hpessi{\piestar}_{R} \in \left\{ \vh\in\vH: \nbr{\vh-\hstaralpha} _{2} \leq \epsilon\right\}}$ occurs with probability at least $1-3\xi.$ Such an $\epsilon$ is guaranteed to exist by Condition \ref{con: high probability event} and Theorem \ref{thm: consistency of the estimator of the regularized version algorithm}.
\begin{theorem}[Regret of the Regularized Algorithm]\label{thm: convergence of sub-optimality reg}
We assume all the assumptions made in Theorem \ref{thm: consistency of the estimator of the regularized version algorithm} hold and additionally impose  Assumptions  \ref{asp: local curvature} and \ref{asp: data coverage}. 
Let $0 < \epsilon_{R} <1$ be any fixed number. We set the regularization parameter $\lambda_n$ to  $\eta_n^{-(1 + \epsilon_{R})}$ in \eqref{eq:pess_reg}. 
Then, conditioning on the event $\cE \cap \cbr{\hpessi{\piestar}_{R} \in \left\{ \vh\in\vH: \nbr{\vh-\hstaralpha}_{2} \leq \epsilon\right\}}$, the regret of $\piepessi_R$ is 
\begin{align*}
    \text{Regret} (\piepessi_{R}) = \cO\rbr{\eta_n^{1 - \epsilon_{R}}}.
\end{align*}

\begin{proof}
    See \S\ref{proof: convergence of sub-optimality reg} for a detailed proof.
    \end{proof}
\end{theorem}

Theorem \ref{thm: convergence of sub-optimality reg} implies that if we set $\lambda_n = \eta_n^{-\rbr{1 + \epsilon_{R}}},$ the regret of $\piepessi_{R}$ is $\cO\rbr{\eta_n^{1 - \epsilon_{R}}}.$ Recall by Theorem \ref{thm: convergence of sub-optimality}, $\text{Regret}(\piepessi)$ is  $\cO(\eta_n)$ when $e_n$ is properly chosen. Thus, the regret of $ \piepessi_{R}$ is larger than that of $\piepessi$ by an infinitesimal amount $\epsilon_{R}$. A small value of $\epsilon_{R}$ slows the convergence of $\hpessi{\piestar}_{R}$ toward $\hstaralpha$. As a result, the constant $c_0$ in Assumption \ref{asp: local curvature} may be large, which can increase the regret. Essentially, the extra $\epsilon_{R}$ term can be viewed as the trade-off when converting an optimization problem with data-dependent constraints into a nonconstrained optimization problem in the nonlinear quantile case. 

Furthermore, when selecting the function spaces $\vH$ and $\vTheta$ as the linear function classes, we have $\eta_n = \tilde{\cO}(n^{-1/2})$ as shown in Theorem \ref{thm: concentration bound_linear} in \S\ref{app: bracketing concentration}. This leads to a regret rate of $\tilde{\cO}(n^{-1/2(1 - \epsilon_{R})})$ for the regularized algorithm, which outperforms the $\tilde{\cO}(n^{-1/3})$ rate associated with the regularized version of the pessimistic offline reinforcement learning algorithm studied in \cite{xie2021bellman}. Since $\epsilon_{R}$ can be infinitesimally small (ignoring $c_0$), our regret bound  also closely matches the $\tilde{\cO}(n^{-1/2})$ regret of the augmented Lagrangian algorithm detailed in \cite{rashidinejad2022optimal}. Note that \citet{rashidinejad2022optimal} requires the knowledge of the propensity score, which is a strong assumption and is not required by our method. Furthermore, we would like to emphasize that the quantile objective considered in our work is a nonlinear functional of the reward distribution,
whereas \cite{xie2021bellman} and \cite{rashidinejad2022optimal} focus on the expected reward, which is a linear functional of the reward distribution. To the best of our knowledge, we are the first to establish the efficacy of the pessimism in the context of nonlinear quantile objectives and unobserved confounding, which is a significant extension of the existing literature on pessimistic offline reinforcement learning.

\section{Simulation Experiments} \label{sec: experiment}

In this section, we evaluate the performance of our approach via numerical simulations. We focus on a synthetic setting where the instrumental variables are available in the offline data.
We compare the pessimistic regularized policy learning algorithm, specified in \eqref{eq:pess_reg}, against a greedy benchmark that does not incorporate pessimism. 
The simulation results demonstrate the benefits of pessimism, particularly in scenarios with insufficient data coverage.

\subsection{Data Generating Model} 
\label{sec:simulation_data}

We consider a contextual bandit model with two-dimensional contexts, denoted by $X = (X_1, X_2)$, a binary action $A \in \{0, 1\}$, and a one-dimensional instrumental variable $Z$. The output variable $Y$ is generated according to \eqref{eq:model_assumption}, where the structural error $\epsilon$ represents an unmeasured confounder that simultaneously influences the action and the outcome.
We detail the data generation model below. 

\paragraph{Context and instrument generation.} 
We let $(X_1, X_2, Z)$ be drawn from a centered trivariate Gaussian distribution:
\[
    \begin{bmatrix} X_1 \\ X_2 \\ Z \end{bmatrix} \sim \mathcal{N} \left( \begin{bmatrix} 0 \\ 0 \\ 0 \end{bmatrix}, \begin{bmatrix} 1 & \rho & \rho \\ \rho & 1 & \rho \\ \rho & \rho & 1 \end{bmatrix} \right),
    \]
where $\rho$ is a correlation coefficient. We set $\rho$ to $0.95$ in all experiments.

\paragraph{Structural error $\epsilon$.} For a given target quantile $\alpha$, we sample 
the noise $\epsilon$ according to $\epsilon \sim \mathcal{N}(-\Phi^{-1}(\alpha), 1)$, where $\Phi(\cdot)$ is the cumulative distribution function (CDF) of a standard normal distribution.  This construction ensures that the $\alpha$-quantile of $\epsilon$ is zero, i.e., $\mathbb{P}(\epsilon \le 0) = \alpha$.

\paragraph{Action $A$.}
Given a treatment assignment score $S$, which is introduced below, we generate a binary action $A$ from a Bernoulli distribution.
In particular, $\PP(A = 1 \given S ) = \text{sigmoid}(S) = 1 / (1 + \exp(-S)) $, where $\text{sigmoid}(\cdot )$ is the sigmoid function.

\paragraph{\textbf{Outcome $Y$.}} 
As shown in \eqref{eq:pess_reg}, we generate $Y$ according to the structural model:
\begin{align}\label{eq:model_simulation}
Y = \beta_1^* \cdot X_1 +  \beta_2^* \cdot X_2 + \beta_3^* \cdot  A \cdot   X_1 + \beta_4^* \cdot  A  \cdot  X_2 + \epsilon,
\end{align}
where $\beta ^* = (\beta_1^*, \beta_2^*, \beta_3^*, \beta_4^*)^\top  = (1, 1, 3,2)^\top $ are the true coefficients.  
Here, the term $\beta_1^* \cdot X_1 +  \beta_2^* \cdot X_2 + \epsilon$ represents the baseline outcome, and $A \cdot (\beta_3^* \cdot X_1  + \beta_4^*\cdot  X_2)$ is the additional effect due to action $A$.

\paragraph{Treatment assignment score $S$.} 
The treatment assignment score $S$ is designed to introduce a lack of coverage in the data.
Specifically, we introduce two parameters, a structured proportion $p \in [0, 1]$ and two treatment biases $b_{+}$ and $b_{-}$. 
For a fraction $p$ of the $n$ samples, we generate $S$  as follows:
\begin{align}\label{eq:assignment_bias}
S = X_1 + 0.5 \cdot X_2^2 + Z + b_{+} \cdot \ind(X_1 \ge 0) + b_{-} \cdot \ind (X_1 < 0) + \epsilon .
\end{align}
Here we set $b_{+} = 8$ and $b_{-} = -5$. Moreover, for the remaining fraction $(1 - p)$ of the samples, we generate $S$ from random noise: $S \sim \mathcal{N}(0, 1)$.

Our construction induces two layers of complexities in  data coverage. 
First, by having a nonzero $p$, we introduce out-of-distribution data points that are not generated from the confounded structural model. 
We set $p$ as $0.7$ and $0.8$ in our experiments.  
Second, by having the bias terms $b_{+}$ and $b_{-}$, we introduce a systematic bias in the treatment assignment. That is, we have more assigned actions for the region $X_1 \ge 0$ than for the region $X_1 < 0$.

\paragraph{Optimal policy $\pi^*$.} 
By \eqref{eq:model_simulation}, it is straightforward to see that the quantile-optimal policy $\pi^*$ is a deterministic policy that outputs action $A = 1$  if and only if $\beta_3^* \cdot X_1 + \beta_4^* \cdot X_2 > 0$, i.e., $3 X_1 + 2 X_2 > 0$.
When this condition is not satisfied, the optimal action is $A = 0$.

\subsection{Algorithm Implementations}

We compare the pessimistic regularized policy learning algorithm (Algorithm \ref{algo:regularized}) against a greedy baseline that does not incorporate pessimism.

\paragraph{Linear hypothesis classes $\mathcal{H}$ and $\Theta$.}
We set $\mathcal{H}$ and $\Theta$ to be linear function classes with known features. 
Specifically, 
motivated by \eqref{eq:model_simulation}, any function $h $ in $\mathcal{H}$ can be written as 
\begin{align}\label{eq:linear_hypothesis_h}
h(A, X; \beta) = \beta_1 X_1 + \beta_2 X_2 + A (\beta_3 X_1 + \beta_4 X_2),
\end{align}
where $\beta = (\beta_1, \beta_2, \beta_3, \beta_4)^\top$ is the parameter vector. 
The test function space $\Theta$ is defined as a linear span of a set of polynomial basis functions involving $X_1, X_2,$ and $Z$: 
$$
\bigl \{X_1, ~X_2, ~Z, ~  X_1X_2,~ X_1Z,~ X_2Z,~  X_1^2,~ X_2^2,~ Z^2, ~ X_1^2Z, ~  X_1X_2^2, ~  Z^3, ~   X_1^2X_2^2\bigr\}.
$$
 
\paragraph{Greedy approach.} This approach first estimates the structural quantile function $h_\alpha(A,X)$ by minimizing the empirical loss function $\mathcal{L}_n(h)$ in  \eqref{def: empirical loss function}. 
Note that supremum over $\theta $ in \eqref{def: empirical loss function} is a quadratic form, which can be solved exactly in closed form.
Thus, we can directly compute the gradient of $\mathcal{L}_n(h)$ with respect to the parameters $\beta$ of $h$.
We  obtain an approximate minimizer $\hat h$  via direct gradient descent, with a constant stepsize $0.05$. After obtaining $\hat h$, we derive the greedy policy by solving $\max_{\pi} v(\hat{h}, \pi)$, which can be computed in closed form.

\paragraph{Pessimistic approach.} We approximately implement Algorithm \ref{algo:regularized} with $\lambda_n = \sqrt{n}$, where $n$ is the sample size. This choice of $\lambda_n$ is suggested by Theorem \ref{thm: convergence of sub-optimality reg} for linear function classes, with $\epsilon_{R}$ set to zero. 
Note that \eqref{eq:pess_reg} involves a minimax optimization and that $\sup_{\pi} v(h, \pi)$ for any fixed $h \in \mathcal{H}$ can be computed in closed form.
We exchange the order of the supremum and infimum in \eqref{eq:pess_reg} and reduce the problem to solving 
\begin{align}\label{eq:pess_reg_new}
\inf_{h \in \mathcal{H}} \bigl\{ V(h) + \lambda _n \cdot \mathcal{L}_n (h)\bigr\}, \qquad \text{where}~~V(h) = v\big(h, \pi(h)\big).
\end{align}
Here we let $\pi(h)$ denote the greedy policy  with respect to $h$. 
For example, for the function $h(\cdot; \beta)$ defined in \eqref{eq:linear_hypothesis_h}, the greedy policy $\pi(h)$ takes action one if and only if $\beta_3 X_1 + \beta_4 X_2 > 0$.

Another challenge is that $V(h)$ in \eqref{eq:pess_reg_new} is not differentiable with respect to the parameters of $h$. 
Thanks to the low dimensionality of $\beta$, to solve this minimization problem, we use a randomized search method. That is, we draw $N=10^4$ random samples of $\beta$ from a Gaussian distribution centered at $\hat h$,  evaluate the objective values in \eqref{eq:pess_reg_new} for each sampled $\beta$, and select the minimizer $\tilde h$. 
The greedy policy corresponding to $\tilde h$ is regarded as an approximation of $\hat \pi_{R}$ in Algorithm \ref{algo:regularized}.

\subsection{Experiment Results} 

\begin{figure}[h]
    \centering
    \begin{tabular}{ccc}
        \includegraphics[width=0.355\textwidth, height=0.177\textheight]{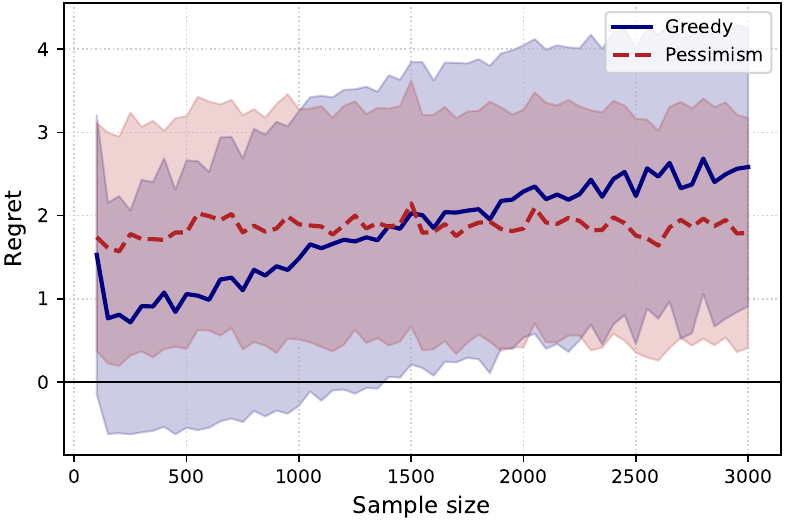} &
        \includegraphics[width=0.335\textwidth, height=0.177\textheight]{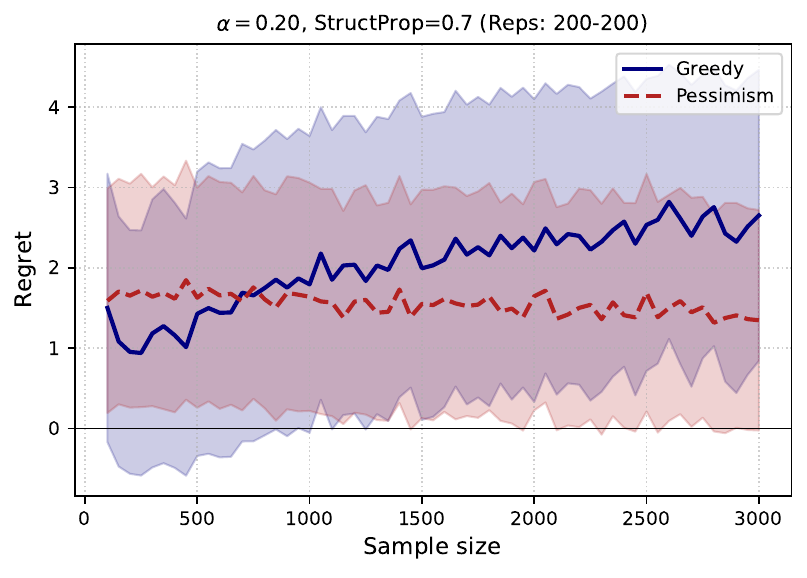} &
        \includegraphics[width=0.331\textwidth, height=0.181\textheight]{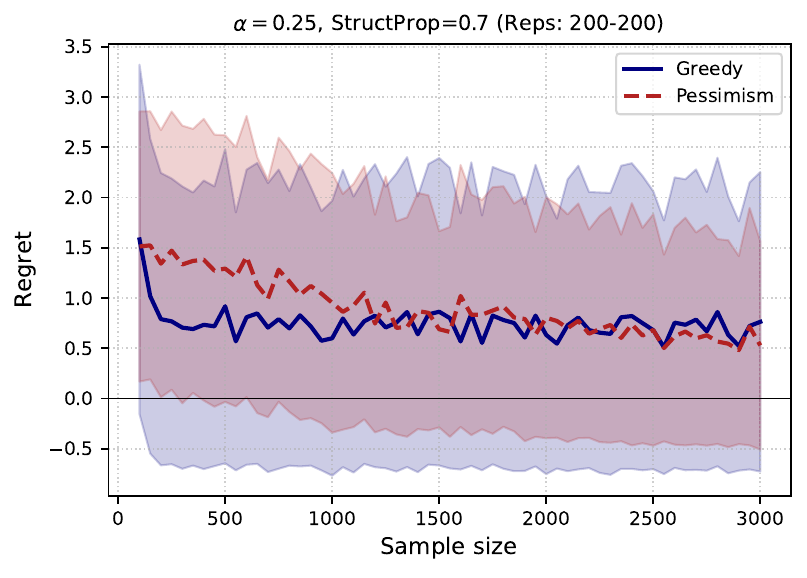} \\
        (a) $(\alpha , p) = (0.15, 0.7)$ & (b) $(\alpha , p) = (0.2, 0.7)$ & (c) $(\alpha , p) = (0.25, 0.7)$ \\
    \end{tabular}
    \caption{Average regret of the greedy and pessimistic approaches for different sample sizes $n$ and quantile levels $\alpha$, with structured proportion $p = 0.7$. The first two plots clearly shows that pessimism outperforms the greedy approach. The third plot shows that their performances are comparable, with pessimism slightly better when $n$ is large.}
    \label{fig:results_prop_7}
\end{figure}

\begin{figure}[htbp]
    \centering
    \begin{tabular}{ccc}
        \includegraphics[width=0.355\textwidth, height=0.179\textheight]{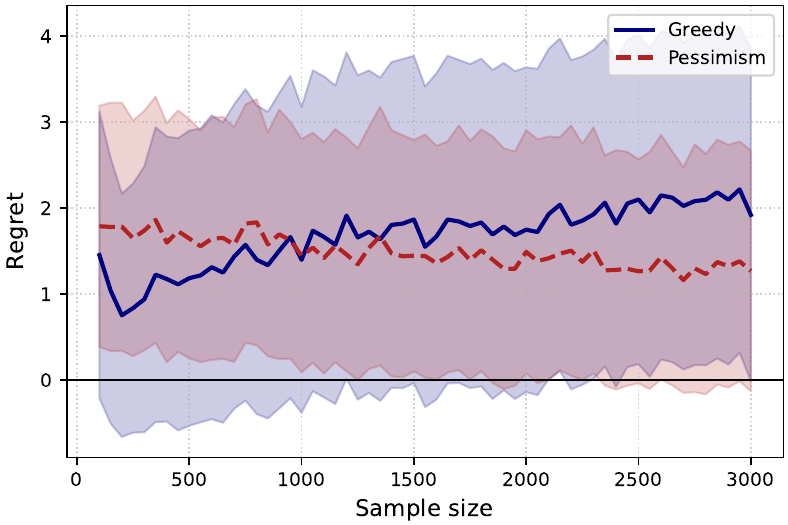} &
        \includegraphics[width=0.335\textwidth]{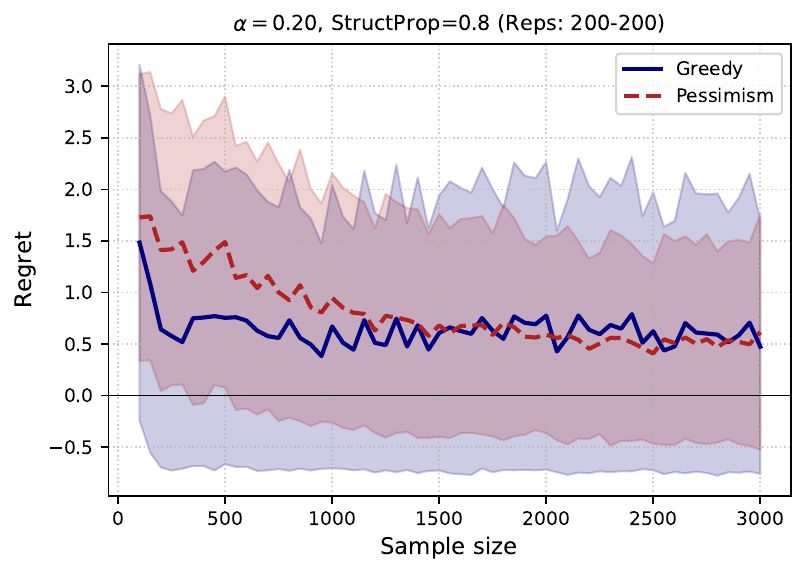} &
        \includegraphics[width=0.331\textwidth]{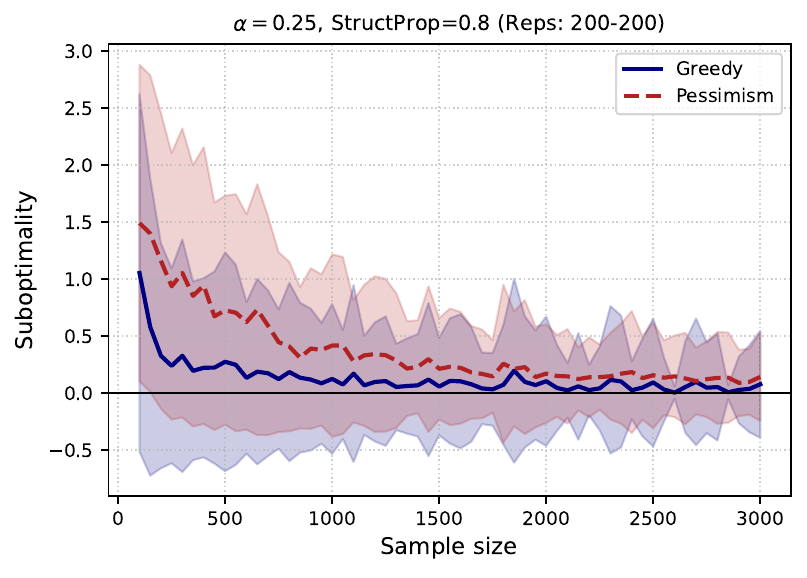} \\
        (a) $(\alpha , p) = (0.15, 0.8)$ & (b) $(\alpha , p) = (0.2, 0.8)$ & (c) $(\alpha , p) = (0.25, 0.8)$\\
    \end{tabular}
    \caption{Average regret of the greedy and pessimistic approaches for different sample sizes $n$ and quantile levels $\alpha$, with structured proportion $p = 0.8$.The first two plots clearly shows that pessimism outperforms the greedy approach. The third plot shows the greedy approach outperforms the pessimistic approach when $\alpha = 0.25$.}
    \label{fig:results_prop_8}
\end{figure}

We introduce the parameter of structured proportion $p$ in Section \ref{sec:simulation_data}. 
We compare our pessimistic algorithm against the greedy baseline under different configurations of $(n, \alpha, p)$, where $n$ is the sample size and $\alpha$ is the target quantile. 
In particular, we choose $n \in \{ 100, 150, 200, \ldots, 3000\}$, $\alpha \in \{ 0.15, 0.2, 0.25\}$, and $p \in \{ 0.7, 0.8\}$. 
For each configuration, we generate $200$ independent datasets and compute the average regret of the learned policies. 
We show the results for $p= 0.7$ and $p=0.8$ in Figures \ref{fig:results_prop_7} and \ref{fig:results_prop_8}, respectively.
The blue and red curves represent the average regret of the greedy and pessimistic approaches, respectively, while the shaded areas around the curves indicate one standard deviation above and below the mean. Our findings are summarized as follows.

\begin{itemize}
\item In all these figures, the width of the shaded areas is similar for both the greedy and pessimistic approaches, indicating that the variability of the regret is comparable for both methods.
\item For the quantile levels $\alpha = 0.15$ and $\alpha = 0.2$, the {\bf pessimistic approach consistently outperforms the greedy approach} when the sample size is large. This clearly demonstrates the advantage of incorporating pessimism, especially when the data coverage is insufficient.
In particular, with $(1-p)$ portion of out-of-distribution data added, 
it seems to suggest that the signal from the observed data is not sufficient to estimate $\hstaralpha$ accurately. In this case, pessimism benefits the learning process by avoiding overfitting to the observed data distribution, leading to lower regret. 

\item Furthermore, it is worth noting that the reason why the greedy approach is inferior is not due to the limited sample size, but rather due to the insufficient coverage of the data generating distribution. In particular, as shown in the first two plots of Figures \ref{fig:results_prop_7} and \ref{fig:results_prop_8}, the margin between the greedy and pessimistic approaches is larger when the sample size is larger. 

\item For the quantile level $\alpha = 0.25$, pessimism is slightly worse than the greedy approach. In particular, when $p = 0.7$, they are comparable when $n$ is large, as shown in Figure \ref{fig:results_prop_7}-(c). For some sample sizes, pessimism is slightly better. 
For $p = 0.8$, the greedy approach outperforms the pessimistic approach, as shown in Figure \ref{fig:results_prop_8}-(c). In this case, both algorithms perform well with small regret. 
This case suggests that when the data coverage is sufficient, greedy algorithm itself can achieve good performance, and the pessimistic approach does not provide significant additional benefits.
Moreover, note that the pessimistic approach is computationally more expensive, requiring random sampling in our implementation, and the optimization error can also contribute to the regret. Thus, when the data coverage is sufficient, the greedy approach can be preferred due to its simplicity and efficiency.

\item Furthermore, if we compare the performances of these algorithms across different values of $p$, we observe that both methods perform better when $p = 0.8$ than when $p = 0.7$. This is expected, as a larger structured proportion $p$ implies that the data is more representative of the underlying distribution, leading to better performance for both methods.
\end{itemize}

\section{Discussion}\label{sec: conclusion}
In this work, we propose a quantile-optimal policy learning algorithm where the goal is to find a policy whose reward distribution has the largest $\alpha$-th quantile for some $\alpha \in (0, 1)$. The idea is to (i) establish a variety of conditional moment restrictions for the identification of the quantile objectives of any policy with the help of instrumental variables and negative controls; (ii) adopt a minimax estimation approach with nonparametric models to solve these conditional moment restrictions, and construct conservative solutions for pessimistic policy learning. The final policy is the one that maximizes the pessimistic objective. In addition, we propose a novel regularized policy learning method that is computationally benign. We deliver a theoretical analysis to show that the proposed method is provably efficient. To the best of our knowledge, this is the first sample-efficient policy learning algorithm for estimating the quantile-optimal policy when there exists unmeasured confounding. Our work requires the observation of either instrumental variables or negative controls. However, we posit that the proposed algorithm is equally effective when the ODCP possesses different underlying causal structures, as long as the quantile treatment effect can be identified by conditional moment restrictions. We leave this investigation to future research.
\newpage
\bibliographystyle{ims}
\bibliography{reference}

\newpage 
\appendix


\section{Discussion of Assumptions and Conditions in Section \ref{sec: theoretical results}} \label{app: discussion assumptions}
\paragraph{Assumption \ref{asp: identifiability and realizability}  (Identifiability and Realizability).} This assumption, often referred to as a global identification assumption, is commonly made in nearly all nonparametric quantile IV literature. For instance, see Assumption 3.2(ii) in \cite{chen2012estimation}, Assumption 1 in \cite{gagliardini2012nonparametric}, and Assumption 1 in \cite{horowitz2007nonparametric}.

 Regarding sufficient conditions for this assumption to hold, \cite{chernozhukov2005iv} provides such conditions in the case where the context $X$ is discrete. For more general setups, \cite{chen2014local} discusses conditions for local identification in detail. 
 The only study we are aware of that carefully outlines conditions for global identification is \cite{wong2022equation}. However, Wong's approach relies on a set of strong regularity conditions to establish these results.

\paragraph{Condition \ref{con: high probability event}:} We only verify that Condition 4.4 holds true for linear function classes because they already encompass a wide variety of functions, including polynomial splines, B-splines, wavelets, and Fourier series basis functions. These function classes can be represented as linear combinations of basis functions, making them instances of the linear function class. By examining our proof in \S\ref{app: bracketing concentration}, we rely only on the property that linear function classes have a covering number and bracketing number bounded by $A_n \log (C_3 / t)$ where $A_n$ is the dimension of the linear function class. Therefore, we can actually obtain different convergence rates $\eta_n$ under the supremum norm for the chosen function classes. The bracketing and covering numbers of various function classes can be found in different literature. For example, \cite{van2013weak} provides an extensive treatment of covering numbers and bracketing numbers. They discuss these concepts for a variety of function classes, including: Sobolev spaces, Besov spaces, Hölder classes, and reproducing kernel Hilbert spaces.

\paragraph{Assumption \ref{asp: regularity of Density} (Regularity of Conditional Density).} This paper assumes that the outcome variable $Y$ is continuous. The primary reason is that we are dealing with a \textbf{nonlinear} functional $\cT \vh$. To link the regret to  $\cT \vh,$, we conduct a local expansion of $\cT \vh$ around the neighborhood of the structural quantile function $\hstaralpha$. For this expansion, it is essential that the density of $Y$ is continuous and bounded, ensuring that the pathwise derivative $\cT \vh$ at $h = \hstaralpha$ exists. We remark that this assumption is needed in most of the nonparametric quantile IV literature. For instance, see Condition 6.1(i), (iv) in \citet{chen2012estimation}, Theorem 3.1 in \citet{abadie2002instrumental}, Condition A.1 in \citet{chernozhukov2005iv}, and Assumption A.1 in \citet{gagliardini2012nonparametric}. While our paper does not address discrete outcomes due to this constraint, we believe a similar algorithm could be applied in this case by first convolving the discrete outcome with Gaussian noise, which would smooth the distribution of $Y$. This smoothing allows for the use of techniques designed for continuous outcomes.

\paragraph{Assumption \ref{asp: local curvature} (Local Curvature).} Assumption \ref{asp: local curvature} asserts that the weaker pseudo metric $\norm{\vh-\hstaralpha}$ is Lipschitz continuous with respect to the population criterion function $\nbr{\vT\vh}_{2}$ within a small neighborhood around $\hstaralpha$. This condition is relatively mild and is commonly employed in the nonlinear functional analysis literature \citep{chen2012estimation, miao2023personalized}. Condition 6.3 (i) in \citet{chen2012estimation} provides a sufficient condition for the assumption to hold in a broad class of functional spaces $\cH$, including the linear function spaces we consider in  \S\ref{app: bracketing concentration}.

\paragraph{Assumption \ref{asp: data coverage} (Change of Measure).}
The Change of Measure Assumption \ref{asp: data coverage} is actually a weaker substitute of the standard concentrability assumption commonly used in the RL literature. In the tabular setting, Assumption \ref{asp: data coverage} is satisfied when the right-hand side of the assumption is uniformly bounded, and the Moore-Penrose inverse of the probability mass matrix $P(Z \given A, X)^{+}$ exists. This condition holds if the IV satisfies the standard completeness assumption: $\text{rank} [P(Z \given A, X)] \geq |\cA| \times |\cX|.$ We note that Theorem \ref{thm: convergence of sub-optimality} and Theorem \ref{thm: convergence of sub-optimality reg} remain valid if Assumption \ref{asp: data coverage} is replaced with the usual single-policy concentrability assumption: there exists a constant $\widetilde{c}>0$ such that $\sup_{x \in \cX, a \in \cA} \tpr(x)\piestar(a\given x)/{p(x, a)} \leq \widetilde{c}.$ Note $\tpr(x)\piestar(a\given x)/{p(x, a)}$ represents the density ratio of the distribution under the interventional process with the oracle policy $\piestar$ over that of the offline data collection process.

\section{Other Applications}\label{sec: other applications}
While the primary focus of this paper is on the general structural quantile model, it's important to note that our algorithm remains applicable as long as the problem can be reformulated into solving a conditional moment restriction. For instrumental variables, the conditional moment restriction $\EE\sbr{W(D; \vh)\given X, Z}=0$, can be replaced by solving $\EE\sbr{\tilde{W}(D; \vh)\given X, Z}=0$, where $\vh$ is the parameter of interest, related to a functional $\tilde{W}$ that is specified by the problem. In the case of NC, we can solve \eqref{eq: PV1} and \eqref{eq: PV2} after substituting $\tilde{W}(D; \vh)$ for $W(D; \vh)$. In particular, our algorithm is effective with more restricted forms of quantile models. 

\begin{example}[Single index quantile model \citep{wu2010single}]\label{ex: single index}
    The structural quantile function is restricted to the form $\vh(A, X)= h_{0}(\theta_0^{T}(X, A)),$ where $h_{0}$ and $\theta_0$ are unknown. The functional $\tilde{W}(D; \vh)$ is then defined as $\mathbb{1}\{Y \leq h_{0}(\theta_0^{T}(X, A))\} - \alpha.$ For this example to fall in our framework, we simply choose $\vH = \tilde{\vH} \times \RR^{\dim\cX} \times \RR^{\dim\cA}$ to be the function class that contains the function of the form: $h: \cX \times \cA \rightarrow \RR$ defined by $h(X, A) = \tilde{h}(\alpha_1^T X + \alpha_2^T A).$
\end{example}

\begin{example}[Partially linear quantile model \citep{lee2003efficient}]\label{ex: partially linear}
    The structural quantile function is restricted to the form $\vh(A, X)= h_{0}(A)+\theta_0^{T}X,$ where $h_{0}$ and $\theta_0$ are unknown. The functional $\tilde{W}(D; \vh)$ is then defined as $\mathbb{1}\{Y \leq h_{0}(A)+\theta_0^{T}X\} - \alpha$. We then choose $\vH = \tilde{\vH} \times \RR^{\text{dim}\cX},$ which contains the function of the form: $h: \cX \times \cA \rightarrow \RR$ defined by $h(X, A) = \tilde{h}(A)+\alpha_1^T X.$
\end{example}

When it is impossible to find a conditional moment restriction, we can still apply our approach in some cases. In Example \ref{ex: quantile-based risk measures}, we extend the method to a more general setting than maximizing the quantile function, demonstrating the flexibility of the proposed algorithm.
\begin{example}[Quantile-based risk measures \citep{dowd2006after}]\label{ex: quantile-based risk measures}
    We consider the spectral risk measure, $\int_{0}^{1}w(\alpha)\hstaralpha d \alpha,$ which is a member of the family of quantile-based risk measures. Here, $w(\alpha)$ is a known weighting function that makes the spectral risk measure coherent. To estimate the spectral risk measure, we consider a finite index set $\cI=\{\alpha_1, \dots, \alpha_m\}$ such that $\alpha_i \in (0, 1)$ for any $i$. We then define
    \begin{align}\label{eq: discrete quantile-based risk measure}
        M_{\phi}(A, X; h):=\sum_{i=1}^{m}\phi(\alpha_i)h_{\alpha_i}(A, X).
    \end{align}
    where $\vh = (h_{\alpha_1}, \dots, h_{\alpha_m}),$ for some known function $\phi$. The idea is that we simultaneously estimate $\{h_{\alpha_i}(A, X)\}_{i=1}^{m}$ for each $\alpha_i$ via the proposed algorithm. Then we approximate $-\int_{0}^{1}w(\alpha)h_{\alpha} d \alpha$ by $M_{\phi}(A, X; h)$. The choice of $\phi$ depends on the numerical method we used to approximate the integral. For example, Newton–Cotes method is one popular choice that has the form of \eqref{eq: discrete quantile-based risk measure}. Given that $\phi(\alpha_i)$ is fixed, we now introduce the policy-learning algorithm that minimizes the spectral risk measure or, equivalently, maximizes the negative spectral risk measure through the use of IV. The case for NC is similar. For each $i \in [m],$ we define  
    \begin{align*}
        W_i(D; \vh_{\alpha_i}) &= \mathbb{1}\cbr{Y \leq \vh_{\alpha_i}(A, X)} - \alpha_i,\\
        \cL_{i, n}(\vh_{\alpha_i}) &= \sup_{\theta\in\Theta}\cbr{\EE_n\sbr{W_i(D; \vh_{\alpha_i}) \theta(\vX, \vZ)}-1/2 \cdot  \nbr{\theta}_{n, 2}^2}.
    \end{align*}
    For the solution set algorithm, we further let  
    \begin{align*}
        \cS_i(e_n) &= \cbr{\vh_{\alpha_i}\in\vH:  \cL_{i, n}(\vh_{\alpha_i}) \le  \inf_{\vh_{\alpha_i}\in\vH}\cL_{i,n}(\vh_{\alpha_i}) + e_n}, \text{and}\\
        \cS(e_n) &= \cS_1(e_n) \times \dots \times \cS_m(e_n). 
    \end{align*}
    The estimated policy is then given by $\piepessi^{\text{RM}} :=\arg\sup_{\pie} \inf_{h\in\cS(e_n)} \EE_{\pin{\pie}}\sbr{M_{\phi}(A, X; h)}$. For the regularized algorithm, we define $\cE_{n}(\vh):= \sum_{i=1}^{m}\cL_{i, n}(\vh_{\alpha_i}) - \sum_{i=1}^{m}\inf_{\vh_{\alpha_i}\in\vH}\vcL_{i, n}(\vh_{\alpha_i})$. The estimated policy is given by $\piepessi^{\text{RM}}_{R}=\argsup_{\pie}\inf_{h\in \vH} \{\EE_{\pin{\pie}}\sbr{M_{\phi}(A, X; h)} + \lambda_n \cdot \bigeps(h)\}$.
We now present a result that characterizes the regret for the algorithms applied in Example \ref{ex: quantile-based risk measures}.
\begin{corollary} (Regret for Quantile-based Risk Measures).\label{cor: regret quantile based risk} 
    Suppose $\max_{i \in [m]}{\phi(\alpha_i)} = M.$
        \begin{itemize} 
        \item [(i)]
        Suppose that the assumptions in Theorem \ref{thm: convergence of sub-optimality} hold. Specifically, Assumption \ref{asp: identifiability and realizability} and Assumption \ref{asp: regularity of function classes} hold for each $\vh_{\alpha_i}^{*}.$ Assumption \ref{asp: local curvature} holds uniformly for all $\vh_{\alpha_i}^{*}$ for some constant $c_0.$  In addition, there exists a change of measure function for each $\alpha_i$ in Assumption \ref{asp: data coverage}. Then, $\text{Regret}(\piepessi^{\text{RM}}) =  \cO(Mm\cdot \eta_n)$ with probability $1-3m\xi.$
        \item[(ii)]   
        Suppose that the assumptions in Theorem \ref{thm: convergence of sub-optimality reg} hold. Specifically Assumption \ref{asp: identifiability and realizability} Assumption \ref{asp: regularity of function classes} hold for each $\vh_{\alpha_i}^{*}$. Assumption \ref{asp: sample criterion} holds for each $\vcL_{i, n}(\vh_{\alpha_i}).$ Assumption \ref{asp: local curvature} holds uniformly for all $\vh_{\alpha_i}^{*}$ for some constant $c_0.$  In addition, there exists a change of measure function for each $\alpha_i$ in Assumption \ref{asp: data coverage}. Then, $\text{Regret}(\piepessi^{\text{RM}}_{R}) = \cO(Mm \cdot \eta_n^{1 - \epsilon_{R}})$  with probability $1-3m\xi.$
    \end{itemize}
    \end{corollary}
    \begin{proof}
        See \S\ref{pro: regret quantile based risk} for a detailed proof.
    \end{proof}
    The results presented in Corollary \ref{cor: regret quantile based risk} align with our intuition: selecting a smaller value for $m$ improves the performance on regret. However, given that $m$ dictates the precision of using \eqref{eq: discrete quantile-based risk measure} to estimate the risk measure $\int_{0}^{1}w(\alpha)\hstaralpha d\alpha$, it essentially represents a trade-off between these two aspects.
\end{example}

\section{Algorithms and Theory in the Negative Control Case}\label{sec: negative controls}
In this section, we  focus on  the scenario where the auxiliary variables are  negative controls. Unless otherwise specified, the notation in this section is consistent with that used previously. Below, we first introduce the identification assumptions and, for the first time in the literature, establish a novel set of integral equations for the quantile treatment effect using negative controls (Theorem \ref{thm: NC conditional moment restrictions}).

\subsection{Causal Identification via Negative Controls}\label{sec: NC causal identification}
During the ODCP, we now observe the negative control exposure (NCE) denoted as $E$ and the negative control outcome (NCO), represented by $V$. The NCE are variables known to not causally affect the reward $Y$. The NCO are variables known to be causally unaffected by either the action $A$ or the NCE $E$. Formally, the following conditions for $V$ and $E$ hold in ODCP:
\begin{assumption}[Negative Controls Assumption]\label{asp: negative controls} For the action $A,$ unmeasured confounder $U$, and the context $X$, 
    \begin{itemize}
        \item[(i)] Latent Unconfoundedness: $V \indep A \given (U, X)$ and $E \indep (Y, V) \given (A, X, U)$;
        \item[(ii)] Completeness Condition: for any $a\in\cA, x\in\cX$,  
        \begin{align*}
            \EE\sbr{\sigma(U, A, X)\given E=e, A=a, X=x}=0
        \end{align*}
        holds for any $e\in\cE$ if and only if $\sigma(U, A, X)\overset{\text{a.s.}}{=} 0$.
        For simplicity, we omit the notation of a.s. in the rest of the paper.
    \end{itemize}
\end{assumption} 
Here, Assumption \ref{asp: negative controls} (i) formalizes the conditional dependence criteria of the negative controls. Assumption \ref{asp: negative controls} (ii) is the completeness condition that implies that $E$ captures the variability of the unmeasured confounders $U$. This condition appears extensively in the NC literature \citep{tchetgen2020introduction}. We can now encode the causal structural quantile function into conditional moment restrictions via  Assumption \ref{asp: negative controls}.
\begin{theorem}[Conditional Moment Restrictions for Negative Controls]\label{thm: NC conditional moment restrictions}
    Suppose that Assumption \ref{asp: negative controls} holds. If there exist bridge functions $\vh^{*}_1: \cA \times \cX$ and $\vh^{*}_2: \cV\times\cA\times\cX\rightarrow \RR$ such that
    \begin{align}
        &\EE\sbr{W(D; \vh^{*}_1) - h^{*}_2(V, A, X)\given E=e, A=a, X=x} = 0,\label{eq: PV ID 1}\\
        &\EE\sbr{h^{*}_2(V, a', X)\given X=x}=0,\label{eq: PV ID 2_}       
    \end{align}
    for any $(e, a, x, a') \in \cE\times\cA\times\cX\times\cA,$ then it follows that
    \begin{align*}
        \prob\sbr{Y \le \vh^{*}_1(X, A) \given X=x, \text{do}(A=a)} = \alpha.
    \end{align*}
    Therefore, the structural quantile function $\vh^{*}_1$ is identified by the conditional moment restrictions \eqref{eq: PV ID 1} and \eqref{eq: PV ID 2_}. We will henceforth refer to $\vh^{*}_1$ and $\hstaralpha$ interchangeably.
\begin{proof}
    See \S\ref{pro: NC conditional moment restrictions} for a detailed proof.
\end{proof}
\end{theorem}

\paragraph{Data augmentation} 
In the conditional moment restriction \eqref{eq: PV ID 2_}, the equality holds pointwise for any $a' \in \cA.$ However, solving this equation becomes computationally intractable if either $|\cA|$ is large or the action space $\cA$ is continuous. To address this issue, we augment the offline dataset by introducing a random variable $A'$ that is distributed uniformly over $\cA$ and is independent of $(X, E, V, A, Y).$ Since
\begin{align*}
    \EE\sbr{h^{*}_2(V, a', X)\given X=x} = \EE\sbr{h^{*}_2(V, A', X)\given A'=a', X=x},
\end{align*}
we can replace \eqref{eq: PV ID 2_} with 
\begin{align}\label{eq: PV ID 2}
    \EE\sbr{h^{*}_2(V, A', X)\given A'=a', X=x}=0. 
\end{align}

In practice, we integrate $A'$ into the offline dataset by appending $a'_i \sim \text{Unif($\cA$)}$ to each sample $(a_i, x_i, e_i, v_i, y_i)$ in ODCP. Consequently, the offline dataset is expanded to $\cbr{a_i, x_i, e_i, v_i, y_i, a'_i}_{i=1}^{n}.$ We can then estimate $h^{*}_2$ by solving \eqref{eq: PV ID 2_} using the augmented offline dataset.

\subsection{Estimation via Causal Identification for Negative Controls}\label{sec: NC loss function}
Denote $h=(h_1, h_2).$ Based on \eqref{eq: PV ID 1} and \eqref{eq: PV ID 2}, we aim to solve simultaneously for $h$ that satisfies:
\begin{align}\label{eq: PV1}
    &\EE\sbr{W(D; h_1) - h_2(V, A, X)\given E, A, X} = 0 \text{ and}\\
    &\EE\sbr{h_2(V, A', X)\given A', X}=0 \label{eq: PV2}.
\end{align}
We can define the RMSE on \eqref{eq: PV1} and \eqref{eq: PV2} with respect to $\vh$ as
\begin{align}\label{def: NC RMSE}
    \nbr{\vT\vh}^2_{2, 2}:= \EE\sbr{ \bigl( \cT_1 \vh(E, A, X) \bigr) ^2} + \EE\sbr{\bigl( \cT_2 \vh(A', X)\bigr)^2}, 
\end{align}
where the operators $\cT_1\vh$ and $\cT_2\vh$ denote 
\begin{align*}
    &\cT_1\vh (\cdot) :=\EE\sbr{W(D; h_1) - h_2(V, A, X)\given (E, A, X)=\cdot},\\
    &\cT_2\vh (\cdot) :=\EE\sbr{h_2(V, A', X)\given (A', X)=\cdot}.
\end{align*}
We then choose the hypothesis function space $\vH=\cH_1\times\cH_2$ and the test function space $\Theta=\Theta_1\times\Theta_2.$  Motivated by the Fenchel duality of the function $x^2/2,$ we can then minimize the empirical loss function $$\cL_n(\vh) := \cL_{1,n}(\vh) + \cL_{2,n}(\vh),$$ where we define $\cL_{1,n}$ and $\cL_{2,n}$ as 
\begin{align}\label{def: NC empirical loss function}
        &\cL_{1,n}\rbr{\vh}:= \sup_{\theta_1\in\Theta_1} \cbr{\frac{1}{n}\sum_{i=1}^{n}\{\sbr{W(D_i; h_1) - h_2(V_i, A_i, X_i)}\theta_1(E_i, A_i, \vX_i)\} - \frac{1}{2n}\sum_{i=1}^{n}\theta_1^2(E_i, A_i, \vX_i)},\\
        &\cL_{2,n}(\vh):= \sup_{\theta_2\in\Theta_2} \cbr{\frac{1}{n}\sum_{i=1}^{n}\sbr{h_2(V_i, A'_i, X_i)}\theta_2(A'_i, \vX_i) - \frac{1}{2n}\sum_{i=1}^{n}\theta_2^2(A'_i, \vX_i)}.
\end{align}
Since both $\cL_{1,n}(\vh)$ and $\cL_{2,n}(\vh)$ can be evaluated from the offline data, we can obtain an estimator of $h_1, h_2$ by minimizing $\cL_n(\vh) = \cL_{1,n}(\vh) + \cL_{2,n}(\vh)$ over $\vH.$
\subsection{Algorithm of Policy Learning via Negative Controls}\label{sec: NC algorithm}
We now present the solution set algorithm for policy learning via NC. We also provide the regularized algorithm in \S\ref{sec: NC reg algorithm}. The underlying concept of the algorithms closely mirrors that of the IV. The primary caveat is that we are now simultaneously solving two conditional moment equations, with our focus primarily on $h_1$, as it represents the estimated structural quantile function of interest. Consequently, this necessitates a minor modification to the definition of the solution set. Apart from this adjustment, the foundational ideas remain consistent.

\paragraph{The solution set algorithm for the negative controls} \label{sec: NC CI algorithm}
We build the solution set for $\vh$ based on the empirical loss function $\cL_n(\vh)$ and the threshold $e_n$:
\begin{align}\label{def: NC CI}
    \cS(e_n) := \cbr{\vh\in\vH:  \cL_{n}(\vh) \le  \inf_{\vh\in\vH}\cL_{n}(\vh) + e_n}.
\end{align}
As we are only interested in $h_1,$ we additionally define the projection of $\cS(e_n)$ onto its first coordinate:
\begin{align*}
    \cS_{1}(e_n) := \cbr{\vh_1\in\vH_1: \exists(\vh_1, \vh_2)\in\cS(e_n)}.
\end{align*}
We then select the policy $\pie$ that optimizes the pessimistic average reward function $v(h_1, \pie)$ over the solution set $\cS_{1}(e_n)$. The full algorithm is summarized in Algorithm \ref{alg: NC meta}.
\begin{algorithm}
    \caption{Quantile action Effect Policy Learning for Negative Controls}\label{alg: NC meta}
    \small
    \begin{algorithmic}
    \REQUIRE Offline dataset $\cbr{a_i, x_i, e_i, v_i, y_i, a'_i}_{i=1}^{n}$ from ODCP, hypothesis space $\vH$, test function space $\Theta,$ and threshold $e_n$.
    \STATE (i) Construct the solution set $\cS(e_n)$ as the level set of $\vH$ with respect to metric $\cL_n(\cdot)$ and threshold $e_n$.
    \STATE (ii) $\piepessi :=\arg\sup_{\pie}\inf_{h_1\in \cS_1(e_n)} v(h_1, \pie)$.
    \ENSURE $\piepessi$.
    \end{algorithmic}
\end{algorithm}

\paragraph{The regularized algorithm for the negative controls}\label{sec: NC reg algorithm}
Similar to the instrumental variables case, the solution set algorithm faces the computational challenges of solving an optimization problem with data-dependent constraints. We introduce a more practical, regularized version of the algorithm.  We denote $\bigeps(h) = \cL_{n}(\vh) -  \inf_{\vh\in\vH}\cL_{n}(\vh).$ The regularized version of the algorithm is summarized in Algorithm \ref{alg: NC regularized version}.
\begin{algorithm}
    \caption{The Regularized Policy Learning Algorithm for Negative Controls}
    \small
    \begin{algorithmic}\label{alg: NC regularized version}
    \REQUIRE Offline dataset $\cbr{a_i, x_i, e_i, v_i, y_i, a'_i}_{i=1}^{n}$ from ODCP, hypothesis space $\vH$, test function space $\Theta,$ and regularization parameter $\lambda_n$.
    \STATE $\piepessi_{R}:=\arg\sup_{\pie}\inf_{h_1\in \vH_1} \cbr{v(h_1, \pie) + \lambda_n\bigeps(h)}$.
    \ENSURE $\piepessi_R$.
    \end{algorithmic}
\end{algorithm}
\subsection{Theoretical Results for Policy Learning Algorithm via the Negative Controls}\label{sec: NC theoretical results}
For NC, we are solving two conditional moment equations simultaneously in contrast to solving one equation in the case of IV. Consequently, the underlying assumptions and theoretical results are analogous to those in Section \ref{sec: theoretical results}. Thus, in this section, we confine our discussion only to the main assumptions and theorems that directly characterize the regret of $\piepessi$ and $\piepessi_{R}.$ See \S\ref{app: theoretical analysis NC} for a complete derivation of the theoretical results. 

A crucial assumption unique to the NC case is Assumption \ref{asp: NC data coverage0}, which parallels the change of measure Assumption \ref{asp: data coverage}. In this instance, we require the existence of two change of measure functions, with each corresponding to one of the two conditional moment restrictions.

\begin{assumption}[Change of Measure for Negative Controls]\label{asp: NC data coverage0}
    For the marginal distribution of context $\tpr$ in the interventional process and the optimal interventional policy $\piestar$, 
       \begin{itemize}
        \item[(i).]There exists a function $b_1: \cE \times \cA \times \cX \rightarrow \RR$ such that $\EE\sbr{b_1^2(E, A, X)} < \infty$ and
        \begin{align}
            \EE\sbr{b_1(E, A, X)p_{y\given E, A, X}(\hstaralpha(A, X))\given A=a, X=x} = \frac{\tpr(x)\piestar(a\given x)}{\pr(x, a)}.\label{cond:NC b1}
        \end{align}
        \item[(ii).] Define $b_2: \cV \times \cE \times \cA \times \cX  \times \cA \rightarrow \RR$ by
        \begin{align}
            b_2(v, e, a, x, a') := \frac{\hpessi{\piestar}_{2}(v, a, x) \pr(v\given e, a, x)}{\hpessi{\piestar}_{2}(v, a', x)\pr(v \given x)}.
            \label{cond:NC b2}
        \end{align}
        We assume $\nbr{b_2}_{\infty} < \infty.$
    \end{itemize}
\end{assumption}

\paragraph{Remark on Assumption \ref{asp: NC data coverage0}} Beyond the change of measure condition outlined in Assumption \ref{asp: data coverage} for the IV, which requires the function of $b$ to have a finite $L_{2}$ norm, Assumption \ref{asp: NC data coverage0} (ii) incorporates more stringent requirements. First, since we now have to solve two conditional moment restrictions, we now need two change of measure functions, $b_1$ and $b_2$. Second, the assumption asserts the change of measure function $b_2$ to have a finite $L_{\infty}$ norm. This is a stricter condition as $L_{\infty}$ is a stronger norm than $L_{2}$. Practically, this implies the necessity for the offline dataset to uniformly cover the covariate space $\cX$ and the negative control outcome space $V$. In addition, to ensure that $\hpessi{\piestar}_{2}(v, a', x)$ is bounded away from $0$, we can do a location shift on the reward.

\begin{theorem}[Informal Version of Theorem \ref{thm: NC CI sub-optimality}]\label{thm: NC CI sub-optimality sketch}
Under appropriate conditions, if the threshold $e_n$ for the solution set is set to $e_n>(2L_h^2 + 5/4)\eta_n^2$, then
    with probability $1 - 5\xi,$ the regret of $\piepessi$ is
    \begin{align*}
        \text{Regret}(\piepessi) = c_1(1+\nbr{b_2}_{\infty})\nbr{b_1}_{2} \cdot \big(\cO\rbr{\sqrt{e_n}} +  \cO\rbr{\eta_n}\big).
    \end{align*}
    \begin{proof}
        See Theorem \ref{thm: NC CI sub-optimality} for the complete statement of the theorem and \S\ref{pro: NC CI suboptimality} for a detailed proof.
    \end{proof}
\end{theorem}

\begin{theorem}[Informal Version of Theorem \ref{thm: NC convergence of sub-optimality reg}]\label{thm: NC convergence of sub-optimality reg sketch}
    Under appropriate conditions, for any $0 < \epsilon_{R} <1,$ if the regularized parameter $\lambda_n$ is set to $\lambda_n = \eta_n^{-(1 + \epsilon_{R})},$ then with probability $1 - 5\xi,$ the regret of $\piepessi_R$ is 
    \begin{align*}
        \text{Regret}(\piepessi_{R}) = \cO\rbr{\eta_n^{1 - \epsilon_{R}}}.
    \end{align*}
    \begin{proof}
        See Theorem \ref{thm: NC convergence of sub-optimality reg} for the complete statement of the theorem and \S\ref{pro: NC regularized suboptimality} for a detailed proof.
    \end{proof}
\end{theorem}

\section{Theoretical Analysis for Algorithm via Negative Controls}\label{app: theoretical analysis NC}

The structure of this section is parallel to that of Section \ref{sec: theoretical results}. The ultimate goal is to link the regret of $\piepessi$ and $\piepessi_R$ to $\nbr{\vT\vh}^2_{2, 2}.$ We first construct two events of high probability that bridge the empirical loss $\vcL_{n}$ to $\nbr{\vT\vh}^2_{2, 2}.$  

\begin{align*}
    \tilde{\cE_1} & := \big\{\abr{\EE_{n}\cbr{\sbr{W(D; h_1) - h_2(V, A, X)}\theta_1(E, A, \vX)}  -  \EE\cbr{\sbr{W(D; h_1) - h_2(V, A, X)}\theta_1(E, A, \vX)} }  \nonumber \\
    &\qquad \qquad \qquad  \le \eta_{n}\left(\nbr{\theta_1}_{2}+\eta_{n}\right),   \quad  \abr{\nbr{\theta_1}^2_{n, 2}-\nbr{\theta_1}^2_{2}}\le \frac{1}{2}\left(\nbr{\theta_1}^2_{2}+\eta_n^2\right), \forall \vh\in \vH, \forall \theta_1\in\Theta_1\big\}. \\
    \tilde{\cE_2} & := \big\{ \abr{\EE_{n}\left\{\sbr{h_2(V, A', X)}\theta_2(V, A', X)\right\} - \EE\left\{\sbr{h_2(V, A', X)}\theta_2(V, A', X)\right\}} \nonumber \\
    &\qquad \qquad \le \eta_{n}\left(\nbr{\theta_2}_{2}+\eta_{n}\right),  \quad \abr{\nbr{\theta_2}^2_{n, 2}-\nbr{\theta_2}^2_{2}}\le \frac{1}{2}\abr{\nbr{\theta_2}^2_{2}+\eta_{n}^2}, \forall \vh\in \vH, \forall \theta_2\in\Theta_2\big\}.
\end{align*}
\begin{assumption}[Regularity of Function Classes for Negative Controls]\label{asp: NC regularity of function classes}
    We assume $\vH$ is compact with respect to the norm $\nbr{\cdot}_{\sup}$ and $\Theta$ is star-shaped. We also assume $\sup_{\vh \in \vH}\nbr{\vh}_{\sup}\le L_H$.
\end{assumption}
 \begin{condition}\label{con: NC high probability event}
    Suppose that Assumption \ref{asp: NC regularity of function classes} holds. For any $\xi>0,$ there exists $\eta_n > 0$ that decreases with $n$ such that the event $\tilde{\cE}=\tilde{\cE_1}\cap\tilde{\cE_2}$ holds with probability at least $1-4\xi.$
\end{condition}
 Note each $\tilde{\cE_i}$ bridges $\cT_i\vh (\cdot)$ to $\nbr{\vT_i\vh}^2_{2}.$ Hence we denote $\tilde{\cE}=\tilde{\cE_1}\cap\tilde{\cE_2},$ so that $\tilde{\cE}$ is an event that controls the difference between the empirical loss function $\vcL_{n}(\vh)$ and $\nbr{\vT\vh}^2_{2, 2}.$ If we choose $\vH_1, \vH_2, \Theta_1$ and $\Theta_2$ to be linear function classes, Condition \ref{con: NC high probability event} holds for some $\eta_n = \tilde{\cO}(n^{-1/2}).$ The reason is that by applying a similar strategy as in \S\ref{app: bracketing concentration}, we can show that each $\tilde{\cE_i}$ holds with probability at least $1 - 2\xi.$ Taking the union bound gives us $1-4\xi.$
We define a norm $\nbr{\cdot}_{\sup}$ on $\vH := \vH_1 \times \vH_2$ as 
$$\nbr{\vh}_{\sup} := \nbr{\vh_1}_{\infty} + \nbr{\vh_2}_{\infty}.$$

\begin{assumption}[Identifiability and Realizability for the Negative Controls]\label{asp: NC identifiability and realizability}  
    Suppose $( h^{*}_1, h^{*}_2) \in \vH$. For any $\vh \in \vH$ that satisfies \eqref{eq: PV1} and \eqref{eq: PV2}, we have $\norm{\vh - h^{*}}_{\sup}=0.$
\end{assumption}
Assumption \ref{asp: NC identifiability and realizability} ensures that the structural quantile function is uniquely identified through the conditional moment restrictions. Consequently, we can substitute $h^{*}_1$ with $\hstaralpha$ without any ambiguity.
\begin{assumption}[Compatibility of Test Function Class]\label{asp: NC compatibility of test function class}
For any $\vh\in\vH,$ $\inf_{\theta\in\Theta} \nbr{\theta - \cT\vh}_{2}=\epsilon_{\Theta},$ and $\epsilon_{\Theta}=\cO(n^{-1/2}).$
\end{assumption}

\subsection{Solution Set Algorithm for the Negative Controls}\label{sec: NC theoretical analysis for the confidence set version algorithm}
We introduce an analogue to Theorem \ref{thm: uncertainty quantification} tailored for NC. This demonstrates that within the event $\tilde{\cE},$ the solution set $\cS_{1}(e_n)$ exhibits several favorable properties.
\begin{theorem}[Uncertainty Quantification for Negative Controls]\label{thm: NC uncertainty quantification}
Suppose that Assumptions \ref{asp: NC identifiability and realizability}, \ref{asp: NC compatibility of test function class}, and \ref{asp: NC regularity of function classes} hold. 
\begin{itemize}
    \item[(i).] On event $\tilde{\cE}$, $\cL_{n}(h^{*}) \le (2L_h^2 + 5/4)\eta_n^2$ where $h^{*} = (\hstaralpha, h_2^{*}).$ Moreover, if we set $e_n>(2L_h^2 + 5/4)\eta_n^2,$ then $\hstaralpha\in\cS_{1}(e_n)$.
    \item[(ii).] On event $\tilde{\cE}$, for all $\vh\in\cS_{1}(e_n)$, we have,
\begin{align*}
    \nbr{\cT\vh}_{2} =  \cO\rbr{\sqrt{e_n}} +  \cO\rbr{\eta_n}.
\end{align*}
\end{itemize}
\begin{proof}
    The proof is identical to the proof of Theorem \ref{thm: uncertainty quantification}, which is in \S\ref{pro: uncertainty quantification}.
\end{proof}
\end{theorem}
\begin{assumption}[Regularity of Density for the Negative Controls]\label{asp: NC regularity of Density}
    We assume that $p_{y\given e, a, x}$, the conditional density of $Y$ given $(E, A, X)$, exists. Moreover, $p_{y\given e, a, x}(y)$ is continuous in $(y, e, a, x)$ and $\sup_{y}p_{y\given e, a, x}(y) < \infty$ for almost all $(y, e, a, x).$
\end{assumption}
    \begin{theorem}[Consistency of Functions in $\cS(e_n)$ for Negative Controls]\label{thm: NC consistency}
    Suppose that Assumptions \ref{asp: NC identifiability and realizability}, \ref{asp: NC compatibility of test function class}, \ref{asp: NC regularity of function classes}, and \ref{asp: NC regularity of Density}  hold. If we set $e_n=\cO(\eta_n^2)$, then on the event $\tilde{\cE},$ for any $\vh \in \cS(e_n),$ $\nbr{\vh_1 - \hstaralpha}_{\infty} = o_{p}\rbr{1}$ and hence $\nbr{\vh_1 - \hstaralpha}_{2} = o_{p}\rbr{1}.$ 
\end{theorem}
\begin{proof}
    Given Lemma \ref{lem: NC continuity of the operator}, the proof is identical to that  of Theorem \ref{thm: consistency}, which is in \S\ref{pro: consistency}.
\end{proof}
Assumption \ref{asp: NC regularity of Density} ensures that the conditional density of the reward $Y$ given $(E, A, X)$ is well-defined and bounded away from infinity. Note we only present the result of consistency of $\vh_1$ in theorem \ref{thm: consistency}. This focus is deliberate, as the upper bound of regret of the NC case is the same as that detailed in Corollary \ref{cor: sub-optimality decomposition}. Thus, the upper bound only concerns the conditional average difference between $\vh_1$ and $\hstaralpha$. Therefore, a local expansion of $\vh_1$ around $\hstaralpha$ suffices to serve our purpose. 

let $\vH_{1 \epsilon} := \left\{ \vh_1\in\vH_1: \nbr{\vh_1-\hstaralpha}_{2} \leq \epsilon\right\}\cap \cS_{1}(e_n)$ where $\epsilon$ is a sufficiently small positive number such that $\prob\rbr{\tilde{\cE} \cap \cbr{\hpessi{\piestar}_1 \in \vH_{1\epsilon}}} \geq 1 - 5\xi.$ Such $\epsilon$ is guaranteed to exist by Condition \ref{con: NC high probability event} and Theorem \ref{thm: NC consistency}.
\begin{assumption}[Local Curvature for the Negative Controls]\label{asp: NC local curvature}
    If we set $e_n>(2L_h^2 + 5/4)\eta_n^2$, then there exists a finite constant $c_1>0$ such that for any $\vh_1 \in \vH_{1 \epsilon},$ 
    $$\norm{\vh_1-\hstaralpha}_{\mathrm{ps}} \leq c_1 \nbr{\EE[W(D; \vh_1)\given E, A, X]}_{2}.$$ 
    Similar to Lemma \ref{lem: directional derivative}, we can show that
      \begin{align*}
        \norm{\vh_1-\hstaralpha}_{\mathrm{ps}} = \sqrt{\EE[(p_{y\given E, A, X}(\hstaralpha(A, X))\cbr{\vh_1(X,A) - \hstaralpha(A, X)})^2]}.
      \end{align*}

\end{assumption}

\begin{theorem}[Regret of Solution Set Algorithm for  Negative Controls]\label{thm: NC CI sub-optimality}
Suppose that Assumption \ref{asp: negative controls} holds. Suppose that Assumptions \ref{asp: NC identifiability and realizability},  \ref{asp: NC compatibility of test function class}, and \ref{asp: NC regularity of function classes} for function classes $\vH$ and $\Theta$ hold. Suppose also that Assumptions \ref{asp: NC local curvature} and \ref{asp: NC data coverage0} hold. If the threshold $e_n$ for the solution set is set to $e_n>(2L_h^2 + 5/4)\eta_n^2$, then conditioning on the event $\tilde{\cE} \cap \cbr{\hpessi{\piestar}_1 \in \vH_{1\epsilon}}$, the regret of $\piepessi$ is 
    \begin{align*}
        \text{Regret}(\piepessi) = c_1(1+\nbr{b_2}_{\infty})\nbr{b_1}_{2} \cdot \big(\cO\rbr{\sqrt{e_n}} +  \cO\rbr{\eta_n}\big).
    \end{align*}
    \begin{proof}
        See \S\ref{pro: NC CI suboptimality} for a detailed proof.
    \end{proof}
\end{theorem}

\subsection{Regularized Algorithm for Negative Controls}\label{sec: NC theoretical analysis regularized version algorithm}
We now detail the theoretical analysis for the regularized algorithm for NC. We define
$\hpessi{\pi}_{R} := (\hpessi{\pi}_{R1}, \hpessi{\pi}_{R2}) :=\arg\inf_{h\in\vH} \cbr{v(h_1, \pie) + \lambda_n \cdot \bigeps(h)}.$
\begin{assumption}[Sample Criterion for Negative Controls]\label{asp: NC sample criterion}
    We assume $\inf_{\vh\in\vH}\vcL_{n}(\vh) = \cO(\eta_n^2).$
\end{assumption}

\begin{theorem}[Consistency of $\hpessi{\pie}_{R}$ for the Negative Controls]\label{thm: NC consistency of the regularized version algorithm}
    Suppose that Assumptions \ref{asp: NC identifiability and realizability}, \ref{asp: NC compatibility of test function class}, \ref{asp: NC regularity of function classes}, \ref{asp: NC regularity of Density} and \ref{asp: NC sample criterion}  hold. For any $\epsilon_{R} > 0$, if we set $\lambda_n \ge \eta_n^{-(1 + \epsilon_{R})},$ then on the event $\tilde{\cE},$ for any $\pie,$ $\|\hpessi{\pi}_{R1} - \hstaralpha\| _{\infty} = o_{p}\rbr{1}$  and hence $\| \hpessi{\pi}_{R1} - \hstaralpha\| _{2} = o_{p}\rbr{1}.$ 
    \begin{proof}
        See \S\ref{pro: NC consistency of the regularized version algorithm} for a detailed proof.
    \end{proof}
\end{theorem}

Given the consistency result, we then perform local expansion of $\hpessi{\piestar}_{R1}$ around $\hstaralpha.$ We now define a restricted space of $\vH$. For a fixed $\lambda_n \ge \eta_n^{-(1 + \epsilon_{R})},$ let
$\vH_{1\epsilon} := \left\{ \vh_1\in\vH_1: \nbr{\vh_1-\hstaralpha}_{2} \leq \epsilon\right\}.$
We then fix a sufficiently small positive number $\epsilon$ such that the event  $\tilde{\cE}_{R\epsilon} = \tilde{\cE} \cap \{\hpessi{\piestar}_{R1} \in \vH_{1\epsilon}\} $ occurs with probability at least $1-5\xi$ by Condition \ref{con: NC high probability event}.

\begin{theorem}[Regret of the Regularized Algorithm for the Negative Controls]\label{thm: NC convergence of sub-optimality reg}
    Suppose that the conditions for negative controls Assumption \ref{asp: negative controls} hold. Suppose that Assumptions \ref{asp: NC identifiability and realizability},  \ref{asp: NC compatibility of test function class}, and 
\ref{asp: NC regularity of function classes} for function classes $\vH$ and $\Theta$ hold. Suppose also that the Assumptions \ref{asp: NC sample criterion}, \ref{asp: NC local curvature}, and \ref{asp: NC data coverage0} hold. For any $0 < \epsilon_{R} <1,$ if the regularized parameter $\lambda_n$ is set to $\lambda_n = \eta_n^{-\rbr{1 + \epsilon_{R}}},$ then conditioning on the event $\tilde{\cE}_{R\epsilon},$ the regret of $\piepessi_R$ is
    \begin{align*}
        \text{Regret}(\piepessi_{R}) = \cO\rbr{\eta_n^{1 - \epsilon_{R}}}.
    \end{align*}
    \begin{proof}
        See \S\ref{pro: NC regularized suboptimality} for a detailed proof.
    \end{proof}
\end{theorem}

\section{Proof of the Main Results of Section \ref{sec: theoretical results}}
In this section, we prove the main results in Section \ref{sec: theoretical results}. The goal is to establish the regret of the solution set algorithm and the regularized algorithm for the IV case.
\subsection{Proof of Theorem \ref{thm: uncertainty quantification}}\label{pro: uncertainty quantification}
Theorem \ref{thm: uncertainty quantification} closely parallels Theorem 6.4 in \citet{chen2023unified}, with adaptations made to suit our specific context. Despite the resemblance of the theorem statement and proof techniques, we nevertheless include a detailed proof to accommodate the shift from a linear operator to a nonlinear operator, which results in several changes in how the final results are presented. In particular, with Assumptions \ref{asp: identifiability and realizability} and \ref{asp: compatibility of test function class}, we prove that $\hstaralpha\in\cS(e_n)$ by deriving an upper bound for $\cL_n(\hstaralpha)$. We then derive the upper bound for $\nbr{\vT\vh}_2$ for any $\vh\in\cS(e_n)$. 

\paragraph{Theorem \ref{thm: uncertainty quantification} (i).}
In this part, we first compute an upper bound of $\cL_n(\hstaralpha)$. We then deduce that $\hstaralpha\in\cS(e_n)$ if $e_n >13/4 \cdot \eta_n^2.$
\begin{proof}
It holds on $\cE$ that
\begin{align*}
    \cL_{n}(\hstaralpha)&=\sup_{\theta\in\Theta}\EE_{n}\sbr{W(D; \hstaralpha) \theta(Z, X)}-\frac 1 2 \nbr{\theta}_{n, 2}^2\nend
    &\le \sup_{\theta\in\Theta} \Big\{\abr{\EE_{n}\sbr{W(D; \hstaralpha) \theta(Z, X)} - \EE\sbr{W(D; \hstaralpha) \theta(Z, X)}}\nend
    &\quad + \frac 1 2 \abr{\nbr{\theta}^2_{n, 2}-\nbr{\theta}^2_{2}}
    + \EE\sbr{W(D; \hstaralpha) \theta(Z, X)} - \frac 1 2 \nbr{\theta}^2_{2}\Big\}\nend
    &\overset{\cE}{\leq} \sup_{\theta\in\Theta} \Big\{\eta_n\rbr{\nbr{\theta}_{n,2}+\eta_n} + \frac 1 4\rbr{\nbr{\theta}^2_{2}+\eta_n^2} \nend 
    &\quad + \EE\sbr{W(D; \hstaralpha) \theta(Z, X)} - \frac 1 2 \nbr{\theta}^2_{2}\Big\}, 
\end{align*}
where the first inequality holds by the triangle inequality and the second inequality holds by the definition of $\cE$. Let $\cL^{\lambda}(\cdot)=\sup_{\theta\in\Theta}\EE\sbr{W(D; \hstaralpha) \theta(Z, X)} - \lambda \nbr{\theta}^2_{2}$. Then $\cL_{n}(\hstaralpha)$ satisfies
\begin{align} \label{eq:3 fr}
   \cL_{n}(\hstaralpha) \le \cL^{1/8}(\hstaralpha) - \inf_{\theta\in\Theta}\rbr{\frac 1 8   \nbr{\theta}^2_{2} - \eta_n\nbr{\theta}_{2}} + \frac 5 4\eta_n^2.
\end{align}
We further let $\theta^\lambda(\cdot;\vh) := \arg\sup_{\theta\in\Theta}\cL^\lambda(\vh)$.
To relate $\cL^{1/8}$ to $\cL^{1/2}$, we observe that
for $\cL^{\lambda_1}(\vh)$ and $\cL^{\lambda_2}(\vh)$ where $0<\lambda_1\le \lambda_2,$ 
\begin{align} \label{eq:4 fr}
    \cL^{\lambda_2}(\vh) &=\sup_{\theta\in\Theta}\EE\sbr{W(D; h) \theta(Z, X)}-\lambda_2 \nbr{\theta}_{2}^2\nend
    &=\frac{\lambda_2}{\lambda_1}\cdot \sup_{\theta\in\Theta}\cbr{\EE\sbr{\frac{\lambda_1}{\lambda_2}W(D; h) \theta(Z, X)}-\lambda_1 \nbr{\theta}_{2}^2}\nend
    &\ge \frac{\lambda_2}{\lambda_1}\cdot \rbr{\EE\sbr{\frac{\lambda_1}{\lambda_2}W(D; h) \cdot \frac{\lambda_1}{\lambda_2}\theta^{\lambda_1}(Z, X;\vh)}-\lambda_1 \nbr{\frac{\lambda_1}{\lambda_2}\theta^{\lambda_1}(Z, X;\vh)}_{2}^2}\nend 
    &\ge \frac{\lambda_1}{\lambda_2} \cL^{ \lambda_1}(\vh),
\end{align}
where the first inequality holds by letting $\theta = \lambda_1/\lambda_2 \cdot \theta^{\lambda_1}.$ Note $\frac{\lambda_1}{\lambda_2}\theta^{\lambda_1}\in\Theta$ as $\lambda_1\le\lambda_2$ and $\Theta$ is star-shaped.
Now substituting $\lambda_1 = 1/8$ and $\lambda_2 = 1/2$ and plugging \eqref{eq:4 fr} into \eqref{eq:3 fr}:
\begin{align} \label{eq:5 fr}
    \cL_{n}(\hstaralpha)
    &\overset{\cE}{\leq} 4\cL^{1/2}(\hstaralpha) - \inf_{\theta\in\Theta}\rbr{\frac 1 8   \nbr{\theta}^2_{2} -\eta_n\nbr{\theta}_{2}} + \frac 5 4 \eta_n^2\nend 
    &\le 4\cL^{1/2}(\hstaralpha) + 
    \frac{13}{4}\eta_n^2, 
\end{align}
where the second inequality holds by computing the minimum value of the quadratic equation of $\nbr{\theta}^2_{2}.$ Also note that
\begin{align*}    \cL^{1/2}(\hstaralpha)&=\sup_{\theta\in\Theta}\cbr{\EE\sbr{W(D; \hstaralpha) \theta(Z)}-\frac 1 2 \nbr{\theta}_{2}^2}\nend
    &\le \sup_{\theta}\cbr{\EE\sbr{W(D; \hstaralpha) \theta(Z)}-\frac 1 2 \nbr{\theta}_{2}^2}\nend
    &=\frac 1 2 \nbr{\cT \hstaralpha}_{2}^2,
\end{align*}
where the second inequality follows by Fenchel duality. Hence \eqref{eq:5 fr} gives
\begin{align} \label{eq:fr 3}
    \cL_n(\hstaralpha)
    &\overset{\cE}{\leq} 2\nbr{\vT\hstaralpha}_{2}^2+ \frac{13}{4}\eta_n^2 = \frac{13}{4}\eta_n^2.
\end{align}
By the non-negativity of $\cL_{n}(\cdot)$ \footnote{See the remark under Assumption \ref{asp: compatibility of test function class}.}, it follows that
\begin{align*}
    \cL_{n}(\hstaralpha) - \inf_{\vh\in\vH}\cL_{n}(\vh) 
    &\overset{\cE}{\leq} \frac {13}{4}\eta_n^2. 
\end{align*}
Therefore, by definition of the solution set in \eqref{def: CI}, with $e_n > 13/4 \cdot \eta_n^2$, it holds on $\cE$ that $\hstaralpha\in\cS(e_n).$
\end{proof}
\paragraph{Theorem \ref{thm: uncertainty quantification} (ii). }
In this part, we derive an upper bound for $\nbr{\cT \vh}_{2}^2$ for any $\vh\in\cS(e_n).$ 

\begin{proof}
We first note it holds for all $\vh\in\cS(e_n)$ that
\begin{align} \label{eq:L_D(h_CI)-LB-1}
    \cL_{n}(\vh) 
    &=\sup_{\theta\in\Theta}\EE_{n}\sbr{W(D; h) \theta(Z, X)}-\frac 1 2 \nbr{\theta}_{n, 2}^2\nend
    &\ge \sup_{\theta\in\Theta} \big\{-\abr{\EE_{n}\sbr{W(D; h) \theta(Z, X)} -\EE\sbr{W(D; h) \theta(Z, X)}} \nend
    &\quad -\frac 1 2 \abr{\nbr{\theta}^2_{n, 2}-\nbr{\theta}^2_{2}} + \EE\sbr{W(D; h) \theta(Z, X)} - \frac 1 2 \nbr{\theta}_{2}^2\big\}\nend
    &\overset{\cE}{\gtrsim} \sup_{\theta\in\Theta} \big\{- \eta_n\rbr{\nbr{\theta}_{2}+\eta_n} - \frac 1 4\rbr{\nbr{\theta}^2_{2}+\eta_n^2}\nend
    &\quad + \EE\sbr{W(D; h) \theta(Z, X)} - \frac 1 2 \nbr{\theta}_{2}^2\big\}.
\end{align}
The first inequality holds by the triangle inequality and the second equality holds by the definition of $\cE$.
Let $\Theta^+(\vh)=\{\theta\in\Theta: \EE\sbr{W(D; h) \theta(X, Z)}>0\}$. For any $\theta^+\in\Theta^+(\vh)$, suppose that $\EE\sbr{W(D; h) \theta^+(X, Z)}=\beta \nbr{\theta^+}_{2}^2$. By definition of $\theta^+$, we have $\beta > 0$.
Let $0<\kappa\le 1$. Since $\Theta$ is star-shaped, we have $\kappa \theta^+\in\Theta$. By plugging in $\kappa \theta^+$ in \eqref{eq:L_D(h_CI)-LB-1},  we have for any $\vh\in\cS(e_n)$ and $\theta^+\in\Theta^+(\vh)$ that
\begin{align}\label{eq:L_D(h_CI)-LB-2}
    \cL_{n}(\vh) 
    &\overset{\cE}{\gtrsim} \kappa\rbr{\beta-\frac 3 4 \kappa}\nbr{\theta^+}_{2}^2 - \eta_n \kappa\nbr{\theta^+}_{2} -  \frac 5 4\eta_n^2.
\end{align}
Recall the definition of the solution set $\cS(e_n)$. For any $\vh\in\cS(e_n)$, it holds that
\begin{align}\label{eq:19}
    \cL_{n}(\vh)
    &\le \inf_{\vh\in\vH}\cL_{n}(\vh) + e_n \le \cL_{n}(\hstaralpha) + e_n \overset{\cE}{\leq} \frac{13}{4}\eta_n^2 + e_n,
\end{align}
where the second inequality holds by noting that $\hstaralpha\in\vH$ and the last inequality follows from \eqref{eq:fr 3}. Combine \eqref{eq:L_D(h_CI)-LB-2} and \eqref{eq:19}, we get
\begin{align}\label{eq:quadratic inequation}
    \kappa\rbr{\beta-\frac 3 4 \kappa}\nbr{\theta^+}_{2}^2 - \eta_n \kappa\nbr{\theta^+}_{2} - \Delta_{n} \overset{\cE}{\leq} 0,
\end{align}
where we define
\begin{align} \label{def:Delta}
    \Delta_{n}= \frac{18}{4}\eta_n^2 + e_n. 
\end{align} 
Note \eqref{eq:quadratic inequation} holds for any $0<\kappa\le 1$.
By setting $\kappa = \min\{1, \beta\}$, we have $\beta- 3/4 \cdot \kappa>0$. Now we solve the quadratic inequality in \eqref{eq:quadratic inequation}, and deduce that on the event $\cE,$ for all $\vh\in\cS(e_n)$ and $\theta^+\in\Theta^+(\vh)$ that
\begin{align} \label{eq:fr 2}
    \nbr{\theta^+}_{2}\overset{\cE}{\leq} \frac{\eta_n \kappa + \sqrt{\rbr{\eta_n \kappa}^2+\kappa\rbr{4\beta-3\kappa}\Delta_{n}}}{\kappa\rbr{2\beta-\frac 3 2 \kappa}}.
\end{align}
Considering the following two cases.
\paragraph{Case (i) when $\beta\ge 1$.} If $\beta \ge 1$, $\kappa=\min\{\beta, 1\}=1$. On $\cE,$  we have 
\begin{align} \label{eq:7 fr}
    \nbr{\theta^+}_{2}^2 &\overset{\cE}{\leq}
    \rbr{\frac{\eta_n+\sqrt{\eta_n^2+\rbr{4\beta-3}\Delta_{n}}}{\rbr{2\beta-\frac 3 2}}}^2\le 4\rbr{\eta_n+\sqrt{\eta_n^2+\Delta_{n}}}^2 \le 8\rbr{2\eta_k^2+\Delta_{n}},
\end{align}
where the first inequality holds by \eqref{eq:fr 2}, the second inequality holds by noting that $\beta=1$ maximizes the right-hand side.
Plugging this result into \eqref{eq:quadratic inequation} and rearranging with $\kappa =1 $, we have 
\begin{align} \label{eq:8 fr}
    \EE\sbr{W(D; h)\theta^+(\vZ)} &\leq \nbr{\theta^+}_{2}^2\nend
    &= \frac{3}{4}\nbr{\theta^+}_{2}^2 +  \frac{1}{4}\nbr{\theta^+}_{2}^2\nend
    &\overset{\cE}{\leq} \frac 3 4 \nbr{\theta^+}_{2}^2 + \eta_n \nbr{\theta^+}_{2} + \Delta_{n}\nend
    &\le 6\rbr{2\eta_k^2+\Delta_{n}} + \eta_n\rbr{4\eta_n+ 3\sqrt{\Delta_{n}}} + \Delta_{n}\nend
    &\le 16\eta_n^2+ 7\Delta_{n} + 3\eta_n\sqrt{\Delta_{n}}\nend
    &\le 20\eta_n^2 + 11\Delta_{n}, 
\end{align}
where the first inequality holds by noting that $\kappa(\beta- 3/4 \cdot \kappa)= 1/4$ and the third inequality holds by the upper bound of $\nbr{\theta^+}_{2}$ in \eqref{eq:7 fr}.
\paragraph{Case (ii) when $\beta<1$. } If $\beta < 1$, we plug in $\kappa=\beta$. It holds on $\cE$ that
\begin{align*}
    \beta \nbr{\theta^+}_{2} \overset{\cE}{\leq} 2\rbr{\eta_n + \sqrt{\rbr{\eta_n \kappa}^2+\Delta_{n}}}\le 2\rbr{2\eta_n + \sqrt{\Delta_{n}}},
\end{align*}
which suggests that 
\begin{align} \label{eq:9 fr}
    \EE\sbr{W(D; h) \theta^+(\vX, \vZ)} =  \beta \nbr{\theta^+}_{2}^2 \overset{\cE}{\leq} 2\rbr{2\eta_n + \sqrt{\Delta_{n}}} \nbr{\theta^+}_{2}.
\end{align}
\paragraph{Combination of Case (i) and Case (ii).}
Combining \eqref{eq:8 fr} and \eqref{eq:9 fr}, we then have for any $\vh\in\cS(e_n)$ and $\theta_k^+\in \Theta_k^+(\vh)$ that
\begin{align*}
    \EE\sbr{W(D; h) \theta^+(\vX, \vZ)} &\overset{\cE}{\leq} \max\cbr{2\rbr{\sqrt{\Delta_{n}}+2\eta_n} \nbr{\theta}_{2}, 20\eta_n^2+ 11\Delta_{n}}\nend
    &\leq\max\cbr{C_n \nbr{\theta}_{2}, C_n^2},
\end{align*}
where $C_n = 5 \eta_n+ 4\sqrt{\Delta_{n}}$.
We then consider the case in which $\EE\sbr{W(D; h) \theta(\vX, \vZ)}<0$ for $\theta\in\Theta\backslash \Theta^+(\vh)$. Then, for any $\vh\in\cS(e_n)$, $\theta\in\Theta$,
\begin{align} \label{eq:fr 1}
    \EE\sbr{W(D; h) \theta(\vX, \vZ)} \overset{\cE}{\leq} \max\cbr{C_n \nbr{\theta}_{2}, C_n^2}.
\end{align}
Now let $\theta^{*}_{\vh}(\vX, \vZ)=\arg\min_{\theta\in\Theta} \nbr{\theta-\cT\vh}_{2}$. By \eqref{eq:fr 1}, it then holds for any $h\in\cS(e_n)$ that
\begin{align} \label{eq:fr 5}
    \EE\sbr{\cT\vh(X, Z) \theta^{*}_{\vh}(\vX, \vZ)} &\overset{\cE}{\leq}
    \max\cbr{C_n \nbr{\theta^{*}_{\vh}}_{2}, C_n^2}.
\end{align}
For the left-hand side of \eqref{eq:fr 5}, we have
\begin{align} \label{eq:fr 6}
    \EE\sbr{\cT\vh(X, Z)\theta^{*}_{\vh}(\vX, \vZ)}
    &=\EE\sbr{\cT\vh(X, Z)\rbr{\theta^{*}_{\vh}(\vX, \vZ)-\cT\vh(X, Z)+\cT\vh(X, Z)}}\nend
    &\ge \nbr{\cT\vh}_{2}^2 - \nbr{\cT\vh}_{2}\nbr{\theta^{*}_{\vh}-\cT\vh}_{2} = \nbr{\cT\vh}_{2}^2 + \cO(\eta_n) , 
\end{align}
where the first inequality holds by the Cauchy-Schwarz inequality and the last equality holds by Assumption \ref{asp: compatibility of test function class}.
For the right-hand side of \eqref{eq:fr 5}, we have,
\begin{align} \label{eq:fr 7}
    \max\cbr{C_n \nbr{\theta^{*}_{\vh}(\vX, \vZ)}_{2}, C_n^2} 
    \le C_n \rbr{C_n + \nbr{\theta^{*}_{\vh}-\cT\vh+\cT\vh}_{2}} \le C_n\rbr{C_n+\nbr{\cT\vh}_{2}},  
\end{align}
where the last inequality holds by the triangle inequality and Assumption \ref{asp: compatibility of test function class}.
Combining \eqref{eq:fr 6} and \eqref{eq:fr 7} with \eqref{eq:fr 5}, for all $\vh\in\cS(e_n),$ we have:
\begin{align*}
    \nbr{\cT\vh}_{2}^2 - C_n^2 - C_n\nbr{\cT\vh}_{2} \overset{\cE}{\leq}0, 
\end{align*}
which implies that
\begin{align*}
    \nbr{\cT\vh}_{2} &\overset{\cE}{\leq} \frac 1 2 \rbr{C_n + \sqrt{C_n^2+4C_n^2}}= \cO(C_n)\nend
    &=  \cO\rbr{5\eta_n + 4\sqrt{18/4 \cdot \eta_n^2 + e_n}}\nend
    & = \cO\rbr{\sqrt{e_n}} + \cO\rbr{\eta_n} , 
\end{align*}
where the third equality holds by the definition of $C_n$ and the definition of $\Delta_{n}$ in \eqref{def:Delta}. 
\end{proof}

\subsection{Proof of Theorem \ref{thm: consistency}}\label{pro: consistency}
We prove that, on event $\cE$, any $\vh\in \cS(e_n)$ converges in probability  to $\hstaralpha$ with respect to $\nbr{\cdot}_{\infty}$ and hence $\nbr{\cdot}_{2},$ when $e_n=\cO(\eta_n^2).$
\begin{proof}
 We first show that it converges in probability in $\nbr{\cdot}_{\infty}$ by using two lemmas that investigate the property of the loss function $\cL_{n}.$ Lemma \ref{lem: lower bound of the empirical loss} below gives a lower bound of the empirical loss on event $\cE.$ Lemma \ref{lem: continuity of the linear operator} shows that the RMSE $\nbr{\cT\vh}_2^2$ is continuous on $(\vH, \nbr{\cdot}_{\infty}).$
 \begin{lemma}[Lower Bound of  Empirical Loss]\label{lem: lower bound of the empirical loss}
Suppose that Assumptions \ref{asp: regularity of function classes} and \ref{asp: compatibility of test function class} hold. On the event $\cE$, for all $\vh  \in \vH,$ we have
\begin{align*}
  \cL_n(\vh)=\sup_{\theta\in\Theta}\EE_n\sbr{W(D; h) \theta(\vX, \vZ)}-\frac 1 2 \nbr{\theta}_{n, 2}^2 \geq \min\cbr{\frac{\eta_n}{2}\nbr{\vT\vh}_{2} \nbr{\vT\vh}^2_{2}}-\cO\rbr{\eta_n^2}.
\end{align*}
\end{lemma}
\begin{proof}
    See  \S\ref{pro: IV loss lower bound} for a detailed proof.
\end{proof}
\begin{lemma}[Continuity of Linear Operator]\label{lem: continuity of the linear operator}
Suppose the Assumption \ref{asp: regularity of Density} holds. $\nbr{\cT\vh}_2^2$ is continuous on $(\vH, \nbr{\cdot}_{\infty}).$
\end{lemma}
\begin{proof}
    See \S\ref{pro: IV continuity} for a detailed proof.
\end{proof}
 
 We are ready to prove consistency. For any $\epsilon > 0,$ we have 
 \begin{align*}
    & \prob\sbr{{\nbr{\vh - \hstaralpha}}_{\infty} > \epsilon, \vh \in \cS(e_n)} \notag\\
    &\qquad \leq \prob\sbr{\inf_{{\vh \in \vH},{\nbr{\vh - \hstaralpha}}_{\infty} > \epsilon}\cL_{n}(\vh)\leq \cO(\eta_n^2)}\nend
    &\qquad \leq \prob\sbr{ \min \cbr{\inf_{{\vh \in \vH},{\nbr{\vh - \hstaralpha}}_{\infty} > \epsilon}\frac{\eta_n}{2}\nbr{\vT\vh}_{{2}}, \inf_{{\vh \in \vH},{\nbr{\vh - \hstaralpha}}_{\infty} > \epsilon}\nbr{\vT\vh}^2_{2}} \leq \cO\rbr{\eta_n^2}}, 
 \end{align*}
 where the second inequality holds by Lemma \ref{lem: lower bound of the empirical loss}. Denote $\varphi_{\epsilon} := \inf_{{\vh \in \vH},{\nbr{\vh - \hstaralpha}}_{\infty}> \epsilon}\nbr{\vT\vh}_{2}.$  Assumption \ref{asp: identifiability and realizability}, \ref{asp: regularity of function classes}, and Lemma \ref{lem: continuity of the linear operator} implies that $\vH$ is compact and $\nbr{\vT\vh}_{{2}}$ is continuous on  $(\vH, \nbr{\cdot }_{\infty}).$ Hence $\varphi_{\epsilon}$ is strictly positive. We then have  
\begin{align*}
   \prob\sbr{{\nbr{\vh - \hstaralpha}}_{\infty} > \epsilon, \vh \in \cS(e_n)} &\leq \prob\sbr{ \min \left\{\frac{\eta_n}{2}\varphi_{\epsilon}, \varphi_{\epsilon}^2\right\} \leq \cO\rbr{\eta_n^2}} ,
 \end{align*}
 which converges to zero as $n$ goes to infinity.
Since the supremum norm is stronger than the $L_2$ norm under a finite measure:
\begin{align*}
   \prob\sbr{{\nbr{\vh - \hstaralpha}}_{2} > \epsilon, \vh \in \cS(e_n)} \leq \prob\sbr{{\nbr{\vh - \hstaralpha}}_{\infty} > \epsilon, \vh \in \cS(e_n)},
\end{align*}
we have $\prob\sbr{{\nbr{\vh - \hstaralpha}}_{2} > \epsilon, \vh \in \cS(e_n)}$ converges to 0 as $n$ goes to infinity. That is, 
\begin{align*}
  \lim_{n \rightarrow \infty} \prob\sbr{{\nbr{\vh - \hstaralpha}}_{2} > \epsilon, \vh \in \cS(e_n)} = 0.
\end{align*}
Thus, we conclude the proof.
\end{proof}
 
\subsection{Proof of Theorem \ref{thm: consistency of the estimator of the regularized version algorithm}}\label{pro: consistency of the estimator of the regularized version algorithm}
We prove that if we set $\lambda_n \ge \eta_n^{-\rbr{1 + \epsilon_{R}}}$ where $\epsilon_{R}$ is an arbitrary positive real number, then on the event $\cE$, $\hpessi{\pie}_R$ converges in probability to $\hstaralpha$ with respect to $\nbr{\cdot}_{\infty}$ and hence $\nbr{\cdot}_{2}.$

\begin{proof}
   We use a similar argument as in the proof of Theorem \ref{thm: consistency}. But we first need to quantify an upper bound of the empirical loss of the regularized estimator: 
\begin{lemma}[Upper Bound of the Empirical Loss of Regularized Estimator]\label{lem: upper bound of the empirical loss of regularized version estimator}
   Suppose that Assumption \ref{asp: identifiability and realizability}, Assumption \ref{asp: regularity of function classes}, and Assumption \ref{asp: sample criterion} hold. If we set $\lambda_n \geq \eta_n^{-(1 + \epsilon_{R})}$ where $0 < \epsilon_{R} <1$ is arbitrary, then on the event $\cE$, $\bigeps(\hpessi{\pie}_R) = \cO\rbr{\eta_n^{1 + \epsilon_{R}}}$ and $\cL_n(\hpessi{\pie}_R) = \cO\rbr{\eta_n^{1 + \epsilon_{R}}}.$
   \begin{proof}
       See \S\ref{pro: upper bound of the empirical loss of regularized version estimator} for a detailed proof.
   \end{proof}
\end{lemma}
   
Therefore, for any $\epsilon > 0,$ we have
  \begin{align*}
      & \prob\sbr{{\nbr{\hpessi{\pie}_R - \hstaralpha}}_{\infty} > \epsilon} \notag\\
      &\qquad \leq \prob\sbr{\inf_{{\vh \in \vH},{\nbr{\vh - \hstaralpha}}_{\infty} > \epsilon}\cL_{n}(\vh)\leq \cO\rbr{\eta_n^{1 + \epsilon_{R}}}}\nend
      &\qquad \leq \prob\sbr{ \min \left\{\inf_{{\vh \in \vH},{\nbr{\vh - \hstaralpha}}_{\infty} > \epsilon}\frac{\eta_n}{2}\nbr{\vT\vh}_{{2}}, \inf_{{\vh \in \vH},{\nbr{\vh - \hstaralpha}}_{\infty} > \epsilon}\nbr{\vT\vh}^2_{2}\right\} \leq \cO\rbr{\eta_n^{1 + \epsilon_{R}}}}, 
   \end{align*}
  where the first inequality holds by Lemma \ref{lem: upper bound of the empirical loss of regularized version estimator}, and the second inequality holds by Lemma \ref{lem: lower bound of the empirical loss}. Denote $\varphi_{\epsilon} := \inf_{{\vh \in \vH},{\nbr{\vh - \hstaralpha}}_{\infty}> \epsilon}\nbr{\vT\vh}_{2}.$  Assumption \ref{asp: identifiability and realizability}, \ref{asp: regularity of function classes}, and Lemma \ref{lem: continuity of the linear operator} implies that $\vH$ is compact and $\nbr{\vT\vh}_{{2}}$ is continuous on  $(\vH, \nbr{\cdot }_{\infty}).$ Hence $\varphi_{\epsilon}$ is strictly positive. We then have  
  \begin{align*}
   \prob\sbr{{\nbr{\hpessi{\pie}_R - \hstaralpha}}_{\infty} > \epsilon} &\leq \prob\sbr{ \min \left\{\frac{\eta_n}{2}\varphi_{\epsilon}, \varphi_{\epsilon}^2\right\} \leq \cO\rbr{\eta_n^{1 + \epsilon_{R}}}},
   \end{align*}
   which converges to zero as $n$ goes to infinity.
   Since the supremum norm is stronger than the $L_2$ norm under a finite measure:
  \begin{align*}
     \prob\sbr{{\nbr{\hpessi{\pie}_R - \hstaralpha}}_{2} > \epsilon} \leq \prob\sbr{{\nbr{\hpessi{\pie}_R - \hstaralpha}}_{\infty} > \epsilon},
  \end{align*}
  we have $\prob\sbr{{\nbr{\hpessi{\pie}_R - \hstaralpha}}_{2} > \epsilon}$ converges to 0 as $n$ goes to infinity:
   \begin{align*}
     \lim_{n \rightarrow \infty} \prob\sbr{{\nbr{\hpessi{\pie}_R - \hstaralpha}}_{2} > \epsilon} =  0.
  \end{align*}
  Thus, we conclude the proof.
\end{proof}

\subsection{Decomposition of the Regret with Pessimism \& Proof of Corollary \ref{cor: sub-optimality decomposition}}\label{proof: pessimism}
In this section, we study the regret of the estimated policy $\piepessi$ with pessimism.
The result in this section will be utilized in \S\ref{proof: subopt} to derive the regret.
Recall the definition of $\hpessi{\pi}$ and $\piepessi$:
\begin{align} \label{eq: hpessi}
    \hpessi{\pi} := \arg\inf_{h\in\cS(e_n)} v(h, \pi),
    \piepessi := \arg\sup_{\pie} \inf_{h\in\cS(e_n)} v(h, \pi) =  \arg\sup_{\pie} v(\hpessi{\pie}, \pi).
\end{align}
The regret of policy $\piepessi$ is bounded by
\begin{align} \label{eq: regret}
    \textrm{Regret}(\piepessi) &=v^{\piestar}_{\alpha}-v^{\piepessi}_{\alpha} \nend
    &= \underbrace{v^{\piestar}_{\alpha} - v(\hpessi{\piestar}, \piestar)}_{\displaystyle \text{(i)}} + \underbrace{v(\hpessi{\piestar}, \piestar) - v(\hpessi{\piepessi}, \piepessi)}_{\displaystyle\textrm{(ii)}} + \underbrace{v(\hpessi{\piepessi}, \piepessi) - v^{\piepessi}_{\alpha}}_{\displaystyle \text{(iii)}}, 
\end{align}
Here, $\text{(ii)}\le 0$ holds by the optimality of $\piepessi$ and $\text{(iii)}\overset{\cE}{\leq} 0$ holds by the definition of $\hpessi{\piepessi}$ in \eqref{eq: hpessi} and the fact that $\hstaralpha\in\cS(e_n)$ on event $\cE$ by Theorem \ref{thm: uncertainty quantification}. 
Therefore, we just need to bound (i). This proves Corollary \ref{cor: sub-optimality decomposition}.

We now establish an upper bound for (i) as follows.  

\subsection{Proof of Theorem \ref{thm: convergence of sub-optimality}} \label{proof: subopt}
We upper bound the regret of the solution set algorithm in the context of IV. By Corollary \ref{cor: sub-optimality decomposition}, we can focus on the estimation error of the average reward function for the optimal interventional policy $\piestar,$ i.e., Term (i) in \eqref{eq: regret}. 
\begin{proof}
By definition of the change of measure function (Assumption \ref{asp: data coverage}), we have:
\begin{align*}
    v^{\piestar}_{\alpha} - v(\hpessi{\piestar}, \piestar)
    &=\EE_{\pin{\piestar}}\sbr{\hstaralpha(A, X)-\hpessi{\piestar}(A, X)}\nend
    &=\EE\sbr{(\hstaralpha(A, X)-\hpessi{\piestar}(A, X)) \cdot \EE\sbr{b(X, Z)\pr_{Y\given A, X, Z}(\hstaralpha(A, X))\given A, X}}.
\end{align*}
Then  by the tower property,
\begin{align*}
    v^{\piestar}_{\alpha} - v(\hpessi{\piestar}, \piestar) 
    &= \EE\sbr{\EE\sbr{(\hstaralpha(A, X)-\hpessi{\piestar}(A, X))b(X, Z)\pr_{Y\given A, X, Z}(\hstaralpha(A, X))\given A, X}}\nend
    &= \EE[\pr_{Y\given A, X, Z}(\hstaralpha(A, X))(\hstaralpha(A, X)-\hpessi{\piestar}(A, X))b(X, Z)]\nend
    &= \EE\sbr{\EE\sbr{p_{Y\given A, X, Z}(\hstaralpha(A, X))(\hstaralpha(A, X)-\hpessi{\piestar}(A, X))\given X, Z}b(X, Z)}\nend
    & \leq \norm{\hpessi{\piestar}-\hstaralpha}_{\mathrm{ps}} \cdot \nbr{b}_{2},
\end{align*}
 where the last inequality holds by the Cauchy-Schwarz inequality. Denote $\cE_{\epsilon}:=\cE \cap \cbr{\hpessi{\piestar} \in \vH_{\epsilon}}.$
Thus we can bound the regret as
\begin{align*}
    \text{Regret}(\piepessi) 
    &\overset{\cE_{\epsilon}}{\leq} \norm{\hpessi{\piestar}-\hstaralpha}_{\mathrm{ps}} \cdot \nbr{b}_{2} \le c_0\cdot  \nbr{\vT\hpessi{\piestar}}_{2} \cdot \nbr{b}_{2} = c_0\cdot \nbr{b}_{2} \cdot [\cO\rbr{\sqrt{e_n}} +  \cO\rbr{\eta_n}],
\end{align*}
where the first inequality holds by the regret decomposition in \S\ref{proof: pessimism}, the second inequality holds by Assumption \ref{asp: local curvature}, and the last inequality holds by the Theorem \ref{thm: uncertainty quantification}. Hence, we complete the proof of Theorem \ref{thm: convergence of sub-optimality}.
\end{proof}

\subsection{Proof of Theorem \ref{thm: convergence of sub-optimality reg} and  Lemma \ref{cor: sub-optimality decomposition reg}} \label{proof: convergence of sub-optimality reg}
In this section, we prove that if the regularized parameter $\lambda_n$ is set to $\lambda_n = \eta_n^{-(1 + \epsilon_{R})}$, then on the event $\cE$ ,the regret of the regularized algorithm is of order $\eta_n^{1 - \epsilon_{R}}$.
\begin{proof}
    We first decompose the regret corresponding to $\piepessi_R$ as:
    \begin{align*}
        \text{Regret}(\piepessi_R) &=v^{\piestar}_{\alpha}-v^{\piepessi_R}_{\alpha} \nend
        &= \sbr{v^{\piestar}_{\alpha} + \lambda_n\bigeps\rbr{\hstaralpha}} - \sbr{v(\hpessi{\piepessi_{R}}_R, \piepessi_R)+ \lambda_n\bigeps(\hpessi{\piepessi_{R}}_R)} + \sbr{v(\hpessi{\piepessi_{R}}_R, \piepessi_R)+ \lambda_n\bigeps(\hpessi{\piepessi_{R}}_R)} \\ &\qquad- \sbr{v\rbr{\hstaralpha, \piepessi_R}+ \lambda_n\bigeps\rbr{\hstaralpha}}.
    \end{align*}
By the optimality of $\hpessi{\piepessi}_R$, we have
\begin{align*} 
    \text{Regret}(\piepessi_R) &\leq \sbr{v^{\piestar}_{\alpha} + \lambda_n\bigeps\rbr{\hstaralpha}} - \sbr{v(\hpessi{\piepessi_{R}}_R, \piepessi_R)+ \lambda_n\bigeps(\hpessi{\piepessi_{R}}_R)}\nend
    &\leq \sbr{v^{\piestar}_{\alpha} + \lambda_n\bigeps\rbr{\hstaralpha}} - \sup_{\pie}\inf_{h\in \vH} \{v\rbr{h, \pie} + \lambda_n\bigeps(h)\}\nend
    &\leq \sbr{v^{\piestar}_{\alpha} + \lambda_n\bigeps(\hstaralpha)} - \sbr{v(\hpessi{\piestar}_R, \piestar)+ \lambda_n\bigeps(\hpessi{\piestar}_R)}.
\end{align*}
After rearranging the terms, we get
\begin{align} \label{eq:subopt reg}
    \text{Regret}(\piepessi_R) &= \EE_{\pin{\piestar}}\sbr{\hstaralpha(A, X)-\hpessi{\piestar}_R(A, X)} + \lambda_n\bigeps(\hstaralpha) -  \lambda_n\bigeps(\hpessi{\piestar}_R)\nend
    &\leq \EE_{\pin{\piestar}}\sbr{\hstaralpha(A, X)-\hpessi{\piestar}_R(A, X)} + \lambda_n\bigeps\rbr{\hstaralpha}.
\end{align}
This concludes the proof of Lemma \ref{cor: sub-optimality decomposition reg}.

Following the same argument as in Section \ref{proof: subopt}, we conclude that for the first term,
\begin{align} \label{eq:upper bound of the first term of subopt reg}
\EE_{\pin{\piestar}}\sbr{\hstaralpha(A, X)-\hpessi{\piestar}_R(A, X)} \overset{\cE_{R\epsilon}}{\leq} c_0\nbr{b}_{2}\cdot \nbr{\vT\hpessi{\piestar}_R}_{2}. 
\end{align}
For the second term, we have by Theorem \ref{thm: uncertainty quantification} (ii) that
\begin{align} \label{eq:upper bound second term subopt reg}
    \lambda\bigeps(\hstaralpha) \leq  \lambda \cL_{n}(\hstaralpha) \leq \cO(\eta_n^{1 - \epsilon_{R}}).
\end{align} 
Moreover, using the fact that $\text{Regret}(\piepessi_R) \geq 0$ and rearranging terms, we have
\begin{align} \label{eq:lower bound of the first term of subopt reg}
    \EE_{\pin{\piestar}}\sbr{\hstaralpha(A, X)-\hpessi{\piestar}_R(A, X)} &\geq \lambda_n \bigeps(\hpessi{\piestar}_R) -  \lambda_n\bigeps\rbr{\hstaralpha}\nend
    &\overset{\cE_{R\epsilon}}{\gtrsim} \lambda_n \cL_n(\hpessi{\piestar}_R) - \lambda_n \cO(\eta_n^2)\nend
    &\geq \lambda_n \min\cbr{\frac{\eta_n}{2}\nbr{\vT\hpessi{\piestar}_R}_{2}, \nbr{\vT\hpessi{\piestar}_R}^2_{2}}- \lambda_n \cO(\eta_n^2).
\end{align}
We now show that $\nbr{\vT\hpessi{\piestar}_R}_{2} = \cO(\eta_n).$ 
When $\eta_n/{2} \cdot \nbr{\vT\hpessi{\piestar}_R}_{2} \geq \nbr{\vT\hpessi{\piestar}_R}^2_{2},$ the result is immediate. When $\eta_n/{2} \cdot \nbr{\vT\hpessi{\piestar}_R}_{2} < \nbr{\vT\hpessi{\piestar}_R}^2_{2},$ combining \eqref{eq:upper bound of the first term of subopt reg} and \eqref{eq:lower bound of the first term of subopt reg}, we have
\begin{align*}
    c_0\nbr{b}_{2}\cdot \nbr{\vT\hpessi{\piestar}_R}_{2} &\overset{\cE_{R\epsilon}}{\gtrsim} \lambda_n \cdot \frac{\eta_n}{2}\nbr{\vT\hpessi{\piestar}_R}_{2}- \lambda_n \cO(\eta_n^2).
\end{align*}
Solving for $\nbr{\vT\hpessi{\piestar}_R}_{2}$ and substituting $\lambda_n = \eta_n^{-(1 + \epsilon_{R})}$ yields 
\begin{align*}
    \nbr{\vT\hpessi{\piestar}_R}_{2} = \cO(\eta_n).
\end{align*}
Substituting this into \eqref{eq:upper bound of the first term of subopt reg}, we know the first term of \eqref{eq:subopt reg} is upper bounded by $\cO(\eta_n)$. By \eqref{eq:upper bound second term subopt reg}, we know the second term of  \eqref{eq:subopt reg} is upper bounded by $\cO(\eta_n^{1 - \epsilon_{R}}).$ By summing up the two terms, we conclude that conditioning on the event $\cE_{R\epsilon}$,
\begin{align*}
    \text{Regret}(\piepessi_R) = \cO\rbr{\eta_n^{1 - \epsilon_{R}}}.
\end{align*}
Therefore, we complete the proof.
\end{proof}

\subsection{Proof of Corollary \ref{cor: regret quantile based risk}} \label{pro: regret quantile based risk}
We present the regret for minimizing the quantile-based risk measures. The idea is to aggregate the individual regret for each $\alpha_i.$
\begin{proof}
For each $i \in [m]$, we  define an event  $\cE_i$ as  
\begin{align}\label{def: Event}
    \cE_i := \Big\{& \abr{\EE_{n}\sbr{W_i(D; h_{\alpha_i}) \theta(\vX, \vZ)} - \EE\sbr{W_i(D; h_{\alpha_i}) \theta(\vX, \vZ)}} \le \eta_n\rbr{\nbr{\theta}_{2}+\eta_n},\nend
    &\qquad \abr{\nbr{\theta}^2_{n, 2}-\nbr{\theta}^2_{2}}\le \frac 1 2\rbr{\nbr{\theta}^2_{2}+\eta_n^2}, 
    \forall \vh_{\alpha_i}\in \vH, \forall \theta\in\Theta\Big\},
\end{align}
and then define $\cE' = \bigcap_{i=1}^{m}\cE_i.$ So $\cE'$ holds with probability at least $1 - 2m\xi$.
We can upper bound the regret of $\piepessi^{\text{RM}}$ as:
\begin{align*}
    \text{Regret}(\piepessi^{\text{RM}}) &= \EE_{\pin{\piestar}^{\text{RM}}}\sbr{\sum_{i=1}^{m}\phi(\alpha_i)\vh^{*}_{\alpha_i}(A, X)} - \EE_{p_{\text{in}}{\piepessi^{\text{RM}}}}\sbr{\sum_{i=1}^{m}\phi(\alpha_i)h_{\alpha_i}(A, X)} \nend
    &\overset{\cE'}{\leq} \EE_{\pin{\piestar}^{\text{RM}}}\sbr{\sum_{i=1}^{m}\phi(\alpha_i)(\vh^{*}_{\alpha_i}(A, X) - \hpessi{\piestar}_{\alpha_i}(A, X))} \nend
    & \leq M\sum_{i=1}^{m} \EE_{\pin{\piestar}}\sbr{\vh^{*}_{\alpha_i}(A, X) - \hpessi{\piestar}_{\alpha_i}(A, X))}.
\end{align*}
For the regularized algorithm, we have a similar upper bound:
\begin{align*}
    \text{Regret}\rbr{\piepessi^{\text{RM}}_{R}} &\leq \EE_{\pin{\piestar}^{\text{RM}}}\sbr{\sum_{i=1}^{m}\phi(\alpha_i)(\vh^{*}_{\alpha_i}(A, X) - \hpessi{\piestar}_{R \alpha_i}(A, X))} + \lambda_n(\bigeps(\vh^{*}) - \bigeps(\hpessi{\piestar}_R) \nend
    &\leq M\sum_{i=1}^{m} \EE_{\pin{\pie}}\sbr{\vh^{*}_{\alpha_i}(A, X) - \hpessi{\piestar}_{R \alpha_i}(A, X)} + \lambda_n(\bigeps(\vh^{*}) - \bigeps(\hpessi{\piestar}_R).
\end{align*}
The rest of the proof now follows from applying the proof in Theorem \ref{thm: convergence of sub-optimality} and Theorem \ref{thm: convergence of sub-optimality reg} to each $i.$ 
\end{proof}

\section{Proof of Main Results of \S\ref{sec: negative controls}}\label{app:NC main}
In this section, we give the proofs of the main theorems in \S\ref{sec: negative controls}. The goal is to establish the regret for the policy learning algorithms in the case that we observe NC in ODCP.
\subsection{Proof of Theorem \ref{thm: NC conditional moment restrictions}} \label{pro: NC conditional moment restrictions}
We show that given Assumption \ref{asp: negative controls}, $\vh^{*}_1$ and $h^{*}_2$ satisfy the conditional moment restrictions given in Theorem \ref{thm: NC conditional moment restrictions}. 
\begin{proof}
    From \eqref{eq: PV ID 1}, we have 
    \begin{align*}
        0 &= \EE\sbr{W(D; \vh^{*}_1) - h^{*}_2(V, A, X)\given E=e, A=a, X=x}\nend
          &= \EE\sbr{\EE\sbr{W(D; \vh^{*}_1) - h^{*}_2(V, A, X)\given U, E, A, X}\given E=e, A=a, X=x}\nend
          &= \EE\sbr{\EE\sbr{W(D; \vh^{*}_1) - h^{*}_2(V, A, X)\given U, A, X}\given E=e, A=a, X=x},
    \end{align*}
    where the last equality holds by the Assumption \ref{asp: negative controls} (i). Now by Assumption \ref{asp: negative controls} (ii), we have 
    \begin{align}
        \EE\sbr{W(D; \vh^{*}_1)\given U, A, X} = \EE\sbr{h^{*}_2(V, A, X)\given U, A, X}. \label{eq: PV proof 1}
    \end{align} 
    From \eqref{eq: PV ID 2_}, $\forall a' \in \cA,$ $\forall a \in \cA,$ we have
    \begin{align}
        0 &= \EE\sbr{h^{*}_2(V, a', X)\given X=x} = \EE\sbr{\EE\sbr{h^{*}_2(V, a', X)\given U, X}\given X=x}\nend
          &= \EE\sbr{\EE\sbr{h^{*}_2(V, a', X)\given U, A=a, X}\given X=x} \label{eq: PV proof 2},
    \end{align}
    where the third equality holds by Assumption \ref{asp: negative controls} (i). Now by combining \eqref{eq: PV proof 1} and \eqref{eq: PV proof 2}, we have
    \begin{align*}
        0 &= \EE\sbr{\EE\sbr{W(D; \vh^{*}_1)\given U, A=a, X}\given X=x}= \EE\sbr{W(D; \vh^{*}_1)\given do(A=a), X=x}.
    \end{align*}
    By the definition of $W(D; \vh^{*}_1),$ we conclude that 
    \begin{align*}
        \prob\sbr{Y \le \vh^{*}_1 \given X=x, \text{do}(A=a)} = \alpha.
    \end{align*}
    Therefore, we complete the proof.
\end{proof}

\subsection{Proof of Theorem \ref{thm: NC CI sub-optimality}} \label{pro: NC CI suboptimality}
In this section, we analyze the regret of the solution set algorithm for the negative controls. 
\begin{proof}
We first note that the regret decomposition for the case of negative control is exactly the same as that of the instrumental variable. Therefore, we begin by considering the term (i) in \eqref{eq: regret}. By the change of measure Assumption \ref{asp: NC data coverage0}, we have:
        \begin{align*}
            v^{\piestar}_{\alpha} - v(\hpessi{\piestar}_1, \piestar)
            &=\EE_{\pin{\piestar}}\sbr{\hstaralpha(A, X)-\hpessi{\piestar}_{1}(A, X)}\nend
            &=\EE\sbr{(\hstaralpha(A, X)-\hpessi{\piestar}(A, X)_{1})\EE\sbr{b_1(E, A, X)p_{y\given E, A, X}(\hstaralpha(A, X))\given A, X}}.
        \end{align*}
Then by Tower property,
        \begin{align*}
       v^{\piestar}_{\alpha} - v(\hpessi{\piestar}_1, \piestar) &= \EE\sbr{\EE\sbr{(\hstaralpha(A, X)-\hpessi{\piestar}_{1}(A, X))b_1(E, A, X )p_{y\given E, A, X}(\hstaralpha(A, X))\given A, X}}\nend
            &= \EE[p_{y\given E, A, X}(\hstaralpha(A, X))(\hstaralpha(A, X)-\hpessi{\piestar}_{1}(A, X))b_1(E, A, X)]\nend
            &= \EE\sbr{\EE\sbr{p_{y\given E, A, X}(\hstaralpha(A, X))(\hstaralpha(A, X)-\hpessi{\piestar}_{1}(A, X))\given E, A, X}b_1(E, A, X)}.
        \end{align*}
Now let $\tilde{\cE}_{\epsilon}:= \tilde{\cE} \cap \cbr{\hpessi{\piestar}_1 \in \vH_{1\epsilon}}.$ By Cauchy-Schwarz and the form of directional derivative in Lemma \ref{lem: directional derivative} (with a slight change of the conditional variables),  we have
            \begin{align}
                \text{Regret}(\piepessi) 
                &\overset{\tilde{\cE}_{\epsilon}}{\leq} \norm{\hpessi{\piestar}_{1}-\hstaralpha}_{\mathrm{ps}} \cdot \nbr{b_1}_{2}\nend
                &\le c_1 \nbr{b_1}_{2} \cdot \nbr{\EE[W(D; \hpessi{\piestar}_{1})\given E, A, X]}_{2}, \label{eq:NC CI proof 1}
            \end{align}
where the last inequality holds by Assumption \ref{asp: NC local curvature}. We now obtain an upper bound for $\nbr{\EE[W\rbr{D; \hpessi{\piestar}_{1}}\given E, A, X]}_{2}:$ 
        \begin{align*}
            \nbr{\EE[W\rbr{D; \hpessi{\piestar}}\given E, A, X]}_{2} &= \nbr{\vT_1\hpessi{\piestar} + \EE\sbr{\hpessi{\piestar}_{2}(W, A, X)\given E, A, X}}_{2}\nend
            &\leq \nbr{\vT_1\hpessi{\piestar}}_{2} + \nbr{\EE\sbr{\hpessi{\piestar}_{2}(W, A, X)\given E, A, X}}_{2}\nend
            &= \nbr{\vT_1\hpessi{\piestar}}_{2} + \nbr{\int_{\cW} \hpessi{\piestar}_{2}(w, A, X)p(w\given E, A, X)\rd w}_{2}.
        \end{align*}
        By Assumption \ref{asp: NC data coverage0} (ii), we have  
        \begin{align*}
         \nbr{\EE[W\rbr{D; \hpessi{\piestar}}\given E, A, X]}_{2} &\leq \nbr{\vT_1\hpessi{\piestar}}_{2} + \nbr{\int_{\cW} \hpessi{\piestar}_{2}(w, A', X)p(w\given X)b_2(w, E, A, X, A')\rd w}_{2}\nend
            &\leq \nbr{\cT_1\hpessi{\piestar}}_{2} + \nbr{b_2}_{\infty}\nbr{\int_{\cW} \hpessi{\piestar}_{2}(w, A', X)\pob(w\given A', X)\rd w}_{2}\nend
            &= \nbr{\vT_1\hpessi{\piestar}}_{2} + \nbr{b_2}_{\infty}\nbr{\vT_2\hpessi{\piestar}}_{2}.
        \end{align*}
        Substitute the above result back to \eqref{eq:NC CI proof 1}, we complete the proof:
        \begin{align*}
            \text{Regret}(\piepessi) 
            &\overset{\tilde{\cE}_{\epsilon}}{\leq} c_1 \nbr{b_1}_{2} \cdot \nbr{\vT_1\hpessi{\piestar}}_{2} + c_1 \nbr{b_1}_{2} \cdot \nbr{b_2}_{\infty}\nbr{\vT_2\hpessi{\piestar}}_{2}\nend
            &= c_1(1+\nbr{b_2}_{\infty})\nbr{b_1}_{2} \cdot \big(\cO\rbr{\sqrt{e_n}} +  \cO\rbr{\eta_n}\big),
        \end{align*}
        where the equality holds by Theorem \ref{thm: NC uncertainty quantification} (ii).
\end{proof}

\subsection{Proof of Theorem \ref{thm: NC consistency of the regularized version algorithm}} \label{pro: NC consistency of the regularized version algorithm}
We show that our estimated structural quantile function $\hpessi{\pi}_{R1}$ is consistent to $\hstaralpha$ in the $\ell_2$-norm. The proof is similar to the proof of Theorem \ref{thm: consistency of the estimator of the regularized version algorithm}, except that the dimension of $\hpessi{\pi}_{R}$ is now two. We deduce the result by first showing that $\hpessi{\pi}_{R} = (\hpessi{\pi}_{R1}, \hpessi{\pi}_{R2})$ is consistent to $h^{*} = \rbr{\hstaralpha, {\vh_2}^{*}}$ in $\nbr{\cdot}_{\sup}.$  
\begin{proof}
    For any $\epsilon > 0,$ we have
       \begin{align*}
           \prob\sbr{{\nbr{\hpessi{\pi}_{R} - h^{*}}}_{\sup} > \epsilon} &\leq \prob\sbr{\inf_{{\vh \in \vH},{\nbr{\vh - h^{*}}}_{\sup} > \epsilon}\cL_{1, n}(\vh)\leq \cO(\eta_n^{1 + \epsilon_{R}})}\nend
           &\leq \prob\sbr{ \min \cbr{\inf_{{\vh \in \vH},{\nbr{\vh - h^{*}}}_{\sup} > \epsilon}\frac{\eta_n}{2}\nbr{\vT_1\vh}_{{2}}, \inf_{{\vh \in \vH},{\nbr{\vh - h^{*}}}_{\sup} > \epsilon}\nbr{\vT_1\vh}^2_{2}} \leq \cO\rbr{\eta_n^{1 + \epsilon_{R}}}}.
        \end{align*}
       Denote $\varphi_{\epsilon} := \inf_{{\vh \in \vH},{\nbr{\vh - h^{*}}}_{\infty}> \epsilon}\nbr{\vT_1\vh}_{2}.$  Since Lemma \ref{lem: NC continuity of the operator} states that $\nbr{\vT_1\vh}_{{2}}$ is continuous on  $(\vH, \nbr{\cdot }_{\sup}),$ $\varphi_{\epsilon}$ is strictly positive. We then have  
       \begin{align*}
        \prob\sbr{{\norm{\hpessi{\pie}_{R} - h^{*}}}_{\sup} > \epsilon} &\leq \prob\sbr{ \min \left\{\frac{\eta_n}{2}\varphi_{\epsilon}, \varphi_{\epsilon}^2\right\} \leq \cO(\eta_n^{1 + \epsilon_{R}})} ,
        \end{align*}
        which converges to 0 as $n$ goes to infinity.
        The result follows from the fact that the supremum norm is stronger than the $L_2$ norm under a finite measure:
       \begin{align*}
          \prob\sbr{{\nbr{\hpessi{\pie}_{R} - h^{*}}}_{2} > \epsilon} \leq \prob\sbr{{\nbr{\hpessi{\pie}_{R} - h^{*}}}_{\infty} > \epsilon},
       \end{align*}
       which converges to 0 as $n$ goes to infinity. Hence, we have $\| \hpessi{\pi}_{R1} - \hstaralpha\|  _{\infty} = o_{p}\rbr{1}$  and  $\| \hpessi{\pi}_{R1} - \hstaralpha\| _{2} = o_{p}\rbr{1}.$ 
\end{proof}

\subsection{Proof of Theorem \ref{thm: NC convergence of sub-optimality reg}} \label{pro: NC regularized suboptimality}
In this section, we analyze the regret of the regularized algorithm for NC. The proof is similar to the proof of Theorem \ref{thm: convergence of sub-optimality reg}.
\begin{proof}
    We first decompose the regret corresponding to $\piepessi_R$ as:
    \begin{align*}
            \text{Regret}(\piepessi_R) &=v^{\piestar}_{\alpha}-v^{\piepessi_R}_{\alpha} \nend
        &= \sbr{v^{\piestar}_{\alpha} + \lambda_n\bigeps\rbr{h^{*}}} - \sbr{v(\hpessi{\piepessi_{R}}_{R1}, \piepessi_R)+ \lambda_n\bigeps(\hpessi{\piepessi_{R}}_R)} + \sbr{v(\hpessi{\piepessi_{R}}_{R1}, \piepessi_R)+ \lambda_n\bigeps(\hpessi{\piepessi_{R}}_R)} \\ &\qquad- \sbr{v\rbr{\hstaralpha, \piepessi_R}+ \lambda_n\bigeps\rbr{h^{*}}}.
    \end{align*}
By the optimality of $\hpessi{\piepessi}_R$, we can drop the last two terms. It follows that
\begin{align*}
    \text{Regret}(\piepessi_R) &\leq \sbr{v^{\piestar}_{\alpha} + \lambda_n\bigeps(h^{*})} - \sbr{v(\hpessi{\piepessi_{R}}_{R1}, \piepessi_R)+ \lambda_n\bigeps(\hpessi{\piepessi_{R}}_R)}\nend
    &\leq \sbr{v^{\piestar}_{\alpha} + \lambda_n\bigeps(h^{*})} - \sup_{\pie}\inf_{h\in \vH} \{v(h_1, \pie) + \lambda_n\bigeps(h)\}\nend
    &\leq \sbr{v^{\piestar}_{\alpha} + \lambda_n\bigeps(h^{*})} - \sbr{v(\hpessi{\piestar}_{R1}, \piestar)+ \lambda_n\bigeps(\hpessi{\piestar}_R)}.
\end{align*}
Then by rearranging terms, we have
\begin{align}\label{eq:NC subopt reg}
    \text{Regret}(\piepessi_R)&= \EE_{\pin{\piestar}}\sbr{\hstaralpha(A, X)-\hpessi{\piestar}_{R1}(A, X)} + \lambda_n\bigeps(h^{*}) -  \lambda_n\bigeps(\hpessi{\piestar}_R)\nend
    &\leq \EE_{\pin{\piestar}}\sbr{\hstaralpha(A, X)-\hpessi{\piestar}_{R1}(A, X)} + \lambda_n\bigeps(h^{*}).
\end{align}
Following the same argument as in the proof of Theorem \ref{thm: NC CI sub-optimality}, we conclude that for the first term,
\begin{align}\label{eq:NC upper bound of the first term of subopt reg}
\EE_{\pin{\piestar}}\sbr{\hstaralpha(A, X)-\hpessi{\piestar}_{R1}(A, X)} \overset{\tilde{\cE}_{R\epsilon}}{\leq} c_1 \nbr{b_1}_{2} \cdot \nbr{\vT_1\hpessi{\piestar}}_{2} + c_1 \nbr{b_1}_{2} \cdot \nbr{b_2}_{\infty}\nbr{\vT_2\hpessi{\piestar}}_{2} .
\end{align}
For the second term, we have by Theorem \ref{thm: NC uncertainty quantification} (ii) that
\begin{align}\label{eq:NC upper bound second term subopt reg}
    \lambda_n\bigeps(h^{*}) \leq  \lambda_n \cL_{n}(h^{*}) \leq \cO(\eta_n^{1 - \epsilon_{R}}).
\end{align} 
Moreover, using the fact that $\text{Regret}(\piepessi_R) \geq 0$ and rearranging terms, we have
\begin{align*}
    \EE_{\pin{\piestar}}\sbr{\hstaralpha(A, X)-\hpessi{\piestar}_{R1}(A, X)} &\geq \lambda_n \bigeps(\hpessi{\piestar}_R) -  \lambda_n\bigeps(h^{*}) \overset{\tilde{\cE}_{\epsilon}}{\geq} \lambda_n \cL_n(\hpessi{\piestar}_R) - \lambda_n \cO(\eta_n^2).
\end{align*}
We can further lower bound it as
\begin{align}\label{eq:NC lower bound of the first term of subopt reg}
    \lambda_n \min \Bigl \{\frac{\eta_n}{2}\nbr{\vT_1\hpessi{\piestar}_R}_{2}, \nbr{\vT_1\hpessi{\piestar}_R}^2_{2}\Bigr\} + \lambda_n \min\Big \{\frac{\eta_n}{2}\nbr{\vT_2\hpessi{\piestar}_R}_{2}, \nbr{\vT_2\hpessi{\piestar}_R}^2_2\Big \} - \lambda_n \cO(\eta_n^2). 
\end{align}
Then by following a similar step as in the proof of Theorem \ref{thm: convergence of sub-optimality reg}, it follows that
\begin{align*}
    \nbr{\vT_1\hpessi{\piestar}}_{2} + \nbr{\vT_2\hpessi{\piestar}}_{2} = \cO(\eta_n).
\end{align*}
Then combining this result with \eqref{eq:NC subopt reg}, \eqref{eq:NC upper bound of the first term of subopt reg} and \eqref{eq:NC lower bound of the first term of subopt reg}, we conclude that conditioning on the event $\tilde{\cE}_{R\epsilon},$
\begin{align*}
    \text{Regret}(\piepessi_R) = \cO(\eta_n^{1 - \epsilon_{R}}).
\end{align*}
Therefore, we complete the proof.
\end{proof}

\section{Supporting Lemmas for Section \ref{sec: theoretical results}}\label{app: technical}
In what follows, we present the statement and proofs of the supporting lemmas used in §\ref{sec: theoretical results}. 

\begin{lemma}[Directional Derivative]\label{lem: directional derivative}
Suppose that Assumption \ref{asp: model} and Assumption \ref{asp: local curvature} hold. Then for any $\vh \in \vH,$
\begin{align*}
    \frac{d\cT\hstaralpha}{d\vh}[\vh-\vh^{*}_{\alpha}] = \EE[p_{Y|{A,X, Z}}(h^{*}_{\alpha}(A, X))\cbr{\vh(X,A) - h^{*}_{\alpha}(A, X)}\given X, Z].
\end{align*}
Hence $\norm{\vh-\hstaralpha}_{\mathrm{ps}} = \sqrt{\EE[(\EE[p_{Y|{A, X, Z}}(h^{*}_{\alpha}(A, X))\cbr{\vh(X,A) - h^{*}_{\alpha}(A, X)}\given X, Z])^2]}.$
\begin{proof}
Recall that $W(D; h^{*}_{\alpha})=\mathbb{1}\{Y \leq h^{*}_{\alpha}(A, X)\} - \alpha.$ Assumption \ref{asp: local curvature} assures that the conditional density $p_{Y\given A, X, Z}$ exists almost surely. Then 

\begin{align*}
    \frac{d\cT\hstaralpha}{d\vh}[\vh - \vh^{*}_{\alpha}] 
    &= \frac{d\EE[W(D;(1-r)\vh^{*}_{\alpha}(A, X) + r\vh(A, X)) \given X, Z]}{dr}\Given_{r=0}\nend
    &= \frac{d\EE[\mathbb{1}\{Y \leq h^{*}_{\alpha}(A, X) + r[\vh(A, X) - h^{*}_{\alpha}(A, X)]\} - \alpha \given X, Z]}{dr}\Given_{r=0}\nend 
    &= \frac{d\mathbb{P}(Y \leq h^{*}_{\alpha}(A, X))+ r(\vh(A, X) - h^{*}_{\alpha}(A, X)\given X, Z)}{dr}\Given_{r=0}  .
\end{align*}
By Assumption \ref{asp: local curvature}, we can write the conditional probability as the integral of the conditional density $p_{Y\given A, X, Z}$:
\begin{align}\label{eq: directional derivative}
    \frac{d\cT\hstaralpha}{d\vh}[\vh - \vh^{*}_{\alpha}] 
    &= \frac{d}{dr}\EE[\int_{-\infty}^{h^{*}_{\alpha}(A, X) + r(\vh(A, X) - h^{*}_{\alpha}(A, X))}p_{Y\given A, X, Z}(y)dy \given X, Z]\Given_{r=0}\nend
    &= \EE[p_{Y\given A, X, Z}(h^{*}_{\alpha}(A, X) + r[\vh - h^{*}_{\alpha}(A, X)])\cbr{\vh(X,A) - h^{*}_{\alpha}(A, X)}\given X, Z]\Given_{r=0}\nend
    &= \EE[p_{Y\given A, X, Z}(h^{*}_{\alpha}(A, X))\cbr{\vh(A, X) - h^{*}_{\alpha}(A, X)}\given X, Z].
\end{align}
Hence $\norm{\vh-\hstaralpha}_{\mathrm{ps}} = \sqrt{\EE[(\EE[p_{Y\given A, X, Z}(h^{*}_{\alpha}(A, X))\cbr{\vh(X,A) - h^{*}_{\alpha}(A, X)}\given X, Z])^2]}$ by the definition of the pseudometric.
\end{proof}
\end{lemma}

\paragraph{Proof of Lemma \ref{lem: lower bound of the empirical loss}}
\begin{proof} \label{pro: IV loss lower bound}
Recall that on the event $\cE,$ $\abr{\EE_{n}\sbr{W(D; h) \theta(\vX, \vZ)} - \EE\sbr{W(D; h) \theta(\vX, \vZ)}} \le \eta_n\rbr{\nbr{\theta}_{2}+\eta_n} \text{and} \nend
    \quad \abr{\nbr{\theta}^2_{n, 2}-\nbr{\theta}^2_{2}}\le \frac 1 2\rbr{\nbr{\theta}^2_{2}+\eta_n^2}.$  Note we define $\theta^{*}_{\vh}(\vX, \vZ)$ as $\arg\min_{\theta\in\Theta} \nbr{\theta-\cT\vh}_{2}.$ We consider the following two cases:

\paragraph{Case (i) where $\nbr{\theta^{*}_{\vh}}_{2}\ge \eta_n$.} Let $r = \eta_n/(2\nbr{\theta^{*}_{\vh}}_{2}).$ As $r \in [0, 1/2]$ and $\Theta$ is star-shaped, $r\theta^{*}_{\vh} \in \Theta.$ Thus 
\begin{align*}
    \cL_n(\vh) \ge \EE_n\sbr{W(D; h) r\theta^{*}_{\vh}(\vX, \vZ)} - \frac{1}{2}\nbr{r\theta^{*}_{\vh}}^2_{n, 2}.
\end{align*}
On the event $\cE,$ for the second term on the right-hand side, 
\begin{align*}
    \frac{1}{2}\nbr{r\theta^{*}_{\vh}}^2_{n, 2} &\leq \frac{r^2}{2}\sbr{\frac{3}{2}\nbr{\theta^{*}_{\vh}}^2_{2} + \eta_n^2} \leq \sbr{\eta_n^2/(4\nbr{\theta^{*}_{\vh}}^2_{2})} \cdot \frac{1}{2}\sbr{\frac{3}{2}\nbr{\theta^{*}_{\vh}}^2_{2} + \eta_n^2}  =  \cO(\eta_n^2).
\end{align*}
For the first term regarding the empirical norm of the $\theta^{*}_{\vh}$,
\begin{align*}
    \EE_n\sbr{W(D; h) r\theta^{*}_{\vh}(\vX, \vZ)} &\geq \EE\sbr{W(D; h) r\theta^{*}_{\vh}(\vX, \vZ)} - \eta_n(\nbr{r\theta^{*}_{\vh}}_{2} + \eta_n)\nend
    &\geq r\EE\sbr{W(D; h)\theta^{*}_{\vh}(\vX, \vZ)}-\cO(\eta_n^2)\nend
    &= r\EE\sbr{\cT\vh(X, Z) \theta^{*}_{\vh}(\vX, \vZ)} -\cO(\eta_n^2).
\end{align*}
By adding and subtracting $\theta^{*}_{\vh}(\vX, \vZ),$ we further have
\begin{align*}
    \EE_n\sbr{W(D; h) r\theta^{*}_{\vh}(\vX, \vZ)} &\geq r\EE\sbr{(\cT\vh(X, Z) - \theta^{*}_{\vh}(\vX, \vZ) + \theta^{*}_{\vh}(\vX, \vZ))\theta^{*}_{\vh}(\vX, \vZ)}-\cO(\eta_n^2)\nend
    &\geq \frac{\eta_n}{2}\nbr{\theta^{*}_{\vh}}_{2}-\cO(\eta_n^2) \geq \frac{\eta_n}{2}\nbr{\cT\vh}_{2}-\cO(\eta_n^2),
\end{align*}
where the second inequality follows by Assumption \ref{asp: compatibility of test function class}. Combining both terms, we have
\begin{align*}
\cL_n(\vh)=\sup_{\theta\in\Theta}\EE_n\sbr{W(D; h) \theta(\vX, \vZ)}-\frac 1 2 \nbr{\theta}_{n, 2}^2 \geq \frac{\eta_n}{2}\nbr{\vT\vh}_{2} -\cO(\eta_n^2).
\end{align*}
\paragraph{Case (ii) where $\nbr{\theta^{*}_{\vh}}_{2}< \eta_n$.} We simply choose $r=1:$
\begin{align*}
    \cL_n(\vh) \ge \EE_n\sbr{W(D; h) \theta^{*}_{\vh}(\vX, \vZ)} - \frac{1}{2}\nbr{\theta^{*}_{\vh}}^2_{n, 2}.
\end{align*}
We can upper bound the second term by
\begin{align*}
    \frac{1}{2}\nbr{\theta^{*}_{\vh}}^2_{n, 2} &\leq \frac{3}{2}\nbr{\theta^{*}_{\vh}}^2_{2} + \eta_n^2 \leq \cO(\eta_n^2).
\end{align*}
Thus the first term can be lower bounded by
\begin{align*}
    \EE_n\sbr{W(D; h) \theta^{*}_{\vh}(\vX, \vZ)} &\geq \EE\sbr{W(D; h) \theta^{*}_{\vh}(\vX, \vZ)} - \eta_n(\nbr{\theta^{*}_{\vh}}_{2} + \eta_n)\nend
    &\geq \EE\sbr{W(D; h)\theta^{*}_{\vh}(\vX, \vZ)}-\cO(\eta_n^2).
\end{align*}
By adding and subtracting $\theta^{*}_{\vh}(\vX, \vZ),$ we have
\begin{align*}
    &= \EE\sbr{(\cT\vh(X, Z) - \theta^{*}_{\vh}(\vX, \vZ) + \theta^{*}_{\vh}(\vX, \vZ))\theta^{*}_{\vh}(\vX, \vZ)}-\cO(\eta_n^2)\nend
    &\geq \nbr{\theta^{*}_{\vh}}_{2}^2-\cO(\eta_n^2)\nend
    &\geq \nbr{\cT\vh}_{2}^2-\cO(\eta_n^2).
\end{align*}
We then complete the proof by combining the above two cases.
\end{proof}

\paragraph{Proof of Lemma \ref{lem: upper bound of the empirical loss of regularized version estimator}}
\begin{proof} \label{pro: upper bound of the empirical loss of regularized version estimator}
By definition of $\hpessi{\pie}_R,$ we have
\begin{align*}
    v(\hpessi{\pie}_R, \pie) + \lambda_n\bigeps(\hpessi{\pie}_R) &= \inf_{h\in\vH} \{v(h, \pi) + \lambda_n\bigeps(\vh)\} \leq v(\hstaralpha, \pie) + \lambda_n\bigeps(\hstaralpha).
\end{align*}
The inequality holds as $\hstaralpha \in \vH$ from Assumption \ref{asp: identifiability and realizability}. After rearranging the terms and noting that by Theorem \ref{thm: uncertainty quantification}(ii), $\bigeps(\hstaralpha) \le \cL_n(\hstaralpha) \le 13/4 \cdot \eta_n^2,$  we have
\begin{align*}
    \bigeps(\hpessi{\pie}_R) &\leq \frac{1}{\lambda_n}\sbr{v(\hstaralpha, \pie) - v(\hpessi{\pie}_R, \pie)} + \bigeps(\hstaralpha) \leq \frac{2L_h}{\lambda_n} + \frac{13}{4}\eta_n^2 \leq 2L_h \eta_n^{1 + \epsilon_{R}} + \frac{13}{4}\eta_n^2.
\end{align*}
Then by Assumption \ref{asp: sample criterion}, we have $\cL_n(\hpessi{\pie}_R) = \bigeps(\hpessi{\pie}_R) + \inf_{\vh\in\vH}\vcL_{n}(\vh) = \cO\rbr{\eta_n^{1 + \epsilon_{R}}}.$
\end{proof}

\paragraph{Proof of Lemma \ref{lem: continuity of the linear operator}}
\begin{proof} \label{pro: IV continuity}
Recall the definition of $\cT$:
\begin{align*}
    \cT \vh (X, Z) &:=\EE\sbr{W(D; h)\given X, Z}= \EE\sbr{\mathbb{1}\{Y \leq \vh(X,A)\}-\alpha\given X, Z}\nend
    &=\EE\sbr{\int_{-\infty}^{\vh(A, X)}\pr_{Y|{A, X, Z}}(y)dy \given X, Z} - \alpha.
\end{align*}
Thus, for any $\vh_1, \vh_2 \in \vH,$ by the mean value theorem and Assumption \ref{asp: regularity of Density} we have 
\begin{align*}
    |\cT\vh_1 - \cT\vh_2| &\leq \EE\sbr{\sup_{t \in [0, 1]}\pr_{Y|{A, X, Z}}(\vh_1(X, A) + t\sbr{\vh_2(X, A) - \vh_1(X, A)})\sbr{\vh_2(X, A) - \vh_1(X, A)} \given X, Z}\nend
    &\leq \EE\sbr{\sup_{t \in [0, 1]}\pr_{Y|{A, X, Z}}(\vh_1(X, A) + t\sbr{\vh_2(X, A) - \vh_1(X, A)}) \given X, Z} \sbr{\sup_{y}|\vh_2(y) - \vh_1(y)|}.
\end{align*}
Then, by observing that $\sup_{(X,Z) \in \cX \times \cZ, h \in \vH}|\cT\vh|\leq 1,$ we have
\begin{align*}
    \nbr{\cT\vh_1}_2^2 - \nbr{\cT\vh_2}_2^2 &\leq 2\EE\sbr{|\cT\vh_2 - \cT\vh_1|}\nend
    &\leq 2\EE\sbr{\sup_{t \in [0, 1]}\pr_{Y|{A, X, Z}}(\vh_1(X, A) + t\sbr{\vh_2(X, A) - \vh_1(X, A)}) \given X, Z} \sbr{\sup_{y}|\vh_2(y) - \vh_1(y)|}\nend
    &\leq 2\EE\sbr{\sup_{t \in [0, 1]}\pr_{Y|{A, X, Z}}(\vh_1(X, A) + t\sbr{\vh_2(X, A) - \vh_1(X, A)}) \given X, Z} \nbr{\vh_2 - \vh_1}_{\infty}.
\end{align*}
Assumption \ref{asp: regularity of Density} now implies the continuity result.
\end{proof}

\section{Supporting Lemmas for \S\ref{sec: NC theoretical results}} \label{app:NC tech}
In what follows, we present the statement and proofs of the supporting lemmas used in \S\ref{sec: NC theoretical results}. 
\begin{lemma}[Upper Bound of the Empirical Loss of Negative Controls]\label{lem: NC upper bound of the empirical loss of regularized version estimator}
    If we set $\lambda_n \geq \eta_n^{-(1 + \epsilon_{R})}$ where $0 < \epsilon_{R} <1$ is arbitrary, then on the event $\tilde{\cE}$, $\bigeps(\hpessi{\pie}_R) = \cO(\eta_n^{1 + \epsilon_{R}})$ and $\cL_n(\hpessi{\pie}_R) = \cO(\eta_n^{1 + \epsilon_{R}}).$
    
    \begin{proof}
    By definition of $\hpessi{\pie}_R,$ we have
    \begin{align*}
        v(\hpessi{\pie}_{R1}, \pie) + \lambda_n\bigeps(\hpessi{\pie}_R) &= \inf_{h\in\vH} \{v(h_1, \pi) + \lambda_n\bigeps(\vh)\} \leq v(\hstaralpha, \pie) + \lambda_n\bigeps(h^{*}).
    \end{align*}
    After rearranging the terms and note that by Theorem \ref{thm: uncertainty quantification}(ii), $\bigeps(h^{*}) \le \cL_n(h^{*}) \le (2L_h^2 + 5/4)\eta_n^2,$  we have
    \begin{align*}
        \bigeps(\hpessi{\pie}_R)&\leq \frac{1}{\lambda_n}\sbr{v(\hstaralpha, \pie) - v(\hpessi{\pie}_{R1}, \pie)} + \bigeps(h^{*})\leq \frac{2L_h}{\lambda_n} + \frac{13}{4}\eta_n^2\nend
        &\leq 2L_h \eta_n^{1 + \epsilon_{R}} + \left(2L_h^2 + \frac{5}{4}\right)\eta_n^2.
    \end{align*}
    Then by Assumption \ref{asp: NC sample criterion}, we have $\cL_n(\hpessi{\pie}_R) = \cO\rbr{\eta_n^{1 + \epsilon_{R}}}.$
\end{proof}
\end{lemma}

\begin{lemma}[Lower Bound of the Empirical Loss for Negative Controls]\label{NC lower bound of the empirical loss}
    Suppose that Assumptions \ref{asp: NC regularity of function classes} and \ref{asp: NC compatibility of test function class} hold. On the event $\tilde{\cE}$, for all $\vh  \in \vH,$ $k=0, 1$ we have 
    \begin{align*}
      \cL_{k, n}(\vh) \geq \min\cbr{\frac{\eta_n}{2}\nbr{\vT_k\vh}_{2}, \nbr{\vT_k\vh}^2_{2}}-\cO(\eta_n^2).
    \end{align*}
    \begin{proof}
        This lemma is a direct variation of Lemma \ref{lem: lower bound of the empirical loss}. The proof is almost the same.
    \end{proof}
\end{lemma}

\begin{lemma}[Continuity of the Operator for Negative Controls]\label{lem: NC continuity of the operator}     
$\nbr{\cT_1\vh}_2^2$ is continuous on $(\vH, \nbr{\cdot}_{\sup}).$
\begin{proof}
    Recall that $\cT_1\vh:=\EE\sbr{W(D; h_1) - h_2(V, A, X)\given (E, A, X)}.$ Thus, for any $\vh_1 = \rbr{\vh^{(1)}_1, \vh^{(2)}_1}, \vh_2 = \rbr{\vh^{(1)}_2, \vh^{(2)}_2} \in \vH,$ we have
    \begin{align}
        \abr{\cT_1\vh_1 - \cT_1\vh_2} &=\abr{\EE\sbr{W(D; \vh^{(1)}_1) - \vh^{(2)}_1(V, A, X)\given (E, A, X)} - \EE\sbr{W(D; \vh^{(1)}_2) - \vh^{(2)}_2(V, A, X)\given (E, A, X)}}\nend
        &\leq \abr{\EE\sbr{W(D; \vh^{(1)}_1) - W(D; \vh^{(1)}_2)\given (E, A, X)}}\nend  &+ \abr{\EE\sbr{\vh^{(2)}_1(V, A, X) - \vh^{(2)}_2(V, A, X)\given (E, A, X)}}. \label{NC continuity 1}
    \end{align}
Also, recall that  
\begin{align*}
    \EE\sbr{W(D; h_1)\given E, A, X} &=\EE\sbr{\mathbb{1}\{Y \leq \vh_1(X,A)\}-\alpha\given E, A, X}\nend
    &=\int_{-\infty}^{\vh(A, X)}p(y|{E, A, X})dy  - \alpha.
\end{align*}
Thus, by the mean value theorem, we have 
\begin{align*}
    \abr{\EE\sbr{W(D; \vh^{(1)}_1) - W(D; \vh^{(1)}_2)\given (E, A, X)}} &\leq \sup_{t \in [0, 1]}\Big\{\abr{p_{Y\given E, A, X}\big(\vh^{(1)}_1(X, A) + t[\vh^{(1)}_2(X, A) - \vh^{(1)}_1(X, A)]\big)} \\
    &\qquad \cdot \abr{\vh^{(1)}_2(X, A) - \vh^{(1)}_1(X, A)}\Big\}\\
    &\leq \sup_{t \in [0, 1]}\Big|p_{Y\given E, A, X}\big(\vh^{(1)}_1(X, A) + t[\vh^{(1)}_2(X, A) - \vh^{(1)}_1(X, A)]\big)\Big|  \\
    &\qquad \cdot \sup_{x, a}\abr{\vh^{(1)}_1(x, a) - \vh^{(1)}_2(x, a)},
\end{align*}
which is upper bounded by
\begin{align}
\sup_{t \in [0, 1]}\Big|p_{Y\given E, A, X}\big(\vh^{(1)}_1(X, A) + t[\vh^{(1)}_2(X, A) - \vh^{(1)}_1(X, A)]\big)\Big|\cdot \nbr{\vh^{(1)}_1 - \vh^{(1)}_2}_{\infty}. \label{NC continuity 2}
\end{align}
Then, by observing that the $|\cT_1\vh_1 + \cT_1\vh_2|$ is upper bounded by $2L_h + 2,$ we have
    \begin{align*}
        \nbr{\cT\vh_1}_2^2 - \nbr{\cT\vh_2}_2^2 &\leq (2L_h + 2)\EE\sbr{|\cT\vh_1 - \cT\vh_2|}\nend
        &\leq (2L_h + 2)\sup_{t \in [0, 1]}\Big|p_{Y\given E, A, X}\big(\vh^{(1)}_1(X, A) + t[\vh^{(1)}_2(X, A) - \vh^{(1)}_2(X, A)]\big)\Big|\cdot \nbr{\vh^{(1)}_1 - \vh^{(2)}_1}_{\infty} \\ &+ (2L_h + 2)\nbr{\vh^{(2)}_1 - \vh^{(2)}_2}_{\infty},
    \end{align*}
where the last inequality holds by \eqref{NC continuity 1} and \eqref{NC continuity 2}. Hence we complete the proof.
\end{proof}
\end{lemma}

\section{Concentration Bounds}\label{app: bracketing concentration}
Throughout this section, we let $\phi:\cX\times\cA \rightarrow \RR^{H_n}$ and $\psi: \cX\times\cZ \rightarrow \RR^{J_n}$ be two feature maps. $H_n$ and $J_n$ represent the dimensions of the embedding spaces, where we assume exponential decay, i.e., $H_n$ and $J_n$ are both $\cO(\log n)$. We then define two linear spaces, $\vH$ and $\vTheta$, as follows:
\begin{align} \label{eq: linear space H}
    \cH=\{\cX\times \cA\rightarrow \beta_1^\top\phi(\cdot): \beta_1\in\RR^{H_n}, \nbr{\beta_1}_2\le C_1, \nbr{\phi}_{2, \infty}\le 1\},
\end{align}
\begin{align}\label{eq: linear space theta}
    \Theta=\{\cX\times \cZ\rightarrow \beta_2^\top \psi(\cdot) : \beta_2\in\RR^{J_n}, \nbr{\beta_2}_2 \le C_2, \nbr{\psi}_{2, \infty}\le 1\}.
\end{align}
    
The goal of this section is to verify that Condition \ref{con: high probability event} holds with this choice of $\vH$ and $\vTheta$ with $\eta_n = \tilde{\cO}(n^{-1/2})$. Recall that the event $\cE$ in Condition \ref{con: high probability event} is an intersection of two events of the uniform concentration bounds of some function classes. The plan is to first show that the event 
\begin{align*}
\cbr{\sup_{w(\cdot)\theta(\cdot) \in \cQ} \abr{\EE_{n}\sbr{W(D; h) \theta(\vX, \vZ)} - \EE\sbr{W(D; h) \theta(\vX, \vZ)}} \le \eta_n(\nbr{\theta}_{2}+\eta_n)}
\end{align*}
holds with high probability in \S\ref{sec: bracketing concentration}. Then we will show that the event
\begin{align*}
    \cbr{\abr{\nbr{\theta}^2_{n, 2}-\nbr{\theta}^2_{2}}\le \frac 1 2\rbr{\nbr{\theta}^2_{2}+\eta_n^2}, \forall \theta\in\Theta}
\end{align*}
holds with high probability in \S\ref{sec: critical radius bound}. Finally, we take a union bound of the two events to complete the proof of Condition  \ref{con: high probability event}.

\subsection{Concepts from the Empirical Process Theory}
To characterize the concentration bound, we first need to introduce some concepts in the empirical process theory.
\paragraph{Bracketing number.} Given two functions $f_1$ and $f_2$ such that $\nbr{f_1 - f_2} \leq t,$  A $t$-bracket $\sbr{f_1, f_2}$ is a subset of functions of the real-valued function class $\cF$ on $\cX$ satisfying for all $f \in \sbr{f_1, f_2}, \forall x \in \cX, f_1(x) \leq f(x) \leq f_2(x)$. The bracketing number $N_{[\,]}(t; \cF, \nbr{\cdot}_2)$ of a function class $\cF$ with respect to the norm $\nbr{\cdot}_2$ is the smallest number of $t$-brackets needed to cover $\cF$. 

\paragraph{Covering number.} The covering number $N(t; \cF, \nbr{\cdot}\|_{\bullet})$ of a function class $\cF$ with respect to some norm $\nbr{\cdot}\|_{\bullet}$ is the smallest number of $t$-balls in $\cF$ needed to cover $\cF$. 

\paragraph{Localized population Rademacher complexity.} The localized population Rademacher complexity with respect to $\{X_i\}_{i=1}^n$ and function class $\cF$ is defined as 
\begin{align*}
    \cR_{n}(\eta; \cF)=\E_{\{\varepsilon_i\}_{i=1}^n, \{X_i\}_{i=1}^n}\sbr{\sup_{f\in\cF, \nbr{f}_{2}\le \eta}\frac 1 n \sum_{i=1}^n\varepsilon_i f(X_i)}.
\end{align*}
where $\{\varepsilon_i\}_{i=1}^n$ are i.i.d. Rademacher random variables.

\paragraph{Critical radius.} Suppose $\cF$ is a real-valued function class on $\cX$ such that $\forall f \in \cF, \nbr{f}_\infty \le c$ for some constant $c\geq0.$ The critical radius of $\cF$ is the largest possible $\eta$ such that $\cR_{n}(\eta;\cF)\le \eta^2/c$.  
\subsection{Concentration bound of the Function Class $\cQ$ Using Bracketing Number} \label{sec: bracketing concentration}
Recall that we define the function class $\cQ$ as $\cQ  =\cbr{W(\cdot  ; h)\times\theta(\cdot ): \vh\in\vH, \theta\in\Theta}.$ In this subsection, we show that the first event of Condition \ref{con: high probability event} holds with high probability.
\begin{theorem} [Rate of Tail Bound of the Function Class $\cQ.$] \label{thm: rate tail bound Q}
If we choose $\cH$ as (\ref{eq: linear space H}) and $\Theta$ as (\ref{eq: linear space H}), then the event $\{\abr{\EE_{n}\sbr{W(D; h) \theta(\vX, \vZ)} - \EE\sbr{W(D; h) \theta(\vX, \vZ)}} \le \eta_n\rbr{\nbr{\theta}_{2}+\eta_n}\}$ holds with probability $1 - \xi$  for some $\eta_n = \tilde{\cO}(n^{-1/2}).$
\end{theorem}
\begin{proof} We proceed with the help of the following two lemmas.
\begin{lemma}[Bracketing Number of the Function Class $\cQ$]\label{lem: bracketing number Q}
    For every $t$ small enough, under Assumption \ref{asp: regularity of Density}, there exists $C_3 > 0$ such that 
    \begin{align}
    \log N_{[\,]} (t; \cQ, \nbr{\cdot}_2) \le A_n \cdot \log (C_3 / t ),
    \end{align}
    where $A_n = 4(H_n + J_n).$
    \begin{proof}
        See \S\ref{pro: Q bracketing} for a detailed proof.
    \end{proof}
\end{lemma}

\begin{lemma}[Lemma 3.11. in \cite{hu2020sharp}]\label{lem: bracketing concentration}
Let $\tilde{\cQ}$ be a class of functions uniformly bounded by one in $\nbr{\cdot}_2$ and $q_{0}$ be a fixed element in $\tilde{\cQ}.$ Let $\tilde{Q}(t)=\{q \in \tilde{\cQ}: \nbr{q- q_{0}}_2) \leq t\}$. Suppose $\log N_{[\,]} (t; \tilde{\cQ}, \nbr{\cdot}_2) \le A_n \log (C/t)$ for some $C>0.$ Then there exist universal positive  constants $C_4$, $C_5$, $C_6$, and $C_7$ such that for any $\xi>0,$
\begin{align}\label{eq: bracket outside ball}
     \prob\rbr{\sup_{\substack{q\in\tilde{\cQ};\\ \nbr{q-q_0}_2> \sqrt{\frac{A_n}{n}}}}\frac{|(\EE_n - \EE)(q)|}{\sqrt{\frac{A_n}{n}}\nbr{q-q_{0}}_2\log(C\nbr{q-q_{0}}_2^{-1})} \leq \frac{C_4}{A_n}\log(C_5/\xi)}
    \ge 1 - \xi/2.
\end{align}
Moreover, for $q\in \tilde{\cQ}(\sqrt{\frac{A_n}{n}}),$ we have 
\begin{align}\label{eq: bracket inside ball}
    \prob\rbr{\sup_{q\in \tilde{\cQ}(\sqrt{\frac{A_n}{n}})}\abr{(\EE_n - \EE)(q)} \leq  \frac{C_6\log(C_7/\xi)}{A_n \log^2(Cn/A_n)} \frac{A_n}{n}\log(Cn/A_n)} \geq 1 - \xi/2.
\end{align}
This lemma is a simple variant of Lemma 5.13 in \cite{geer2000empirical}.
\end{lemma}

By Lemma \ref{lem: bracketing number Q}, we can substitute $q_0$ as the zero function and $\tilde{\cQ}= \cQ/C_1$ in Lemma \ref{lem: bracketing concentration}. Then the event in \eqref{eq: bracket outside ball} concerns the supremum over $q \in \cQ$ and $\nbr{q}_2 \geq C_1\sqrt{A_n/n}.$ Since when $\nbr{q}_2 \geq C_1\sqrt{A_n/n},$ $\log(C_3 C_1/\nbr{q}_2) \leq \log(C_3\sqrt{n/A_n}),$ \eqref{eq: bracket outside ball} becomes:
\begin{align} \label{eq: bracket outside ball 2}
     \prob\left(\sup_{\substack{q\in \cQ;\\ \nbr{q}_2> C_1\sqrt{\frac{A_n}{n}}}}\frac{\abr{(\EE_n - \EE)(q)}}{\sqrt{\frac{A_n}{n}}\nbr{q}_2\log\rbr{C_3\sqrt{\frac{n}{A_n}}}} \leq \frac{C_4}{A_n}\log(C_5/\xi)\right)
    \ge 1 - \xi/2.
\end{align}
Moreover, the restriction $q\in \tilde{\cQ}(\sqrt{A_n/n})$ of the event in \eqref{eq: bracket inside ball} is replaced with $q\in \cQ(C_1\sqrt{A_n/n}).$ We now set 
\begin{align*}
\eta_n = & \max\cbr{\sqrt{\frac{A_n}{n}}\log\left( C_3 \sqrt{\frac{n}{A_n}} \right)\cdot \frac{C_4}{A_n} \cdot \log\left(\frac{C_5}{\xi}\right),~~    \bigg [\frac{C_1 A_n}{n}\cdot \frac{C_6\log(C_7/\xi)}{A_n \log^2(C_3n/C_1A_n)} \cdot \log(C_3n/C_1 A_n) \bigg]^{1/2}}.
\end{align*}
Then combine \eqref{eq: bracket inside ball} and \eqref{eq: bracket outside ball 2} and take the union bond, we conclude that 
\begin{align*}
\prob &\rbr{\sup_{w(\cdot), \theta(\cdot) \in \cQ} \abr{\EE_{n}\sbr{W(D; h) \theta(\vX, \vZ)} - \EE\sbr{W(D; h) \theta(\vX, \vZ)}} \le \eta_n\left(\nbr{\theta}_{2}+\eta_n\right)} \\
    &\geq \prob \left( \sup_{w(\cdot), \theta(\cdot) \in \cQ\left(C_1\sqrt{\frac{A_n}{n}}\right)} \abr{\EE_{n}\sbr{W(D; h) \theta(\vX, \vZ)} - \EE\sbr{W(D; h) \theta(\vX, \vZ)}} \le \eta_n^2, \right. \\
    &\qquad \left. \sup_{\nbr{w(\cdot)\theta(\cdot)}_2 \geq C_1\sqrt{\frac{A_n}{n}}} \abr{\EE_{n}\sbr{W(D; h) \theta(\vX, \vZ)} - \EE\sbr{W(D; h) \theta(\vX, \vZ)}} \le \eta_n\nbr{\theta}_2 \right) \\
    &\geq 1 - \xi.
\end{align*}
We complete the proof by noticing that  $\eta_n = \tilde{\cO}(n^{-1/2})$ by choice.
\end{proof}

\subsection{Concentration Bound Using Critical Radius} \label{sec: critical radius bound}
We now demonstrate that there exists $\eta_n = \tilde{\cO}(n^{-1/2})$ such that the event $\{|\nbr{\theta}^2_{n, 2}-\nbr{\theta}^2_{2}|\le \frac 1 2\rbr{\nbr{\theta}^2_{2}+\eta_n^2}, \forall \theta\in\Theta\}$ holds with high probability when $\Theta$ is chosen to be the linear space \ref{eq: linear space theta}. Subsequently, we can then take the union bound to verify Condition \ref{con: high probability event}.
\begin{lemma} \label{lem: localization law of theta}
    For any $\xi>0,$ there exists $\eta_n = \tilde{\cO}(n^{-1/2})$ such that the event $\{|\nbr{\theta}^2_{n, 2}-\nbr{\theta}^2_{2}|\le \frac 1 2\rbr{\nbr{\theta}^2_{2}+\eta_n^2}, \forall \theta\in\Theta\}$ holds with probability $1-\xi$ when $\Theta$ is chosen to be the linear space (\ref{eq: linear space theta}).
\begin{proof}
We proceed with the help of two lemmas:
\begin{lemma}[Lemma F.4 in \cite{chen2023unified}]\label{lem: covering number linear}
The covering numbers $N(t;\cH, \nbr{\cdot}_{\infty})$ and $N(t;\Theta, \nbr{\cdot}_{\infty})$ are upper bounded by $\rbr{1+2C_1/t}^{H_n}$ and $\rbr{1 + 2C_2/t}^{J_n},$ respectively.
\end{lemma}
\begin{lemma}[Theorem 14.1 in \cite{wainwright2019high}]\label{lem: 2nd order localized consistency}
Let $t_n$ be the critical radius of $\Theta.$ There exist universal positive constants $k_0$ and $k_1$ such that for any $\eta_n \ge t_n + k_0\sqrt{\log(k_1/\xi)/n}$, we have
$$|\nbr{\theta}_n^2-\nbr{\theta}_2^2| \le \frac{1}{2}\nbr{\theta}_2^2+\frac{1}{2}\eta_n^2, \quad \forall \theta\in\Theta$$
with probability at least $1-\xi$.
\end{lemma}
     By Lemma \ref{lem: covering number linear}, we can substitute $\Theta$ for $\cF$ in \ref{lem: critical radius & covering number}. Hence the critical radius of $\Theta$ is bounded by $\cO(\sqrt{(J_n\log n)/n}) = \tilde{\cO}(n^{-1/2}).$ We then use Lemma \ref{lem: 2nd order localized consistency} to complete the proof.
\end{proof}
\end{lemma}
\begin{theorem}[Main Theorem of Concentration Bound]\label{thm: concentration bound_linear}
    Condition \ref{con: high probability event} holds when choosing $\vH$ and $\vTheta$ as in (\ref{eq: linear space H}) and (\ref{eq: linear space theta}) with some $\eta_n = \tilde{\cO}(n^{-1/2})$. 
\begin{proof}
    By Theorem \ref{thm: rate tail bound Q}, we have 
    \begin{align*}
        \prob \rbr{\abr{\EE_{n}\sbr{W(D; h) \theta(\vX, \vZ)} - \EE\sbr{W(D; h) \theta(\vX, \vZ)}} \le \eta_n\left(\nbr{\theta}_{2}+\eta_n, \forall \vh\in \vH, \forall \theta\in\Theta\right)} \geq 1 - \xi.
    \end{align*}
    Combine with Lemma \ref{lem: localization law of theta} and take the union bound, and we complete the proof.
\end{proof}
\end{theorem}

\section{Supporting Lemmas of  \S\ref{app: bracketing concentration}}\label{app: bracketing tech}
In what follows, we introduce and prove supporting lemmas concerning the bracketing and covering numbers of certain function classes of interest, as discussed in \S\ref{app: bracketing concentration}. These lemmas play a crucial role in verifying Condition \ref{con: high probability event}.
\subsection{Proof of Lemma \ref{lem: bracketing number Q}} 
\begin{proof}\label{pro: Q bracketing}
    We prove this Lemma with the help of the two Lemmas below.
    \begin{lemma}\label{lem: bracketing number linear} The bracketing numbers $N_{[\,]}(t; \Theta, \nbr{\cdot}_2)$ is upper bounded by $\rbr{1+4C_1/t}^{J_n}.$
\begin{proof}
Note that $\cH$ is indexed by the set $\tilde{\beta} = \{ \beta_2 \in \RR^{J_n} : \nbr{\beta_2}_2 \le C_2 \}$. Let $F(x,z) \equiv 1$. For any $\beta_2^T\psi(\cdot), \tilde{\beta}_2^T\psi(\cdot) \in \vTheta$, i.e., $\beta_2, \tilde{\beta}_2 \in \tilde{\beta}$, we have 
$$
|\beta_2^T \psi(x, z) - \tilde{\beta}_2^T \psi(x, z)| \leq \nbr{\beta_2 - \tilde{\beta}_2}_2 \nbr{\psi}_{2, \infty} F(x,z) \leq \nbr{\beta_2 - \tilde{\beta}_2}_2
$$
almost surely in $(X, Z)$. So by Theorem 2.7.11 in \citet{van2013weak}, 
$$
N_{[\,]}(t; \Theta, \nbr{\cdot}_2) \leq N (t/2;\tilde{\beta}, \nbr{\cdot}_2) \leq N (t/2; \Theta, \nbr{\cdot}_2) \leq N(t/2;\Theta, \nbr{\cdot}_{\infty}) \leq \rbr{1 + 4C_2/t}^{J_n},
$$
where the last step leverages Lemma \ref{lem: covering number linear}. 
\end{proof}
\end{lemma}

\begin{lemma}[Theorem 3 \& Example 5.1 in \citet{chen2003estimation}]\label{lem: bracketing number indicator} 
Let the indicator function class be
\begin{align*}
    \cW  :=\cbr{W(\cdot  ; h): \vh\in\vH}.
\end{align*}
Suppose Assumption \ref{asp: regularity of Density} holds. Then the bracketing number $N_{[\,]}(t; \cW, \nbr{\cdot}_2)$ is upper bounded by $\rbr{1+8C_1 K^2/t^2}^{H_n}$ for $t$ small enough, where $K > 0$ uniformly upper bounds $\sup_{y}\pr_{Y\given A, X, Z}(y)$ for almost all $(A, X, Z).$ 
\begin{proof}
In the proof of Theorem 3 in \citet{chen2003estimation}, for $t$ small enough, with Assumption \ref{asp: regularity of Density}: 
$$
N_{[\,]}(t; \cF_2, \nbr{\cdot}_2) \le N ((t/2K)^{\frac{1}{s}};\vTheta', \nbr{\cdot}_{\infty}) N ((t/2K)^{\frac{1}{s}}; \vH, \nbr{\cdot}_{\infty}).
$$ 
In our case of indicator functions, as also indicated in Example 5.1 in \citet{chen2003estimation}, choose $s = 1/2$ and $\cF_2 = \cW$. Let $\vTheta' = \{\alpha\}$ and combine with Lemma \ref{lem: covering number linear}, we have 
$$
N_{[\,]} (t, \cW, \nbr{\cdot}_2) \le 1 \cdot N ((t/2K)^2; \cH, \nbr{\cdot}_\infty) \le \rbr{1 + \frac{8C_1 K^2}{t^2} }^{H_n}.
$$
Therefore, we complete the proof.
\end{proof}
\end{lemma}
We prove the upper bound of the function class $\cQ$ by combining the upper bounds of the function classes $\vTheta$ and $\cW$. Consider $t \in (0,1)$ small enough, we have
\begin{align*}
    \log N_{[\,]} (t, \cQ, \nbr{\cdot}_2) &\le \log N_{[\,]} (t/2; \cW, \nbr{\cdot}_2) + N_{[\,]} (t/2; \vTheta, \nbr{\cdot}_2) \\
    &\le H_n \log (1 + 8C_1 K^2/t^2) + J_n \log (1 + 4C_2/t) \\
    &\le (H_n + J_n) \log \rbr{ (1 + 8C_1 K^2/t^2)(1 + 4C_2/t^2) } \\
    &\le 2(H_n + J_n)\log \rbr{\frac{1 + \max \{ 8C_1 K^2, 4C_2 \} }{t^2}},
\end{align*}
where the second inequality holds by Lemma \ref{lem: bracketing number linear} and Lemma \ref{lem: bracketing number indicator}.  We then let $A_n = 4(H_n + J_n)$ and $C_3 = \sqrt{1 + \max \{ 8C_1 K^2, 4C_2} \}$ to complete the proof.
\end{proof}

\begin{lemma}[Lemma F.7 in \cite{chen2023unified}]\label{lem: critical radius & covering number} 
    If the covering number of $C_2$-uniformly bounded function class $\cF$ satisfies $\log N(t;\cF, \nbr{\cdot}_{\infty})\le A\log\rbr{1+2C_2/t}$ for some constant $A$, the critical radius of $\cF$ is upper bounded by $\cO(\sqrt{(A\log n)/n})$.
\end{lemma}

\end{document}